%% file: tro25_pinns_habich.tex
\documentclass[journal]{IEEEtran}

\IEEEoverridecommandlockouts                              

\newcommand\CPP{C\nolinebreak[4]\hspace{-.05em}\raisebox{.4ex}{\relsize{-3}{\textbf{++}}}}
\usepackage{cite}
\usepackage{amsmath,amssymb,amsfonts}
\DeclareMathOperator*{\minimize}{minimize}
\usepackage{algorithmic}
\usepackage{graphicx}
\usepackage{textcomp}
\usepackage{gensymb}
\usepackage{xcolor}
\usepackage{color}		
\usepackage{multirow}		
\usepackage{array}										
\usepackage{colortbl}
\usepackage{siunitx}
\usepackage[english]{babel}
\usepackage{mathtools}
\usepackage[hidelinks]{hyperref} 
\usepackage{blindtext}
\usepackage[ruled, linesnumbered]{algorithm2e}
\usepackage{pifont}

\title{
Generalizable and Fast Surrogates:\\Model Predictive Control of Articulated Soft \\Robots using Physics-Informed Neural Networks
}

\author{\highlightredfinal{Tim-Lukas Habich, Aran Mohammad, Simon F. G. Ehlers, Martin Bensch, Thomas Seel, and Moritz Schappler}%
\thanks{\highlightredfinal{All authors are with the Leibniz University Hannover, Institute of Mechatronic Systems, 30823 Garbsen, Germany
	(corresponding author: Tim-Lukas Habich, e-mail: \href{mailto:habich@imes.uni-hannover.de}{habich@imes.uni-hannover.de}).}}%
\thanks{\highlightredfinal{This work was supported by the Deutsche Forschungsgemeinschaft (DFG, German Research Foundation) --- 433586601 and INST 187/742-1 FUGG.}}%
}

\newif\ifcopyright
	\copyrighttrue
	
\newif\ifhighlightchanges

\ifhighlightchanges
\newcommand{\highlightredfinal}[1]{\textcolor{red}{#1}}
\else
\newcommand{\highlightredfinal}[1]{#1}
\fi

\newcommand{\highlightred}[1]{#1}
\definecolor{imesorange}{rgb}{0.902,0.478,0.157} 
\definecolor{imesblue}{rgb}{0,0.31,0.6}
\definecolor{imesgreen}{rgb}{0.78,0.82,0.09}
\newcommand{\mm}[1]{\boldsymbol{#1}}

\newcommand{\orange}[1]{\textcolor{imesorange}{#1}}

\newcommand{\e}[2]{\begin{equation} #1 \label {eq:#2} \end{equation}}
\newcommand{\mc}[1]{\mathcal{#1}}
\newcommand{\ind}[1]{\mathrm{#1}}
\newcommand{\R}{\mathbb{R}}
\usepackage[mathscr]{euscript}
\newcommand{\fra}[1]{{\mathscr{F}}_{#1}}

\newcommand{\transpose}{^\mathrm{T}}
\definecolor{Gray}{gray}{0.85}

\newcolumntype{M}[1]{>{\centering\arraybackslash}m{#1}}
\newcolumntype{N}{@{}m{0pt}@{}}
\makeatletter
\newcommand{\removelatexerror}{\let\@latex@error\@gobble}
\makeatother

\begin{document}
\ifcopyright
\thispagestyle{empty}
\pagestyle{empty}
{\LARGE IEEE Copyright Notice}
\newline
\fboxrule=0.4pt \fboxsep=3pt

\fbox{\begin{minipage}{1.1\linewidth}  
\textcopyright\,\,2025\,\,IEEE. Personal use of this material is permitted. Permission from IEEE must be obtained for all other uses, in any current or future media, including reprinting/republishing this material for advertising or promotional purposes, creating new collective works, for resale or redistribution to servers or lists, or reuse of any copyrighted component of this work in other works. \\

Accepted to be published in: IEEE Transactions on Robotics (T-RO), 2025.\\

DOI: 10.1109/TRO.2025.3631818
		
\end{minipage}}
\else
\fi
\graphicspath{{./images/}}

\maketitle
\thispagestyle{empty}
\pagestyle{empty}

\begin{abstract}
Soft robots can revolutionize several applications with high demands on dexterity and safety.
When operating these systems, real-time estimation and control require fast and accurate models.
However, prediction with first-principles (FP) models is slow, and learned black-box models have poor generalizability.
Physics-informed machine learning offers excellent advantages here, but it is currently limited to simple, often simulated systems without considering changes after training.
We propose physics-informed neural networks (PINNs) for articulated soft robots (ASRs) with a focus on data efficiency.
The amount of expensive real-world training data is reduced to a minimum --- one dataset in one system domain.
Two hours of data in different domains are used for a comparison against two gold-standard approaches:
In contrast to a recurrent neural network, the PINN provides a high generalizability.
The prediction speed of an accurate FP model is exceeded with the PINN by up to a factor of~\highlightred{467} at slightly reduced accuracy.
This enables nonlinear model predictive control (MPC) of a pneumatic ASR.
\highlightred{Accurate position tracking with the MPC running at 47 Hz is achieved in six dynamic experiments.}
\end{abstract}
\section{Introduction}\label{intro}
Building soft robots has been an emerging research field for several years. 
In contrast to conventional rigid robots, they are made of significantly softer materials. 
The resulting compliance makes them suitable robot candidates for intrinsically safe interaction with humans, as less damage is caused in the event of a collision~\cite{Rus.2015}. 
Modeling and controlling such rubber-like robots is challenging mainly due to complex geometries, material nonlinearities, and air compressibility in case of pneumatic actuation~\cite{Xavier.2022}.
Therefore, handcrafted models built with conventional first-principles approaches lack accuracy. 
Even if the accuracy is high due to advanced modeling/identification techniques:
\textit{Prediction with first-principles models is slow} due to numerical integration with small time steps.
This prevents the application of these models in real-time estimation and control.

\begin{figure}[tbp]
		\centering
		\resizebox{1\linewidth}{!}{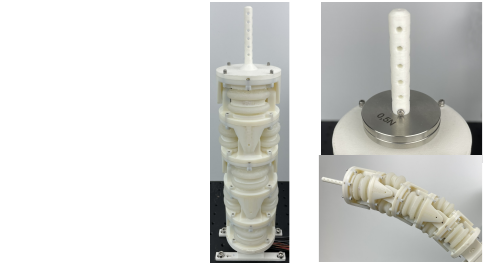}
		\caption{(a)~The objective of this work is to solve the trade-off between model accuracy/generalizability and prediction speed in the field of soft robotics by using physics-informed machine learning (ML). (b)~During training, data from \textit{one} training domain is available. Trained surrogate models extrapolate for changed system dynamics in \orange{unseen test domains}. We consider changing the payload and the orientation of the robot base as possible modifications during operation.} \label{fig:cover}
\end{figure}

In contrast, \textit{learning-based modeling} only requires input-output data of the system, and the prediction with learned forward models is possible with large time steps. 
Therefore, high prediction speeds are possible and enable such real-time applications.
However, a \textit{huge amount of real-world data} is necessary.
The recording is not only expensive and time-consuming, but there is another central problem:
\textit{Learned models only show good performance within the seen data space} (interpolation). 
For changing domains after training (extrapolation), there is usually poor generalization for such black-box approaches~\cite{Nelles.2020}.
Combining the advantages of both modeling worlds in a \textit{hybrid strategy} (and omitting the disadvantages), as sketched in Fig.~\ref{fig:cover}(a), is a widespread goal in various research fields.

Within soft robotics, hybrid approaches incorporating both physical knowledge and machine learning are a highly promising line of research~\cite{DellaSantina.2023} and an unexplored area~\cite{GeorgeThuruthel.2018}.
\highlightred{Laschi et al.~\cite{Laschi.2023} conclude that\textit{ traditional modeling techniques must be incorporated into existing learning strategies} for new advancements in the soft-robotics field.
Recently, Falotico et al.~\cite{Falotico.2024} also declare the integration of physical principles into machine-learning approaches as a perspective for solving current issues in terms of accuracy and computational efficiency.
They \textit{explicitly mention PINNs}, which is the focus of our work.}

As an emerging strategy, PINNs~\cite{Raissi.2019} have been used in a wide variety of applications to solve ordinary differential equations~(ODEs), e.g., for modeling nonlinear structural systems, and partial differential equations~(PDEs) such as Navier-Stokes equations~\cite{Cuomo.14.01.2022}. 
We use PINNs to provide fast and accurate surrogate models of articulated soft robots.
One central requirement is the \textit{generalization to unseen dynamics}, which were not present when recording training data.
\highlightred{Examples are }an additional mass or a changed base orientation, illustrated in Fig.~\ref{fig:cover}(b).

\highlightred{The \textit{four major scientific contributions} of this article are:
	\begin{enumerate}
		\item For the first time, the original form of physics-informed neural networks --- introduced by Raissi et al.~\cite{Raissi.2019} and formulated for state-space modeling in~\cite{Antonelo.2024,Krauss.2024} --- is applied to a real soft robot with multiple degrees of freedom (DoF).
		For this purpose, a first-principles model of our ASR is derived and identified, which is very accurate, but the forward simulation requires significant computing time.
		This justifies the need for a fast surrogate model with high accuracy to enable real-time applications of model-based estimation/control.
		\item Two existing PINN architectures (PINC~\cite{Antonelo.2024} and DD-PINN~\cite{Krauss.2024}) are extended to achieve generalizability despite system changes after training.
		During training, we perform a systematic hyperparameter optimization (HPO).
		This is often neglected in the literature due to excessive training times of several days.
		\item The proposed model is compared against the gold standard of both modeling worlds, namely a hyperparameter-optimized recurrent neural network (RNN) and an identified FP model, regarding prediction speed and accuracy. For this purpose, we recorded two hours of real-world data with variable soft-robot dynamics.
		\item To demonstrate one possible use case for the proposed PINN, a real-time application is realized.
		We choose nonlinear MPC (NMPC) of the real soft robot, which places high demands not only on prediction accuracy but also on prediction speed due to online optimization. For the first time, NMPC with PINNs is realized for multi-DoF soft robots and validated in several dynamic real-world experiments, including comparison with a baseline controller.
	\end{enumerate}}
\highlightred{To enable reproducibility, we further provide the community with \textit{open-source contributions}\footnote{\label{foot:pinn}\highlightredfinal{\url{https://tlhabich.github.io/sponge/pinn_mpc}}}. This includes the whole codebase for \highlightredfinal{PINN training (PINC and DD-PINN), hyperparameter optimization and learning-based NMPC with PINNs} for possible applications within and beyond soft robotics.
The two hours of experimental data \highlightredfinal{(13 real-world datasets of the articulated soft robot)} are also publicly available~\cite{HabichData.2025}.}

\highlightred{In sum, we make \textit{three claims}: 
First, our surrogate model outperforms an accurate physics-driven model in terms of prediction speed by orders of magnitude with only slightly reduced accuracy.
Second, by incorporating physical knowledge during training, the PINN achieves higher prediction accuracies and generalization to out-of-distribution data compared to an RNN.
Throughout this article, all models receive data from only one domain during training/identification and are tested in unseen test domains, as depicted in Fig.~\ref{fig:cover}(b).
Third, accurate learning-based nonlinear MPC with PINNs is enabled for different soft-robot dynamics without the need to retrain the system model or to retune the controller.
This is intended to represent one practical scenario with high demands on prediction speed and accuracy. 
\textit{All claims are proven experimentally.}}
Please refer to the supplementary video of this article to see the soft robot and dynamic movements during control.

\highlightred{The remainder of this article is organized as follows: 
After an overview of related work (Sec.~\ref{relatedwork}), preliminaries are introduced (Sec.~\ref{preliminaries}). 
State-space modeling of the soft robot using first principles and PINNs is presented, followed by the architecture for NMPC (Sec.~\ref{main_pinn}).
In Sec.~\ref{experiments}, the identification and learning results are shown, and all models are compared regarding prediction speed and accuracy.
Also, control experiments are presented.
The article ends with conclusions, including future work directions (Sec.~\ref{conclusions}).} 

\section{Related Work}\label{relatedwork}
\highlightred{A brief overview of the two modeling worlds, namely first-principles modeling~(Sec.~\ref{first_princ_model}) and black-box learning~(Sec.~\ref{soa_blackbox}), is given.
This is followed by presenting directions of hybrid learning~(Sec.~\ref{hybrid_soa}), whose various approaches combine physical knowledge with machine learning.
Further, related work on physics-informed ML for (soft) robots~(Sec.~\ref{PI_machinelearning_soa}) are discussed.}

\highlightred{\subsection{First-Principles Modeling}\label{first_princ_model}}
\highlightred{Soft robots can be divided according to two design paradigms: 
soft continuum robots~(SCRs) vs. ASRs~\cite{GeorgeThuruthel.2018,DellaSantina.2020}.
Depending on the robot design, a different modeling approach is recommended, which is described briefly below.}

\highlightred{Due to the deformable structure along a continuous backbone, SCRs have infinite DoF.
Approaches such as piecewise constant curvature~\cite{Webster.2010}, rod models~\cite{Alessi.2024}, or finite element method (FEM)~\cite{Coevoet.2017} are therefore required for kinematic and dynamic modeling, to name just a few.
For a structured overview, please refer to~\cite{Armanini.2023}.}

\highlightred{The focus of this article is on ASRs. 
	They comprise a vertebrate-like structure with rigid links and compliant joints.
	The use of traditional methods for kinematics, such as the Denavit-Hartenberg notation, and multi-body dynamics, such as the Lagrange formalism, are applicable here.
	Lumped mass models are therefore recommended for such systems~\cite{DellaSantina.2023}, which have been used in various publications for ASRs~\cite{Best.2015,Hoffmann.2023,Chhatoi.2023}.}

\highlightred{Whether ASR or SCR, measurement data from the real robot is usually used for the identification of unknown system parameters.
	This is somewhat similar to black-box learning but with a strong inductive bias on the model architecture derived from physics.
	Thus, \textit{high generalizability and data efficiency} can be achieved with physics-based approaches.
	However, first-principles models can have a low accuracy due to simplifying assumptions.
	The \textit{main disadvantage of accurate first-principles models for soft robots is the slow prediction speed}.
	This is due to the numerical integration of forward models with very fine step sizes, which can complicate model-based estimation and control.}

\subsection{Black-Box Learning}\label{soa_blackbox}
Within soft robotics, data-driven approaches for control are still in their infancy~\cite{Wang.2021b}, with much progress having been made in recent years. 
In~\cite{Braganza.2007}, \textit{feedforward neural networks} (NNs) were first applied to feedforward control of soft extensible continuum manipulators. 
Many works also use feedforward NNs for MPC, e.g., of a soft actuator with one DoF~\cite{Gillespie.2018}, a soft continuum joint~\cite{Cheney.2024}, and a six-DoF soft robot~\cite{Hyatt.2019,Hyatt.2020}.
In~\cite{Habich.2023}, Gaussian processes are trained to realize learning-based position and stiffness feedforward control of a soft actuator.

Nonlinear material behavior and friction cause hysteresis effects that can be captured using \textit{recurrent neural networks}~\cite{Lipton.29.05.2015}. 
These are used in~\cite{Thuruthel.2017} to learn the forward dynamics of a soft manipulator with a nonlinear autoregressive exogenous model. 
Only an open-loop predictive control policy was implemented, which was further developed in~\cite{Thuruthel.2019} for closed-loop control.
Data-driven MPC was successfully realized for a single actuator by using long short-term memory (LSTM) units~\cite{Luong.2021} and for the ASR from Fig.~\ref{fig:cover}(b) via gated recurrent units (GRUs)~\cite{Schafke.2024}.

Recording real-world data requires a lot of effort, and trained networks only achieve \textit{good accuracy within the seen data space}.
In case of system changes, training must be repeated with new data due to\textit{ poor extrapolation}. 
In addition, convergence is more difficult for high-dimensional NNs required to model multi-DoF robots. 
For this reason, the prediction of the entire state vector was not possible with the chosen network architecture and the existing data in~\cite{Hyatt.2019}.
Thus, only the velocity at the next time step was learned. 
Data-driven control of a soft robot using the Koopman operator is realized in~\cite{Bruder.2021}, and, more recently, a data-efficient method based on neural ODEs~\cite{Chen.2018} was used to model a soft manipulator~\cite{Kasaei.2023}. 
However, the authors do not use physical knowledge, which results in poor generalization.

For \textit{different payloads}, reinforcement learning using an LSTM network as a forward-dynamics model was conducted in~\cite{Centurelli.2022}.
However, even for only $\nu{=}1$ varying model parameter (payload), different real-world datasets with variable payload conditions are necessary as training data.
We do not believe that such black-box approaches are scalable for real-world applications, i.e., $\nu{>}1$: 
For the minimum requirement of two levels per model parameter (e.g., minimum and maximum payload), $2^{\nu}$~datasets are already required.
This implies that for variations of several parameters, a time-consuming and expensive data acquisition is necessary.
In addition, some parameters require finer variations and, therefore, more than two levels due to poor interpolation.  
\subsection{Directions of Hybrid Learning}\label{hybrid_soa}
\highlightred{In summary, both first-principles and black-box approaches have problems \textit{when generalizability and prediction speed are considered jointly}.
The term generalizability describes the aim for high accuracy not only in known data domains but also for unseen system domains with changed dynamics.
Hybrid approaches are promising here and apply \textit{physical knowledge in combination with black-box learning}}.

Within the pioneering work of Nguyen-Tuong and Peters for rigid robots~\cite{NguyenTuong.2010}, knowledge of the parametric dynamics model is incorporated into the nonparametric Gaussian process via mean or kernel function. 
Finite-element models have been used to analyze soft robotic segments, thus generating vast amounts of (simulated) training data for feedforward NNs~\cite{Runge.2017}. 
External loads such as gravity or contact forces are not taken into account. 

\textit{Residual/Error learning} of physical models using data is one main direction of hybrid learning.
This has been done for a soft continuum joint~\cite{Johnson.2021}, a soft parallel robot~\cite{Huang.2024}, and soft continuum robots~\cite{Reinhart.2017, Gao.2024, Lou.2024, Jiahao.2024}. 
The main requirement for these approaches is that \textit{the physical model must be efficiently evaluated} during real-time control or estimation, which is often not the case for accurate models of soft robots \highlightred{(cf.~Sec.~\ref{first_princ_model})}.
Another possible disadvantage is the need for real experimental data from all domains.
This indicates, due to poor extrapolation, that a new error model must be learned if the system changes after training. 
Consequently, the authors of~\cite{Lou.2024} collect training data in several domains (different payloads between $\SI{0}{\gram}$ and~$\SI{300}{\gram}$) in order to ensure generalization within this data region (interpolation).
However, we argue that trained models should extrapolate across the available training data and that data acquisition for every possible system change is not practical.
In line with this, it was recently concluded that the \textit{generalization beyond bounded training data and the handling of novel dynamical events} should be examined in future work~\cite{Gao.2024}.
Also, the authors of~\cite{Huang.2024} declare robust control with learned models in changing environments as future work.

Alternatively to error learning, \textit{prior knowledge for constraining NN training} results in less data required. 
This was realized in~\cite{NEUMANN.2013} using extreme learning machines with only one hidden layer. 
Due to the simple network structure, linear constraints, such as the monotonic behavior of the drives or physical restrictions of the soft robot, can be taken into account using quadratic optimization. 
In~\cite{Tariverdi.2021}, a soft manipulator was discretized according to continuum-mechanics principles, and recurrent neural networks were trained for each node, taking applied forces and moments for every node as inputs. 
The authors conclude that the supervised learning used might not entirely respect underlying physics, and \textit{trained models are not applicable to changing dynamics without retraining}.

The aforementioned problems are countered by \textit{physics-informed machine learning}~\cite{Karniadakis.2021}, also known as scientific machine learning~\cite{Cuomo.14.01.2022}. 
This is done by including physical knowledge into the training process so that \textit{trained regressors will be aware of governing physical laws}. 
Regarding modeling and control of dynamical systems, physics-informed ML is an alternative approach compared to classical gray-box modeling: 
Instead of (or in addition to) identifying parameters of white-box models consisting of simplified modeling assumptions, physics-informed ML comes from the black-box side without modeling simplifications and is equipped with additional constraints such as symmetries, causal relationships, or conservation laws~\cite{Nghiem.2023}.
Underlying ODEs/PDEs can act as learning bias during the training process, limiting the high-dimensional search space and, therefore, the necessary amount of real-world data while still maintaining a high generalization. 
This is beneficial since real-world experiments are expensive, and the \textit{training data will rarely represent all possible system conditions during inference}.
Besides high generalizability despite less data, trained surrogate models are considerably faster to evaluate since they do not rely on fine spatiotemporal discretization required for numerical integration of conventional ODEs/PDEs. 
This significantly speeds up the solution of hard optimization problems, such as during nonlinear MPC, which is outlined in a recent overview~\cite{Nghiem.2023} as one of the main opportunities for physics-informed~ML.

\subsection{Physics-Informed Machine Learning in (Soft) Robotics}\label{PI_machinelearning_soa}
For rigid systems, physics-inspired networks using Lagrangian and Hamiltonian mechanics were proposed~\cite{Lutter.2019,Lutter.05.10.2021}. 
Mass matrix, centrifugal, Coriolis, gravitational, and frictional forces as well as (potential and kinetic) energies are learned unsupervised by minimizing the ODE residual. 
In~\cite{Nicodemus.2022}, PINNs are employed for the nonlinear MPC of a simulated rigid robot with two degrees of freedom.

Within soft robotics, hybrid RNNs using responses from the physical model as additional inputs for state observation of one-DoF soft actuators (McKibben actuator and soft finger) are utilized in~\cite{Sun.2022}. 
A physics-informed simulation model is trained from finite-element data and is used as \textit{efficient-to-evaluate surrogate model} within MPC in~\cite{Lahariya.2022}. 
A simulated one-DoF soft finger is considered, and the authors identify the transfer to physical prototypes and extension to multi-DoF tasks as a future research direction. 
Similarly, PINNs are trained as fast surrogate models of soft robotic fingers in~\cite{Wang.2024,Beaber.2024}.
To enable fast prediction of Cosserat rod statics, PINNs are trained for one simulated tendon-driven segment of a continuum robot~\cite{Bensch.2024}.

\highlightred{A real one-segment tendon-driven soft robot was modeled in~\cite{Liu.2024} using Lagrangian neural networks (LNNs) and a real-world dataset.
This has two drawbacks.
Since the network is trained with offline-recorded data, the generalization to unseen dynamics (due to system changes not present in the training data distribution) would be poor.
Also, LNNs still require numerical integration to predict future states since only system quantities (such as inertia or damping) are learned with neural networks.
For this purpose, the authors~\cite{Liu.2024} use Runge-Kutta integration.
Similar to the forward integration of our FP model, a slow prediction speed results due to necessary fine step sizes.}

The authors of~\cite{Yoon.2024} propose a kinematics-informed neural network for pneumatic and tendon-driven soft robots with one segment. 
The generalization ability is examined for motor commands, which are unseen during network training. 
Although this is already an important study, generalization to changed systems would be even more crucial. 
Learning system dynamics instead of pure kinematics would also be advantageous for optimal control, which is realized in~\cite{Chhatoi.2023} for an ASR using differential dynamic programming. 
Thereby, forward dynamics are simulated via inversion of the inertia matrix and they declare the realization of MPC as future work. 
For our robot, forward simulation via inversion of the inertia matrix and numerical integration of the \textit{stiff ODE} is very time-consuming. 
Therefore, a fast and accurate surrogate model is necessary for MPC. 

In~\cite{Mendenhall.2024}, experimental data is extended with simulated finite-element data of \textit{out-of-distribution} (unseen) loading/deformation cases to train a PINN of a single-bellows actuator. 
The main motivation here is to build a fast surrogate of a computationally expensive finite-element model.
Although the NN is 435 times faster compared to the FEM, one simulation still takes ${>}\SI{20}{\second}$ for the parallel actuator consisting of five bellows. 
This computation time further increases for soft robots with more bellows making it unsuitable for real-time applications. 
Also, only statics are modeled to predict deformations.

Most similar to our proposed approach, the authors of~\cite{Wang.2024c} formulate a physics-informed LSTM network with a \textit{variable system property as model input}. 
This allows the training of the model for unseen system properties, which are not present when recording training data. 
However, it is only applied to a simple one-DoF mass-spring-damper system.
Control is also not considered.

\highlightred{To summarize, related soft-robotics work only considers simple (often simulated) actuators and not PINNs for state-space models~\cite{Antonelo.2024,Krauss.2024}, which are specifically designed for real-time MPC.
To the best of our knowledge, no published work considers PINNs for fast MPC of --- neither real nor multi-DoF --- soft robotic systems.
Dynamics changes after training are also predominantly neglected.}

\section{Preliminaries}\label{preliminaries}
This section covers the basics of RNNs~(Sec.~\ref{rnns}) and PINNs~(Sec.~\ref{pinns_prel}). 
The method used for optimizing their hyperparameters~(Sec.~\ref{hpo_prel}) is also briefly described.
\subsection{Recurrent Neural Networks}\label{rnns}
In contrast to utilizing physical knowledge, forward dynamics can be trained with black-box approaches on existing training data.
\highlightred{Recurrent neural networks are the gold standard for time-series learning and can approximate dynamical system behavior by incorporating information from previous time steps.}
They can represent state-space models~\cite{Schussler.2019} and are promising for learning long-term dependencies, such as hysteresis. 
The predicted states
\e{[\mm{h}(T_\ind{s}),\hat{\mm{x}}(T_\ind{s})]=\mm{f}_\mathrm{RNN}(\mm{h}_0,\mm{x}_0,\mm{u}_0)}{RNN}
for one step with sample time $T_\ind{s}$ can be computed given hidden state $\mm{h}_0{=}\mm{h}(0)$, initial state $\mm{x}_0{=}\mm{x}(0)$, and input $\mm{u}_0{=}\mm{u}(0)$ at time\footnote{For better comparison, the continuous notation with time $t$ is used throughout the article for all models, although the implementation of RNNs is discrete.} $t{=}0$.
\highlightred{Note that we use the current state $\mm{x}_0$ as network input due to the MPC context of this work.}

Gated recurrent units~\cite{Cho.2014} are well suited for time-series learning due to the ability to prevent vanishing and exploding gradients during backpropagation.
\highlightred{Such network architecture outperformed LSTM networks for the used ASR~\cite{Schafke.2024}.
We, therefore, use GRUs as the reference method for a purely data-driven approach in this work.}

\highlightred{The network is trained batch-wise via the common backpropagation-through-time (BPTT) algorithm~\cite{Werbos.1990}, which is briefly described in the following.}
During simulation with known initial values $\mm{x}_0$ and $\mm{u}_0$, the hidden state is initialized to $\mm{h}_0{=}\mm{0}$ and then recursively passed for each time step in the same way as the predicted state $\hat{\mm{x}}(T_\ind{s})$ for a given input trajectory $\mm{u}(t) \forall t$.
\highlightred{This behavior is imitated during training on a given dataset, such that the network predictions are compared to the ground-truth data.
Since the RNN outputs are recursively passed as inputs for the next time step, backpropagation must be done through the whole time series, which requires extensive computational effort and memory.
This makes it often unfeasible in practice for larger datasets~\cite{Nelles.2020}.
Multi-step truncation enables a feasible approximation of the gradient~\cite{Nelles.2020}, whereby BPTT is performed only for subsets (batches) of the whole training set.
We perform such truncated BPTT for each batch of $n_\ind{b}$ datapoints and detach the gradients of $\hat{\mm{x}}$ and $\mm{h}$ at the beginning of each new batch.
For each batch, the Adam optimizer adjusts the weights depending on the error between the predictions and the ground-truth data.}


\subsection{Physics-Informed Neural Networks}\label{pinns_prel}
Purely black-box models require a large amount of experimental data, which is very expensive (e.g., costs for staff and maintenance of the system) to acquire in the real world. 
Usually, the available data only describe a few operating points, and trained regressors, therefore, only perform well within these data regions. 
PINNs~\cite{Raissi.2019} provide an excellent opportunity in this \textit{small-data regime} to incorporate existing domain knowledge and thus regularize the training process by additional physics-based loss terms. 
Here, the basic idea is that the loss function for optimizing the network weights does not only consist of a data loss $\mc{L}_\ind{d}$ as in RNNs but also takes into account additional knowledge using physics~$\mc{L}_\ind{p}$ and initial-condition/boundary $\mc{L}_0$ losses. 
The additional losses are evaluated on a finite set of \textit{sampled collocation points} in the entire input space and do not require additional real-world datasets. 
This leads to encoding physical knowledge into the NNs to achieve broad generalizability.

Based on this idea, Antonelo et al.~\cite{Antonelo.2024} reformulate such PINN for control (PINC) to enable simulations with variable, long-range time horizons.
These PINCs are designed for fast MPC of dynamical systems with state-space models from first principles
\e{\hat{\dot{\mm{x}}}(t)=\mm{f}_\ind{FP}(\mm{x}(t),\mm{u}(t))}{FP_SSM}
with states $\mm{x}(t)$ and inputs $\mm{u}(t)$. 
This is done by the use of several NN inputs such as the time $t$, initial state $\mm{x}_0$, and control action $\mm{u}_0$. 
Prediction for indefinitely long horizons is realized by recursively feeding back the PINN output
\e{\hat{\mm{x}}(T_\ind{s})=\mm{f}_\ind{PINN}(T_\ind{s},\mm{x}_0,\mm{u}_0)\approx{{\mm{x}}}(T_\ind{s})}{antonelo_pinn} 
as initial value $\mm{x}_0$ at the subsequent time step with new input~$\mm{u}_0$. 
The latter is considered constant within each time interval $t{\in}[0,T_\ind{s})$ (zero-order hold assumption). 
Therefore, the NN is only trained for small time steps, and forecasting is done via \textit{self-loop prediction}, which is similar to RNNs.  

The physics loss is determined by means of collocation points evaluated with (\ref{eq:antonelo_pinn}) and compared with (\ref{eq:FP_SSM}) using automatic differentiation $\frac{\partial \hat{\mm{x}}}{\partial t}$.
In addition, the initial condition 
\e{\mm{x}_0=\mm{f}_\ind{PINN}(0,\mm{x}_0,\mm{u}_0)}{b} 
is learned using an initial-condition loss. 
In fact, no real data was used for training a surrogate model to solve the ODEs of two benchmark systems (Van-der-Pol oscillator and four-tank system) in~\cite{Antonelo.2024}.
Their main motivation is the development of an efficient-to-evaluate model instead of the time-consuming numerical integration of (\ref{eq:FP_SSM}). 
This is due to the fact that PINNs are trained to directly compute $\hat{\mm{x}}(T_\ind{s})$, cf. (\ref{eq:antonelo_pinn}), while integrators such as Runge-Kutta methods require considerably smaller step sizes due to instabilities. 

The high prediction speed is accompanied by a considerably increased training effort of several days compared to RNNs.
Especially for dynamical systems with several states, the PINC training is very time-consuming.
To counter this, the domain-decoupled PINN (DD-PINN) was proposed in~\cite{Krauss.2024}, which approximates the solution
\e{\hat{\mm{x}}(T_\ind{s})=\mm{f}_\ind{PINN}(T_\ind{s},\mm{x}_0,\mm{u}_0)=\mm{x}_0+\mm{a}(\mm{\alpha}(\mm{x}_0,\mm{u}_0),T_\ind{s})}{DD-PINN-Pred}
with an ansatz function
\begin{multline}\label{eq:Ansatz_DD-PINN}
\mm{a}(\mm{\alpha},t){=}\sum_{i=1}^{n_\ind{a}} \mm{\alpha}_{1i}{\odot}\big(\exp{({-}\mm{\alpha}_{4i}t)}{\odot}\sin{(\mm{\alpha}_{2i}t{+}\mm{\alpha}_{3i})}\\
{-}\sin{(\mm{\alpha}_{3i})} \big)
\end{multline}
using Hadamard product~(${\odot}$) and element-wise functions $\exp{(\cdot)}$ and $\sin{(\cdot)}$. 
\highlightred{As solution~\cite{Krauss.2024}, a superposition of damped harmonic oscillations with amplitude $\mm{\alpha}_{1i}$, damping term $\exp{({-}\mm{\alpha}_{4i}t)}$ and oscillation term $\sin{(\mm{\alpha}_{2i}t{+}\mm{\alpha}_{3i})}$ with angular frequency $\mm{\alpha}_{2i}$ and phase shift $\mm{\alpha}_{3i}$ is assumed.
The subtraction of $\sin{(\mm{\alpha}_{3i})}$ in (\ref{eq:Ansatz_DD-PINN}) ensures the compliance with the initial condition $\hat{\mm{x}}(0){\equiv}\mm{f}_\ind{PINN}(0,\mm{x}_0,\mm{u}_0){\equiv}\mm{x}_0$ due to $\mm{a}(\mm{\alpha},0){\equiv}\mm{0}$.}

In constrast to the PINC~(\ref{eq:antonelo_pinn}), only the ansatz vector
\e{\mm{\alpha}=[\mm{\alpha}_1\transpose,\mm{\alpha}_2\transpose,\mm{\alpha}_3\transpose,\mm{\alpha}_4\transpose]\transpose\in\R^{4\ind{dim}(\mm{x})n_\ind{a}}}{Ansatz_params}
is predicted by the feedforward NN.
Each $\mm{\alpha}_j{=}[\mm{\alpha}_{j1}\transpose,\hdots,\mm{\alpha}_{j n_\ind{a}}\transpose]\transpose$ consists of $n_\ind{a}$ vectors of length $\ind{dim}(\mm{x})$.
The \highlightred{key} advantage is that the training converges much faster~\cite{Krauss.2024}.
\highlightred{This is mainly} because the computationally expensive automatic differentiation is avoided.
\highlightred{Instead, $\frac{\partial \hat{\mm{x}}}{\partial t}{=}\dot{\mm{a}}(\mm{\alpha},t)$} is calculated in closed form by time differentiation of (\ref{eq:Ansatz_DD-PINN}), which results to
\begin{multline}\label{eq:Ansatz_DD-PINN_diff}
	\dot{\mm{a}}(\mm{\alpha},t){=}\sum_{i=1}^{n_\ind{a}} \mm{\alpha}_{1i}{\odot}\exp{({-}\mm{\alpha}_{4i}t)}{\odot}\big({-}\mm{\alpha}_{4i}{\odot}\sin{(\mm{\alpha}_{2i}t{+}\mm{\alpha}_{3i})}\\
	{+}\cos{(\mm{\alpha}_{2i}t{+}\mm{\alpha}_{3i})}{\odot}\mm{\alpha}_{2i} \big)
\end{multline}
with element-wise function $\cos{(\cdot)}$.
\highlightred{As already described in the last paragraph,} by decoupling the time domain $t$ from the feedforward NN and constructing the ansatz function (\ref{eq:Ansatz_DD-PINN}), the initial condition is always maintained. 
\highlightred{We therefore do not need a separate initial-condition loss, as $\mc{L}_0{\equiv}0$ applies.
This further simplifies the network training.}
Both PINC and DD-PINN are extended for variable domains in this work so that generalization can be achieved even when the system changes after training.

\subsection{Hyperparameter Optimization}\label{hpo_prel}
The hyperparameter optimization of the NNs in this work is conducted with the asynchronous successive halving algorithm~(ASHA)~\cite{Li.} to obtain optimal network architectures. 
ASHA samples within the given boundaries of the hyperparameters and starts multiple training runs based on the available computing resources. 
By monitoring the validation loss of each hyperparameter configuration (trial) during training, poorly performing configurations are stopped by choosing a reduction factor, and the resources are used for new training runs.
All trials are minimally trained until the user-defined grace period is reached.
The latter is a crucial parameter and should be chosen sufficiently large, as otherwise, good trials with slower convergence will be rejected.
In contrast, a grace period that is too long unnecessarily extends the time required for the HPO.
After a defined number of trials, the hyperparameters of the training result with the lowest validation loss are chosen as the optimal set.

\section{\highlightred{PINNs for Articulated Soft Robots}}\label{main_pinn}
\begin{figure}[tbp]
		\centering
		\resizebox{\linewidth}{!}{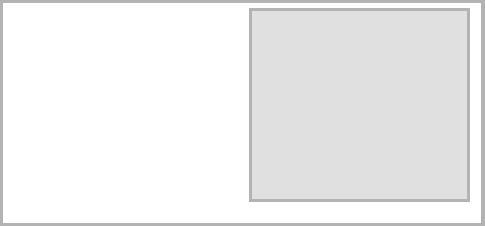}
		\caption{\highlightred{Graphical overview of the article's main part (Sec.~\ref{main_pinn}).}} \label{fig:overview_main}
	\end{figure}

\highlightred{The content of this section is graphically illustrated in Fig.~\ref{fig:overview_main}.
After describing the soft-robot platform~(Sec.~\ref{snake_robot}), a state-space model is obtained using first principles and system identification~(Sec.~\ref{first_princ}).
Based on this, the generalizable and fast surrogates are introduced~(Sec.~\ref{pinn_subsection}) including implementation details~(Sec.~\ref{pinn_impl}) and a note on their practical usability~(Sec.~\ref{dom_know}).
The proposed PINNs are integrated into a nonlinear MPC as \textit{one possible real-time application}~(Sec.~\ref{mpc}).}

\subsection{Soft-Robot Platform}\label{snake_robot}
\begin{figure}[tbp]
		\centering
		\resizebox{1\linewidth}{!}{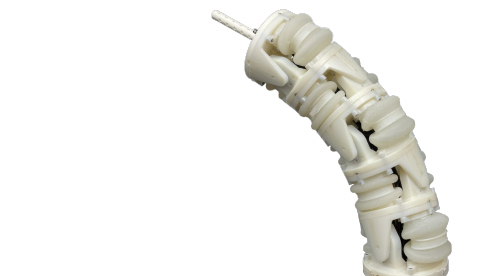}
		\caption{(a) Kinematic chain of the articulated soft robot with $n{=}5$ rotational actuators of height $h{=}\SI{53.4}{\milli\meter}$ and coordinate frames $\fra{i}$. In this work, dynamics are changed by attaching a variable mass $m_\ind{e}$ to the last segment with mass $m_5$ and by changing the base orientation $\beta_\ind{g}$ against gravity. (b)~Real robot with joint angles $q_i$ and antagonistic actuation via pneumatic pressures $p_{i1}$ and $p_{i2}$.} \label{fig:kinematic_chain}
\end{figure}

The open-source soft pneumatic robot with antagonistic bellows from~\cite{Habich.2024} is used in this work with $n{=}5$ stacked actuators, which is shown in Fig.~\ref{fig:kinematic_chain}. 
Antagonistically arranged cast bellows with air pressures $\mm{p}_i{=}[p_{i1},p_{i2}]\transpose$ actuate each joint $i$. 
Bellows pressures ${\mm{p}}{=}[\mm{p}_1\transpose,\dots,\mm{p}_n\transpose]\transpose$ and joint angles $\mm{q}{=}[q_1,\dots,q_n]\transpose$ are measured.
The latter is (first-order) low-pass filtered with \SI{1}{\hertz}, and joint velocities~$\dot{\mm{q}}$ are obtained online via numerical differentiation. 
Desired bellows pressures $\mm{p}_\ind{d}$ are set using external proportional valves. 

Further information regarding the semi-modular ASR can be found in the supporting video of~\cite{Habich.2024}.
The published design has been minimally optimized with regard to various points: joint-friction reduction via smaller shaft diameters, reconstruction of the frames to reduce plastic deformation, increased maximum working pressures by using thicker bellows, and larger tube diameters for faster pressure dynamics. 
The improved version is freely available\footnote{\highlightredfinal{\url{https://tlhabich.github.io/sponge/designs/semi_modular/main}}}.
To further reduce friction, we removed the cable/tube guides in each actuator.

The additional end-effector mass~$m_\ind{e}$ and the base orientation~$\beta_\ind{g}$ can be varied, summarized in the \textit{domain variable}~$\mm{\delta}{=}[m_\ind{e},\beta_\ind{g}]\transpose$. 
Data $\mm{D}_\ind{t}$ with several training samples of~$[\mm{q}\transpose,\dot{\mm{q}}\transpose,\mm{p}\transpose,\mm{p}_{\ind{d}}\transpose]\transpose$ in the training domain~$\mm{\delta}_\ind{t}$ (constant additional mass and constant base orientation) is acquired for the identification/training of the FP~model (Sec.~\ref{first_princ}), black-box model (Sec.~\ref{rnns}) and the hybrid model (Sec.~\ref{pinn_subsection}). 
\textit{Afterward}, the domain is modified, which aims to represent a change in the real system that occurs after recording training data. 
Our method is intended to generalize even for such changes that are unknown during training. 
As a general illustration of our method, $\nu{=}\ind{dim}(\mm{\delta}){=}2$ different parameters are investigated. 
This can be extended arbitrarily.

\subsection{First-Principles Modeling and Identification}\label{first_princ}
\subsubsection{Robot Dynamics}

Forward kinematics of the serial robot are obtained using Denavit-Hartenberg notation. 
Based on this, dynamics
\e{\overbrace{\underbrace{\mm{\tau}(\mm{p}){=}\mm{g}(\mm{q}){+}\mm{s}(\mm{q})}_{\ind{I}}{+}\mm{M}(\mm{q})\ddot{\mm{q}}{+}\mm{c}(\mm{q},\dot{\mm{q}}){+}\mm{d}(\dot{\mm{q}})}^{\ind{II}}{+}\mm{b}(\mm{q},\dot{\mm{q}})}{dyn}
are modeled by means of Euler-Lagrange equations with gravitational~$\mm{g}$, Coriolis/centrifugal~$\mm{c}$, and inertial torques~$\mm{M}\ddot{\mm{q}}$ with mass matrix $\mm{M}$ and accelerations $\ddot{\mm{q}}$.
The latter is only necessary for offline system identification and is determined via offline low-pass filtering and (two-times) numerical differentiation of~$\mm{q}$.

To capture joint friction and the bellows behavior, (\ref{eq:dyn}) additionally contains viscous and Coulomb friction $\mm{d}{=}\mm{k}_\ind{v}{\odot}\dot{\mm{q}}{+}\mm{k}_\ind{C}{\odot}\tanh(\dot{\mm{q}}\pi/\dot{q}_\ind{C})$, and stiffness torques $\mm{s}{=}\mm{k}_\ind{s}{\odot}\mm{q}$. 
The hyperbolic tangent with threshold $\dot{q}_\ind{C}$ is used instead of the discontinuous signum function. 

For modeling the contact torques $\mm{b}{=}[b_1,\dots,b_n]\transpose$ at the compliant joint limits, we adapt the nonlinear model of contact normal force~\cite{Azad.2014}, which is an improved version of the Hunt-Crossley model~\cite{Hunt.1975}. 
This leads to
\e{b_{i}=\left\{
	\begin{array}{ll}
		+|\Delta q_i|^{3/2}k_\ind{bs}{+}\sqrt{|\Delta q_i|}\dot{q}_i k_\ind{bd} & \textrm{for}\,q_i>q_\ind{bt} \\
		-|\Delta q_i|^{3/2}k_\ind{bs}{+}\sqrt{|\Delta q_i|}\dot{q}_i k_\ind{bd} & \textrm{for}\,q_i<-q_\ind{bt} \\
		0 &  \textrm{otherwise}\\
	\end{array}
	\right.}{}
for each joint $i$ with $\Delta q_i{=}q_i{-}q_\ind{bt}$, threshold of the soft boundaries $q_\ind{bt}$, and contact parameters $\mm{k}_\ind{b}{=}[k_\ind{bs},k_\ind{bd}]\transpose$.

\begin{figure}[tbp]
		\centering
		\resizebox{1\linewidth}{!}{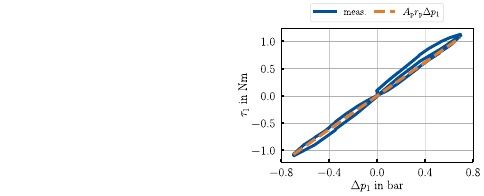}
		\caption{Validation of the actuation model: (a)~Experimental setup with torque sensor attached to the first actuator. (b)~The mapping from pressures to torque with factor $A_\ind{p}r_\ind{p}$ applies for the entire pressure range.} \label{fig:KMS_experiment}
\end{figure}

The left-hand side of~(\ref{eq:dyn}) consists of pressure differences $\Delta\mm{p}{=}[\Delta p_1,\dots,\Delta p_n]\transpose$ with $\Delta p_i{=}p_{i1}{-}p_{i2}$, which are mapped to joint torques~$\mm{\tau}{=}A_\ind{p}r_\ind{p}\Delta\mm{p}$ via pressure area~$A_\ind{p}{=}\SI{639.8}{\milli\meter^2}$ and lever arm~$r_\ind{p}{=}\SI{24.1}{\milli\meter}$ of each actuator. 
The simple linear actuation model is validated experimentally by coupling a torque sensor to the first actuator, which is shown in Fig.~\ref{fig:KMS_experiment}.
For an exemplary pressure trajectory, it is shown that the actuation model is valid for the whole pressure range with maximum pressure $p_\ind{max}{=}\SI{0.7}{\bar}$.

\subsubsection{Identification Parameters}
The dynamics equation~(\ref{eq:dyn}) consists of several unknown parameters 
\e{\mm{k}=[\mm{k}_\ind{s}\transpose,\mm{k}_\ind{v}\transpose,\mm{k}_\ind{C}\transpose,\mm{k}_\ind{b}\transpose]\transpose\in\R^{3n+2}.}{ident_params} 
All other parameters are either adopted from CAD software (actuator height, center-of-gravity positions, and entries of the inertia tensors) or measured (segment masses), whereby the bellows, sensor cables, and tubes are neglected. 
The kinematic and inertia parameters of the first $n{-}1$ actuators are equal due to the repetitive structure.

Note that due to the casting process with high reproducibility, the contact parameters $\mm{k}_\ind{b}$ are assumed to be identical for all joints.
The same simplification could be applied to the parameters $\mm{k}_\ind{s}, \mm{k}_\ind{v}$ and $\mm{k}_\ind{C}$ for all joints.
However, the tubes/cables of the last actuator have to overcome considerably higher friction than those of the first actuator.
Similarly, the tubes/cables may also influence the joint stiffness. 
Therefore, we identify different stiffness and friction parameters for each joint.

A three-step least-squares identification is conducted, whereby we choose $\dot{q}_\ind{C}{=}\SI{1}{\degree/\second}$ and ${q}_\ind{bt}{=}\SI{10}{\degree}$ heuristically. 
The identification dataset $\mm{D}_\ind{t}$ is split into three subsets: $\mm{D}_\ind{tI}$, $\mm{D}_\ind{tII}$, and $\mm{D}_\ind{tIII}$. 
Each subset is used to identify different parameters of~(\ref{eq:ident_params}). 
Such a multi-step identification is common to isolate the influence of the different static/dynamic terms and improve the identification result. 
The procedure is explained in the following.

\subsubsection{Identification of Stiffness}
The factors \mbox{$\mm{k}_\ind{s}{=}[k_{\ind{s}1}\dots,k_{\ind{s}n}]\transpose$} are identified using static data inside the soft boundaries: $\mm{D}_\ind{tI}{\subset}\mm{D}_\ind{t}$ with $|\dot{q}_i|{\leq}\dot{q}_\ind{C}{\approx}0\land|{q}_i|{\leq} q_\ind{bt}$. 
The dynamics equation simplifies to (\ref{eq:dyn})-$\ind{I}$ for these data points, as velocity- and acceleration-dependent terms are assumed to be negligible, and contact torques are zero.
This results in the parameter-linear form
\e{\mm{\tau}_{\ind{I}j}=\mm{\tau}_j-\mm{g}_j=\mm{Q}_{\ind{I}j}\mm{k}_\ind{s}}{parameter-linear-I}
for an arbitrary measurement $j$. 
Vertically stacking\footnote{Note that for each datapoint $j$, the dataset categorization ($\ind{I,\,II,\,III}$) is done separately for each actuator $i$. The rows of $\mm{Q}_\diamond$ and $\mm{\tau}_\diamond$ are therefore deleted if the conditions for the respective dataset $\mm{D}_\ind{t\diamond}$ are not fulfilled.
	This ensures that all data is used and only occurs once in all datasets.
	Such separation of the dataset neglects the coupling between the different robot segments.
	If this simplification does not apply to other systems, suitable identification trajectories must be recorded.\label{ident_I_II_III}} $\mm{\tau}_{\ind{I}j}$ and $\mm{Q}_{\ind{I}j}{=}\ind{diag}(\mm{q}_j)$ for all measurements leads to $\mm{\tau}_\ind{I}$ and $\mm{Q}_\ind{I}$, respectively. 
The latent parameters $\mm{k}_\ind{s}{=}\mm{Q}_\ind{I}^\dagger\mm{\tau}_{\ind{I}}$ are obtained by Moore-Penrose pseudo-inversion ($\dagger$).
\subsubsection{Identification of Friction}
The friction parameters $\mm{k}_\ind{vC}{=}[k_{\ind{v}1},k_{\ind{C}1},\dots,k_{\ind{v}n},k_{\ind{C}n}]\transpose$ are identified using dynamic data inside the soft boundaries: $\mm{D}_\ind{tII}{\subset}\mm{D}_\ind{t}$ with $|\dot{q}_i|{>}\dot{q}_\ind{C}{\land}|{q}_i|{\leq} q_\ind{bt}$. 
The dynamics equation simplifies to~(\ref{eq:dyn})-$\ind{II}$ for these data points, which can be transformed into the parameter-linear form
\begin{equation}
	\label{eq:parameter-linear-II}
	\begin{split}
		\mm{\tau}_{\ind{II}j}&=\mm{\tau}_j-\mm{g}_j-\mm{s}_j-\mm{M}_j\ddot{\mm{q}}_j-\mm{c}_j \\
		& = \underbrace{\begin{bmatrix}
				\dot{q}_{1j}\quad f_\ind{C}(\dot{q}_{1j})&\mm{0}&\mm{0} \\
				\mm{0}&\ddots & \mm{0}\\
				\mm{0}&\mm{0}&\dot{q}_{nj}\quad f_\ind{C}(\dot{q}_{nj})
		\end{bmatrix}}_{\mm{Q}_{\ind{II}j}\in\R^{n\times 2n}}
		\mm{k}_\ind{vC}
	\end{split}
\end{equation}
by using $f_\ind{C}(\dot{q}){=}\tanh(\dot{q}\pi/\dot{q}_\ind{C})$.
Thereby, the stiffness torques $\mm{s}_j$ are computed with the \textit{previously identified} parameters $\mm{k}_\ind{s}$.
Similar to above, $\mm{\tau}_\ind{II}$ and $\mm{Q}_\ind{II}$ are obtained by vertically stacking\footref{ident_I_II_III} $\mm{\tau}_{\ind{II}j}$ and $\mm{Q}_{\ind{II}j}$, respectively. 
This enables the computation of the friction parameters $\mm{k}_\ind{vC}{=}\mm{Q}_\ind{II}^\dagger\mm{\tau}_{\ind{II}}$.

\subsubsection{Identification of Contact Parameters}
The factors $\mm{k}_\ind{b}$ are identified using data outside the soft boundaries $\mm{D}_\ind{tIII}{\subset}\mm{D}_\ind{t}$ with $|{q}_i|{>}q_\ind{bt}$. 
The entire dynamics equation (\ref{eq:dyn}) applies to these data points, which can be transformed into the parameter-linear form
\begin{equation}
	\label{eq:parameter-linear-III}
	\begin{split}
		\mm{\tau}_{\ind{III}j}&=\mm{\tau}_j-\mm{g}_j-\mm{s}_j-\mm{M}_j\ddot{\mm{q}}_j-\mm{c}_j-\mm{d}_j \\
		& = \underbrace{\begin{bmatrix}
				\ind{sgn}(q_{1j})|\Delta q_{1j}|^{3/2}&\sqrt{|\Delta q_{1j}|}\dot{q}_{1j}\\
				\vdots&\vdots \\
				\ind{sgn}(q_{nj})|\Delta q_{nj}|^{3/2}&\sqrt{|\Delta q_{nj}|}\dot{q}_{nj}
		\end{bmatrix}}_{\mm{Q}_{\ind{III}j}\in\R^{n\times 2}}
		\mm{k}_\ind{b}
	\end{split}
\end{equation}
using \textit{previously identified} parameters of stiffness and friction. 
The parameters $\mm{k}_\ind{b}{=}\mm{Q}_\ind{III}^\dagger\mm{\tau}_{\ind{III}}$ are computed in analogy to above\footref{ident_I_II_III}.

\subsubsection{Forward Prediction}
The inverse dynamics (\ref{eq:dyn}) can be transformed into state-space form
\e{\hat{\dot{\mm{x}}}(t)=\mm{f}_{\ind{FP}\mm{\delta}}(\mm{x},\mm{u},\mm{\delta})}{state-space}
with additional domain input~$\mm{\delta}$.
We denote $\mm{x}{=}[\mm{q}\transpose,\dot{\mm{q}}\transpose]\transpose{\in}\R^{2n}$ as state of the dynamical system and $\mm{u}{=}\mm{p}_\ind{d}{\in}\R^{2n}$ as system input. 
Thereby, $\mm{p}_\ind{d}{=}\mm{p}$ is assumed.
This is valid due to the fast pressure control of the proportional valves so that the desired and measured pressures match closely with a time delay of $\SI{10}{}{-}\SI{80}{\milli\second}$.
If this simplification does not hold, the pressure dynamics could be modeled via a simple first-order system.

Conventional numerical integration of (\ref{eq:state-space}) by using Runge-Kutta methods can be used to predict the evolution of the states.  
However, very small step sizes ${\leq}\SI{100}{\micro\second}$ are necessary for stable forward simulation of the system due to the stiff ODE. 
This is impractical for use in real-time nonlinear MPC not only due to the excessive computation time, which is further evaluated in Sec.~\ref{performance}.
Also, to enable online optimization during control, the prediction horizon is usually bounded to a few time steps. 
The fine discretization of the model would, therefore, lead to a very short prediction horizon, which considerably reduces the time available to solve an MPC problem.
Thus, \textit{a fast-to-evaluate surrogate model with large time steps is necessary for real-time control}.
\subsection{\highlightred{Generalizable and Fast PINNs}}\label{pinn_subsection}
Both first-principles (Sec.~\ref{first_princ}) and data-driven modeling \mbox{(Sec.~\ref{rnns})} have disadvantages.
The former has good generalizability.
However, prediction with numerical integration is computationally expensive. 
The latter learns input-output relationships purely from real data, ignoring any physical principles. 
This leads to poor generalizability, as trained models are strongly overfitted to one (training) domain of the system.
However, numerical integration with fine discretization is not necessary, as it enables fast inference with large time steps.
We combine the advantages of both approaches in the following.
For this purpose, physics-informed neural networks provide an excellent architecture.

The original approach for state-space modeling using PINNs~(\ref{eq:antonelo_pinn}) consists of a network with three inputs: time $t$, initial state $\mm{x}_0$, and input signal $\mm{u}_0$. 
It is trained on artificially sampled collocation points, for which the ODE does not need to be solved.
Real-world data can additionally be used if the first-principles model is not accurate enough.
According to~\cite{Wang.02.07.2021}, a trustworthy and reliable physics-informed model should be able to extrapolate to systems with different parameters, external forces, or boundary conditions while maintaining high accuracy.
However, both PINC~\cite{Antonelo.2024} and DD-PINN~\cite{Krauss.2024} represent the state space for one selected parameterization of the dynamical system, but not if this domain changes after training.
For example, the NNs can be trained only for \textit{one} domain (e.g., one particular mass at the end effector $m_\ind{e}$ and base orientation $\beta_\ind{g}$).
Instead, we would like to train a PINN that generalizes to \textit{all} realistic domains during inference to avoid time-consuming data acquisition and retraining. 
For this, the domain~$\mm{\delta}$ is defined as another input of the PINN, which can be any quantity of the first-principles model.

	\begin{figure}[tbp]
		\centering
		\resizebox{\linewidth}{!}{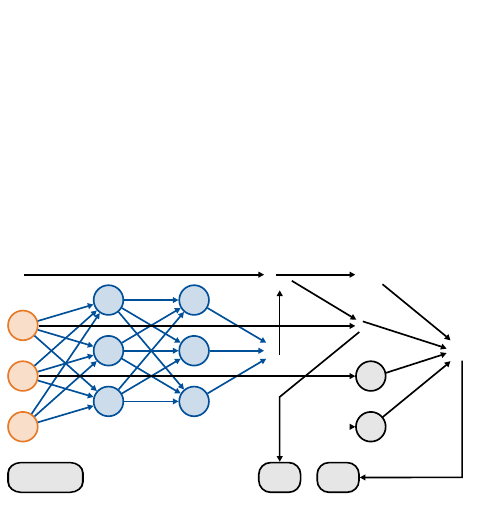}
		\caption{PINN structures with inputs in orange, feedforward network in blue with output in green. Both networks are extended by an additional domain input $\mm{\delta}$: (a)~The PINC directly predicts the state $\hat{\mm{x}}$. During training, the network requires computationally expensive automatic differentiation for each collocation point (in each training epoch) and contains an additional loss term~$\mc{L}_\ind{0}$ for the initial condition. (b)~The DD-PINN predicts the ansatz vector~$\mm{\alpha}$ of an ansatz function~$\mm{a}(\mm{\alpha},t)$. The latter can be differentiated in closed form. Also, $\mm{a}(\mm{\alpha},0){\equiv}\mm{0}$ applies so that no initial-condition loss is necessary. Both drastically speed up the training time.} \label{fig:pinn_structures}
	\end{figure}

Similar to (\ref{eq:state-space}), the forward pass of the proposed PINN architecture results in
\e{\hat{\mm{x}}(t)=\mm{f}_{\ind{PINN}\mm{\delta}}(t,\mm{x}_0,\mm{u}_0,\mm{\delta}),}{pinn_forward}
which is visualized in Fig.~\ref{fig:pinn_structures} for both PINC and DD-PINN. 
The feedforward NN consists of $n_\ind{h}$~hidden layers, each with $n_\ind{n}$~neurons and hyperbolic tangent activation function. 
Whereas the feedforward part of the PINC directly predicts~$\hat{\mm{x}}(t)$, the ansatz vector $\mm{\alpha}$ (\ref{eq:Ansatz_params}) is computed within the DD-PINN, which is then used to calculate the state
\e{\hat{\mm{x}}(t)=\mm{x}_0+\mm{a}(\mm{\alpha}(\mm{x}_0,\mm{u}_0,\mm{\delta}),t).}{DD-PINN-Pred-domain}

Collocation points are sampled in $\R^{1{+}4n{+}\nu}$ with user-specified boundaries\footnote{All inputs and outputs of the NN are scaled between $-1$ and $1$ by using this user-specified minimum/maximum values. For the sake of clarity, we do not introduce new symbols for each variable since min-max scaling is straightforward and must be considered during implementation.} $\mc{T}{\subset} \R$, $\mc{X}_0{\subset} \R^{2n}$, $\mc{U}_0{\subset} \R^{2n}$ and $\mc{D}{\subset} \R^\nu$.
In this way, \textit{domains can be trained even if no real data is available for that domain}. 
During prediction, the current domain $\mm{\delta}$ can then be specified.
We use latin hypercube sampling within the range ${\pm} 1$ for sampling the scaled quantities $t$, $\mm{u}_0$, and $\mm{\delta}$. 
For the scaled states~$\mm{x}_0$, we utilize a multivariate normal distribution with zero mean and standard deviation of $0.4$.
This importance sampling has the advantage that there are fewer collocation points at the broadly selected sampling boundaries, and, for example, more samples are taken to consider the high slope of the Coulomb friction term for $\dot{q}_i{\approx}0$.

\highlightred{Note that we could also provide the RNN with an additional domain input, i.e., $\mm{f}_{\mathrm{RNN}\mm{\delta}}(\mm{h}_0,\mm{x}_0,\mm{u}_0,\mm{\delta})$. 
	However, this has no added value in the restrictive context of this work, as we only provide training data from one domain $\mm{\delta}_\ind{t}$. 
	Therefore, the RNN would be trained with a permanently constant input $\mm{\delta}{=}\mm{\delta}_\ind{t}$.
	Purely data-based approaches are not able to learn the influence of the domain input here, so that it can be omitted.
	The domain input for the RNN would only be appropriate for training with datasets from different discrete domains ($\nu{>}1$).
	However, we do not believe that it is practical to record training data in all possible domains (cf.~Sec.~\ref{soa_blackbox}).
	This, again, underlines the motivation for utilizing physical knowledge.}

The total (multi-objective) loss $\mc{L}$ is composed of several terms ($\mc{L}_\ind{d},\mc{L}_\ind{p}$ and $\mc{L}_\ind{0}$), whose weights ($\eta_\ind{d},\eta_\ind{p}$ and $\eta_\ind{0}$) are calculated after the first training epoch (cf. Sec.~\ref{pinn_impl}).
The mean squared error (MSE) between network \textit{prediction} and \textit{ground truth} is determined, whereby the determination of both quantities varies depending on the loss.

\subsubsection{Physics Loss $\mc{L}_\ind{p}$}
On collocation points, the residuum
\e{\mm{r}_\ind{p}=\frac{\partial\hat{\mm{x}}}{\partial t}-\mm{f}_{\ind{FP}\mm{\delta}}(\hat{\mm{x}},\mm{u}_0,\mm{\delta})}{r_physics}
is calculated with the modeled system dynamics (\ref{eq:state-space}), zero-order hold assumption for $\mm{u}_0$, and the time derivative of the predicted states $\frac{\partial\hat{\mm{x}}}{\partial t}{=}\frac{\partial}{\partial t}\mm{f}_{\ind{PINN}\mm{\delta}}$.
The latter is done by means of automatic differentiation for the PINC, and can be computed in closed form for the DD-PINN via $\frac{\partial\hat{\mm{x}}}{\partial t}{=}\dot{\mm{a}}$ using~(\ref{eq:Ansatz_DD-PINN_diff}).

\subsubsection{Initial-Condition Loss $\mc{L}_0$}
In order to train a PINN, it is essential to consider the governing initial/boundary conditions. 
For state-space modeling, this loss is defined for the initial condition $\mm{x}_0$, which must match the network output $\mm{f}_{\ind{PINN}\mm{\delta}}(0,\mm{x}_0,\mm{u}_0,\mm{\delta})$. 
The loss is also computed on the sampled collocation points with time $t$ set to zero.
For the DD-PINN, this loss is not necessary since $\mc{L}_0{\equiv}0$ due to the choice of $\mm{a}$.

\subsubsection{Data Loss $\mc{L}_\ind{d}$}
This optional loss term is determined with existing real measurement data in the training domain $\mm{\delta}_\ind{t}$.
Similar to black-box learning, the prediction $\hat{\mm{x}}(T_\ind{s}){=}\mm{f}_{\ind{PINN}\mm{\delta}}(T_\ind{s},\mm{x}_0,\mm{u}_0,\mm{\delta}_\ind{t})$ is compared to the measured ground truth $\mm{x}(T_\ind{s})$ using the specified error metric.

\highlightred{To conclude, the PINN is trained for large timesteps similarly to the RNN.
	Further, synthetic collocation points in arbitrary domains are sampled during training, for which physical system knowledge is utilized.
	By means of a weighted, multi-objective loss, it is possible to integrate real system data.
	Note that there is only good generalization capability for quantities that are considered with the domain input $\mm{\delta}$.
	For example, if significantly stiffer bellows are used, the accuracy of the PINN (as for the FP model and RNN) will decrease.
	The disassembly of one soft actuator after training to an ASR with $n{=}4$ joints would also result in less accuracy for the PINN (as for the RNN).
	In contrast, the FP approach has the advantage that the dynamics model for $n{=}4$ can be symbolically generated and used with the identified parameters.
	The PINN would have to be retrained with the updated FP model.}

\subsection{Implementation}\label{pinn_impl}
\begin{figure}[t]
	\removelatexerror
	\vspace{2mm}
	\begin{algorithm}[H]
		\caption{Training of PINC or DD-PINN}\label{alg:pinn}
		{\small 
			\SetKwInOut{Input}{In}
			\SetKwInOut{Output}{Out}
			\Input{$n_\ind{e},n_\ind{s},n_\ind{p},n_0,n_\ind{b},n_\ind{a},n_{\lambda},\lambda_\ind{0},\lambda_\ind{min},\mm{X}_\ind{d},\mm{Y}_\ind{d}$}
			\Output{$\mm{w}$}
			$\mm{\eta},\lambda,\mc{L}_0\gets\{[1,1,1]\transpose,\lambda_\ind{0},0\}$\;
			\ForEach{$epoch \in [0,\dots,n_\mathrm{e}-1]$}{
				\If{$epoch\bmod n_\ind{s}=0$}{\label{1st:line:sampling}
					$\mm{X}_\ind{p}\gets$ Sample $n_\ind{p}$ points\;
					\uIf{$n_\ind{a}=0$}{
						\tcp{PINC}
						$\mm{X}_\ind{0}\gets$ Take $n_\ind{0}$ points from $\mm{X}_\ind{p}$ and set $t=0$\;
						$\mm{Y}_\ind{0}\gets\mm{x}_0$ from all samples $\mm{X}_\ind{0}$, cf. (\ref{eq:b})\;
						$\mm{X},\mm{Y}\gets$ Shuffle $[\mm{X}_\ind{d},\mm{Y}_\ind{d}],[\mm{X}_\ind{p},\mm{0}],[\mm{X}_\ind{0},\mm{Y}_\ind{0}]$\;\label{lst:line:Yp}
					}
					\Else{
						\tcp{DD-PINN}
						$\mm{X},\mm{Y}\gets$ Shuffle $[\mm{X}_\ind{d},\mm{Y}_\ind{d}],[\mm{X}_\ind{p},\mm{0}]$\;\label{2st:line:Yp}						
						}
					$\mm{D}'_\ind{t},\mm{D}'_\ind{v}\gets$ Split $\mm{X},\mm{Y}$ in batches of size $n_\ind{b}$\;
				}\label{2st:line:sampling}
				$\bar{\mc{L}}_\ind{v},\bar{\mm{\mc{L}}}_\ind{t}\gets\{0,\mm{0}\}$\;
				\ForEach{$\mm{D}'\in[\mm{D}'_\ind{t},\mm{D}'_\ind{v}]$}{
					\ForEach{$\mm{B}\in\mm{D}'$ with $b$ batches}{
						$\mc{L}_\ind{d}\gets\mathrm{MSE}(\mm{Y}_{\ind{d}\mm{B}},\hat{\mm{Y}}_{\ind{d}\mm{B}})$\;\label{lst:line:Ld}
						$\mc{L}_\ind{p}\gets\mathrm{MSE}(\mm{F}(\hat{\mm{Y}}_{\ind{p}\mm{B}},\mm{X}_{\ind{p}\mm{B}}),\frac{\partial}{\partial t}\hat{\mm{Y}}_{\ind{p}\mm{B}})$\;\label{lst:line:Lp}
						\If{$n_\ind{a}=0$}{
							$\mc{L}_\ind{0}\gets\mathrm{MSE}(\mm{Y}_{\ind{0}\mm{B}},\hat{\mm{Y}}_{\ind{0}\mm{B}})$\;\label{lst:line:Lic}
						}
						$\mc{L}\gets[\mc{L}_\ind{d},\mc{L}_\ind{p},\mc{L}_\ind{0}]\mm{\eta}$\;\label{lst:line:L_total}
						\uIf{$\mm{D}'=\mm{D}'_\ind{t}$}{
							$\bar{\mm{\mc{L}}}_\ind{t}\gets\bar{\mm{\mc{L}}}_\ind{t}+b^{-1}[\mc{L}_\ind{d},\mc{L}_\ind{p},\mc{L}_\ind{0}]\transpose$\;
							$\mm{w}\gets$ Optimize network weights with $\mc{L},\lambda$\;
						}
						\Else{$\bar{\mc{L}}_\ind{v}\gets\bar{\mc{L}}_\ind{v}+b^{-1}\mc{L}$\;}				
					\If{$epoch=0$ and last batch of $\mm{D}'_\ind{t}$}{
						\uIf{$n_\ind{a}=0$}{
							$\mm{\eta}\gets\max(\bar{\mm{\mc{L}}}_\ind{t})[1/\bar{\mc{L}}_\ind{td},1/\bar{\mc{L}}_\ind{tp},1/\bar{\mc{L}}_\ind{t0}]\transpose$\;}\label{lst:line:eta}
						\Else{
							$\mm{\eta}\gets\max(\bar{\mm{\mc{L}}}_\ind{t})[1/\bar{\mc{L}}_\ind{td},1/\bar{\mc{L}}_\ind{tp},1]\transpose$\;\label{2st:line:eta}
						}
					}}
			}
		\If{$epoch>n_\lambda$}{
		$\lambda\gets\mathrm{ReduceLROnPlateau}(\bar{\mc{L}}_\ind{v},n_{\lambda},\lambda_\ind{min})$\;\label{reducelr}}}	
		
		}
		
	\end{algorithm}
\end{figure} 

For a comprehensive overview, the PINN training described in the previous section is provided in Algorithm~\ref{alg:pinn} as pseudo-code.
We implemented the training in PyTorch~\cite{Paszke.2019}.
Necessary inputs for training a neural network with $n_\ind{n}$~neurons and $n_\ind{h}$~hidden layers are:
\begin{itemize}
	\item $n_\ind{e}$: Number of training epochs
	\item $n_\ind{s}$: The collocation points are resampled all $n_\ind{s}$ epochs for faster convergence and to prevent overfitting
	\item $n_\ind{p}$, $n_\ind{0}$: Number of collocation/initial-condition points
	\item $n_\ind{b}$: Number of collocation, initial-condition and data points in one batch
	\item $n_\ind{a}$: Number of ansatz functions of DD-PINN ($n_\ind{a}{=}0$ implies PINC training)
	\item $n_{\lambda},\,\lambda_\ind{0},\,\lambda_\ind{min}$: The initial learning rate $\lambda_\ind{0}$ is halved when there is no improvement of the mean validation loss $\bar{\mc{L}}_\ind{v}$ after $n_{\lambda}$ epochs until $\lambda_\ind{min}$ is reached (line~\ref{reducelr})
	\item $\mm{X}_\ind{d},\,\mm{Y}_\ind{d}$: Real measurement data from $\mm{D}_\ind{t}$ (optional)
\end{itemize} 

In general, $\mm{X}_\diamond$ describes the input data with $n_\diamond$ points for the respective loss terms. 
Associated with this, $\mm{Y}_\diamond$ is the ground-truth output, which is compared to the network predictions $\hat{\mm{Y}}_\diamond$. 
A mini-batch training of a given feedforward network is conducted.
After resampling at defined intervals, all data points are split into batches such that each batch consists of points for all losses~(lines \ref{1st:line:sampling}--\ref{2st:line:sampling}). 
There is no ground truth $\mm{x}(t)$ necessary when calculating the physics loss. 
This is indicated in line~\ref{lst:line:Yp} (or~\ref{2st:line:Yp}) with $\mm{Y}_\ind{p}{=}\mm{0}$. 
With a $70/30$ split, the training dataset\footnote{We use the $\mm{D}'_\diamond$ to denote the datasets during PINN training, which are not equal to the measured dataset in the training domain $\mm{D}_\ind{t}$.} $\mm{D}'_\ind{t}$ and the validation dataset $\mm{D}'_\ind{v}$ are formed.

For each batch $\mm{B}$ of the two datasets, the losses are calculated (lines~\ref{lst:line:Ld}--\ref{lst:line:L_total}). 
The physics loss is determined with modeled dynamics $\mm{f}_{\ind{FP}\mm{\delta}}(\hat{\mm{x}},\mm{u}_0,\mm{\delta})$, which is denoted with the function $\mm{F}(\cdot,\cdot)$ for the entire batch (line~\ref{lst:line:Lp}). 
Finally, the weights of the network~$\mm{w}$ are updated via the Adam optimizer using the current training loss and the current learning rate.

At the beginning, the weighting factors $\mm{\eta}{=}[\eta_\ind{d},\eta_\ind{p},\eta_\ind{0}]\transpose$ are initialized to ones.
After the first epoch, these are adjusted using the mean losses during all training batches $\bar{\mm{\mc{L}}}_\ind{t}$ (line~\ref{lst:line:eta} or~\ref{2st:line:eta}).  
With these scaling factors, the loss weighting is done during the remaining epochs.

\subsection{Domain Knowledge in Practice}\label{dom_know}
\highlightred{The PINN presentation ends} with a brief note regarding the main requirement of our approach: the knowledge of the current domain $\mm{\delta}$.
Such a requirement is also used as a basis in comparable work.
For example, various payloads are known to the controller a priori in~\cite{Centurelli.2022}.
The model in~\cite{Wang.2024c} also receives the load as input.
For easily measurable quantities, these can be determined online in the application, e.g., measurement of the base inclination using an accelerometer.
Alternatively, unmeasurable quantities could be estimated online.
To this end, the proposed PINN architecture with the domain as input is ideally suited due to the high prediction speed.

\subsection{Real-Time Application: Nonlinear MPC with PINNs}\label{mpc}
As one possible PINN application, a nonlinear model predictive control of the soft robot is realized. 
Since there is a high demand for prediction speed, the benefits of fast and accurate PINNs can be illustrated here.
Besides considerably improved generalizability of PINNs compared to RNNs, the simple structure of a feedforward NN is another advantage of PINNs.
MPC with RNNs requires the correct initialization of the non-measurable hidden states, which complicates their use in MPC~\cite{Schafke.2024}.

\begin{figure}[tbp]
		\centering
		\resizebox{\linewidth}{!}{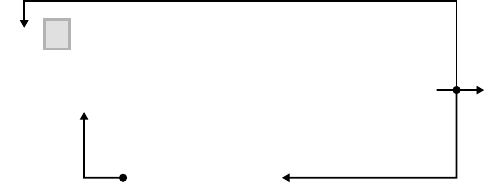}
		\caption{\highlightred{Control architecture: The nonlinear MPC uses the PINN as the dynamics model. Instead, a PI controller~\cite{Habich.2024} is used in this work for comparison. Changing between the two strategies is visualized with the switch.}} \label{fig:control_architecture}
	\end{figure}

The proposed control architecture is illustrated in Fig.~\ref{fig:control_architecture}.
MPC uses the discrete model of the system to predict the behavior for a prediction horizon of $m$ time steps.
The MPC solver searches for an input trajectory to minimize a user-defined cost function within this prediction horizon.
Only the first input signal $\mm{u}^*_0$ of this optimized input trajectory $[\mm{u}^*_\ind{0},\dots,\mm{u}^*_{m{-}1}]{\in}\R^{2n{\times}m}$ is applied to the real system, and the optimization is solved again with current measured states \highlightred{$\mm{x}$} and new desired \highlightred{states $[\mm{x}_\ind{d1},\dots,\mm{x}_{\ind{d}m}]{\in}\R^{2n{\times}m}$ with $\mm{x}_{\ind{d}k}{=}[\mm{q}_{\ind{d}k}\transpose,\dot{\mm{q}}_{\ind{d}k}\transpose]\transpose$} for $m$ future time steps.
\highlightred{To improve the speed of the solver, a control horizon of length one is selected.
	This effectively means that the input is kept constant throughout the prediction horizon so that $\mm{u}_0{=}\mm{u}_{k}{\forall} k$ applies.
	Therefore, only one input $\mm{u}_0$ must be optimized.} 
The optimization problem is formulated as

\highlightred{\begin{multline}\label{eq:optimization_problem}
	\minimize_{\mm{u}_0} \enspace  \sum_{k=1}^{m-1}  \big(\lVert \mm{q}_{\ind{d}k}-\hat{\mm{q}}_k \rVert_{\mm{Q}_{\ind{s}q}}^{2}+\lVert \dot{\mm{q}}_{\ind{d}k}-\hat{\dot{\mm{q}}}_k \rVert_{\mm{Q}_{\ind{s}\dot{q}}}^{2}\big)+\\
	\lVert \mm{q}_{\ind{d}m}-\hat{\mm{q}}_{m} \rVert_{\mm{Q}_{\ind{t}q}}^{2} + \lVert \dot{\mm{q}}_{\ind{d}m}-\hat{\dot{\mm{q}}}_{m} \rVert_{\mm{Q}_{\ind{t}\dot{q}}}^{2}+\sum_{k=0}^{m-1} \lVert \mm{u}_{k} \rVert_{\mm{R}_\ind{s}}^{2}
\end{multline}}
subject to\footnote{For the sake of clarity, we use the index notation $\hat{\mm{x}}_k{=}\hat{\mm{x}}(kT_\ind{s})$.} $\hat{\mm{x}}_{k+1}{=}\mm{f}_{\ind{PINN}\mm{\delta}}(T_\ind{s},\hat{\mm{x}}_k,\mm{u}_k,\mm{\delta})$, $\mm{u}_{\ind{min}} {\leq}  \mm{u}_{k} {\leq} \mm{u}_{\ind{max}}$, and $\hat{\mm{x}}_0$ obtained from measurements for each MPC cycle.

Since the PINN uses scaled inputs/outputs, the entire optimization problem is formulated with scaled (unitless) quantities.
\highlightred{The diagonal weighting matrices for the stage costs of the positions $\mm{Q}_{\ind{s}q}$, velocities $\mm{Q}_{\ind{s}\dot{q}}$ and inputs $\mm{R}_\ind{s}$, and for the terminal cost of the positions $\mm{Q}_{\ind{t}q}$ and velocities $\mm{Q}_{\ind{t}\dot{q}}$ consist of constant diagonal entries ${Q}_{\ind{s}q}$, ${Q}_{\ind{s}\dot{q}}$, ${R}_\ind{s}$, ${Q}_{\ind{t}q}$, and ${Q}_{\ind{t}\dot{q}}$.}
The input limits are $\mm{u}_{\text{min}}$ and $\mm{u}_{\text{max}}$, which can be obtained by min-max scaling of the system-specific limits with a pressure range of $0{-}\SI{0.7}{\bar}$.
\highlightred{The defined pressure limits inherently eliminate any risk of system damage. 
Consequently, imposing state constraints (on position $\mm{q}$ or velocity $\dot{\mm{q}}$) would offer no practical benefit in this application. 
Instead, it would increase the computational burden of the online optimization process.
We, therefore, refrain from imposing state constraints in the MPC.}
The nonlinear MPC problem was implemented with CasADi~\cite{Andersson.2019} using the interior-point method.

\highlightred{For a comparison, we use the PI controller from~\cite{Habich.2024} with weighting matrices ($\mm{Q}_\ind{P}$ and $\mm{Q}_\ind{I}$), anti-windup and output saturation.
This outputs the desired pressure differences $\Delta\mm{p}_\ind{PI}{\in}\R^{n}$ of all bellows pairs, which is mapped to each antagonist and agonist via 
\begin{equation}
	\begin{split}
		\mm{u}_\ind{PI}&{=}\mm{g}_\ind{PI}(\Delta \mm{p}_\ind{PI})\\
		& {=} \bar{\mm{p}}+[\Delta p_\ind{PI1},{-}\Delta p_\ind{PI1},\dots,\Delta p_{\ind{PI}n},{-}\Delta p_{\ind{PI}n}]\transpose/2
	\end{split}
\end{equation}
with $\bar{\mm{p}}=\frac{p_\ind{max}}{2}[1,\dots,1]\transpose$.}

\section{Experiments}\label{experiments}
After describing the pneumatic test bench~(Sec.~\ref{test_bench}), the identification and HPO results are presented~(Sec.~\ref{ident_learning_res}).
All these optimized (first-principles, hybrid, and black-box) models are then experimentally validated~(Sec.~\ref{performance}).
\highlightred{Afterwards, experiments on learning-based MPC are conducted with the proposed PINN~(Sec.~\ref{control_res}).
The section ends with an overall discussion of the experiments~(Sec.~\ref{results_discussion}).}
\subsection{Test Bench}\label{test_bench}

	\begin{figure}[tbp]
		\centering
		\resizebox{1\linewidth}{!}{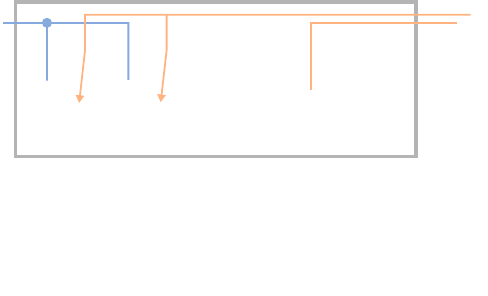}
		\caption{Test-bench architecture: Orange color represents communication elements and blue color represents pneumatic components.} \label{fig:test_bench_n_Act}
\end{figure}
The used test bench was presented in~\cite{Habich.2024} and is only briefly described below.
The main structure is illustrated in Fig.~\ref{fig:test_bench_n_Act}.
\subsubsection{Components}
The pneumatic system consists of a supply unit (pressure supply, shut-off valve, and filter regulator) and three-way proportional piezo valves (Festo VEAA-B-3-D2-F-V1-1R1, resolution: $\SI{5}{mbar}$) with integrated pressure control for each bellows.
Industrial compressors generate the compressed air centrally with negligible pressure fluctuations.
Each actuator is equipped with a Hall encoder (Megatron ETA25K, resolution: $0.09\degree$).
\subsubsection{Communication}
The EtherCAT protocol is used with the corresponding open-source tool EtherLab, which was modified with an external-mode patch and a shared-memory real-time interface\footnote{\highlightredfinal{\url{https://github.com/SchapplM/etherlab-examples}}} for Matlab/Simulink.
This enables reading in or setting several values (current pressures~$\boldsymbol{p}$, joint angles~$\boldsymbol{q}$, and desired pressures~$\boldsymbol{p}_{\ind{d}}$) with a cycle time of $\SI{1}{\milli\second}$ via the EtherCAT real-time bus with input and output terminals (Beckhoff EL3702 and EL4102).
To avoid damage, an output saturation of the desired pressures~$\boldsymbol{p}_{\ind{d}}$ is implemented with limits $0{-}p_\ind{max}$.

Data can be logged, and settings can be altered during runtime with the development computer~(Dev-PC, \SI{3.6}{\giga\hertz} Intel Core i7-12700K CPU with \SI{16}{\giga\byte} RAM) using Matlab/Simulink 2018b.
The compiled model is executed on the real-time computer~(RT-PC, \SI{4.7}{\giga\hertz} Intel Core i7-12700K CPU with \SI{16}{\giga\byte} RAM).
\subsubsection{MPC Implementation}
CasADi was integrated using its fast \CPP~API for real-time control instead of using Python.
The MPC was decoupled from the real-time system using a ROS~\highlightred{service}.
Once the MPC problem is solved, the modified $\mm{u}^*_0$ is used. 
Therefore, no fixed MPC frequency was specified, but rather, the fastest possible frequency was set automatically depending on the choice of MPC parameters.
The trained PINN is exported from PyTorch and manually recreated in CasADi.
Due to the use of the widespread ROS framework, this implementation\footref{foot:pinn} can also be beneficial for other applications.

\subsection{Model Identification/Learning}\label{ident_learning_res}
All approaches receive an identical dataset of $\SI{15}{\minute}$ for identification/learning in the training domain $\mm{\delta}_\ind{t}{=}[\SI{0}{\gram},\SI{0}{\degree}]\transpose$.
The data was logged at a frequency of $\SI{1}{\kilo\Hz}$ and then downsampled to $\SI{50}{\Hz}$.
Random pressure combinations limited to~$\SI{0.7}{\bar}$ with a linear transition of $\SI{1}{\second}$ were applied to each bellows, with each combination being held for $\SI{3}{\second}$.

\subsubsection{Parameter Identification}\label{ident_res}
With the chosen parameters ($\dot{q}_\ind{C}{=}\SI{1}{\degree/\second}$ and ${q}_\ind{bt}{=}\SI{10}{\degree}$), the datasets $\mm{D}_\ind{tI}$, $\mm{D}_\ind{tII}$, and $\mm{D}_\ind{tIII}$ comprise \SI{40}{\percent}, \SI{17}{\percent}, and \SI{43}{\percent} of all datapoints, respectively. 
Various combinations of $\dot{q}_\ind{C}$ and ${q}_\ind{bt}$ were tested, whereby the identified parameters only changed marginally.

\begin{figure}[tbp]
		\centering
		\resizebox{\linewidth}{!}{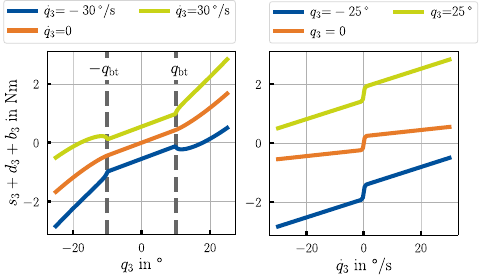}
		\caption{Stiffness, damping, and contact characteristics of joint $i{=}3$ with identified parameters for a simulative variation of (a)~position~$q_3$ with a kink at the threshold of the soft boundaries ${\pm} q_\ind{bt}$ and (b)~velocity~$\dot{q}_3$. Curves illustrate the identified parameters from joint $i{=}3$, and are qualitatively similar for the other joints.} \label{fig:ident_results}
	\end{figure}

The results are presented in Table~\ref{tab:ident_params}.
For both the stiffness~$\mm{k}_\ind{s}$ and the friction parameters~$\mm{k}_\ind{v}$ and~$\mm{k}_\ind{C}$, a trend can be seen that these increase slightly towards the distal end.
As already mentioned, this is attributed to the routing of the cables and tubes in the robot body.
The parameters shown remain unchanged after this offline identification and serve as the training foundation for the PINNs.
To illustrate the effects of the various parameters, Fig.~\ref{fig:ident_results} shows the characteristics of different positions and velocities of an exemplary joint after identification.
\renewcommand{\arraystretch}{1.2}
\begin{table}[b]
	\centering
	\caption{Identified parameters $\mm{k}$ of (\ref{eq:dyn})}
	\begin{tabular}{|c|c|c|c|c|c|c|}
		\hline
		param.&unit& $i{=}1$&$i{=}2$&$i{=}3$&$i{=}4$&$i{=}5$ \\
		\hline
		$k_{\ind{s}i}$ &\SI{}{\newton\meter/\degree}&$0.035$&$0.034$&$0.043$&$0.035$&$0.041$\\
		\hline
		$k_{\ind{v}i}$ &\SI{}{\newton\meter\second/\degree}&$0.008$&$0.008$&$0.010$&$0.011$&$0.011$\\
		\hline
		$k_{\ind{C}i}$ &\SI{}{\newton\meter}&$0.171$&$0.214$&$0.233$&$0.204$&$0.232$\\
		\hline
		$k_\ind{bs}$ &\SI{}{\newton\meter/(\degree)^{3/2}}&\multicolumn{5}{c|}{$0.010$}\\
		\hline
		$k_\ind{bd}$ &\SI{}{\newton\meter\second/(\degree)^{3/2}}&\multicolumn{5}{c|}{$0.005$}\\
		\hline
	\end{tabular}
	\label{tab:ident_params}
\end{table}

\subsubsection{PINC vs. DD-PINN}\label{training_res}
To first get an impression of the training, the proposed PINC and DD-PINN with domain input $\mm{\delta}$ were each trained with identical parameters ($n_\ind{s}{=}250$, $n_\ind{p}{=}\SI{100000}{}$, $n_\ind{b}{=}512$, $n_\lambda{=}50$, $\lambda_\ind{0}{=}\SI{5e{-4}}{}$, $\lambda_\ind{min}{=}\SI{5e{-5}}{}$).
Both trainings were conducted on one core with \SI{8}{\giga\byte} RAM of a computing cluster (\SI{2.6}{\giga\hertz} Intel Xeon Gold 6442Y CPU).
The PINC receives $n_\ind{0}{=}0.2n_\ind{p}$ initial-condition points.
In contrast, the DD-PINN does not require these points, as $\mc{L}_\ind{v0}{\equiv}\mc{L}_\ind{t0}{\equiv}0$ applies due to the structural property of the used ansatz function (\ref{eq:Ansatz_DD-PINN}) with $n_\ind{a}{=}50$.
A simple feedforward NN with $n_\ind{n}{=}100$ neurons and $n_\ind{h}{=}2$ hidden layers is trained.
The user-specified boundaries are set to 
\begin{equation} \label{boundaries}
	\begin{split}
		\mc{T}&=[0,\kappa T_\ind{s}], \\
		\mc{X}_0&=[-\kappa q_\ind{max},\kappa q_\ind{max}]^{n}\times[-\kappa \dot{q}_\ind{max},\kappa \dot{q}_\ind{max}]^{n},\\
		\mc{U}_0&=[0,\kappa p_\ind{max}]^{2n}\,\ind{and}\\
		\mc{D}&=[0,\kappa m_\ind{emax}]{\times}[0,\kappa \beta_{\ind{gmax}}]
	\end{split}
\end{equation}
with $q_\ind{max}{=}\SI{25}{\degree}$, $\dot{q}_\ind{max}{=}\SI{30}{\degree/\second}$, $p_\ind{max}{=}\SI{0.7}{\bar}$, $m_\ind{emax}{=}\SI{200}{\gram}$ and $\beta_{\ind{gmax}}{=}\SI{90}{\degree}$.
The limits are system-specific, and the factor $\kappa{=}1.25$ is used to train for slightly larger ranges, also done in~\cite{Nicodemus.2022}.
We chose the sample time $T_\ind{s}{=}\SI{20}{\milli\second}$, which is further discussed in Sec.~\ref{HPO_res}.

Note that no real data points ($n_\ind{d}{\equiv}0$) are used for all PINNs in this work, similar to~\cite{Antonelo.2024,Krauss.2024}. 
In our case, adding data points has no advantage due to the high accuracy of the first-principles model (cf. Sec.~\ref{pred_acc}), which is \textit{already identified with the real data} from the training domain.
Moreover, this would slow down and complicate the training even further due to the additional loss term.

\begin{figure}[t]
	\vspace{-0.5mm}
	\centering{\includegraphics[width=\linewidth]{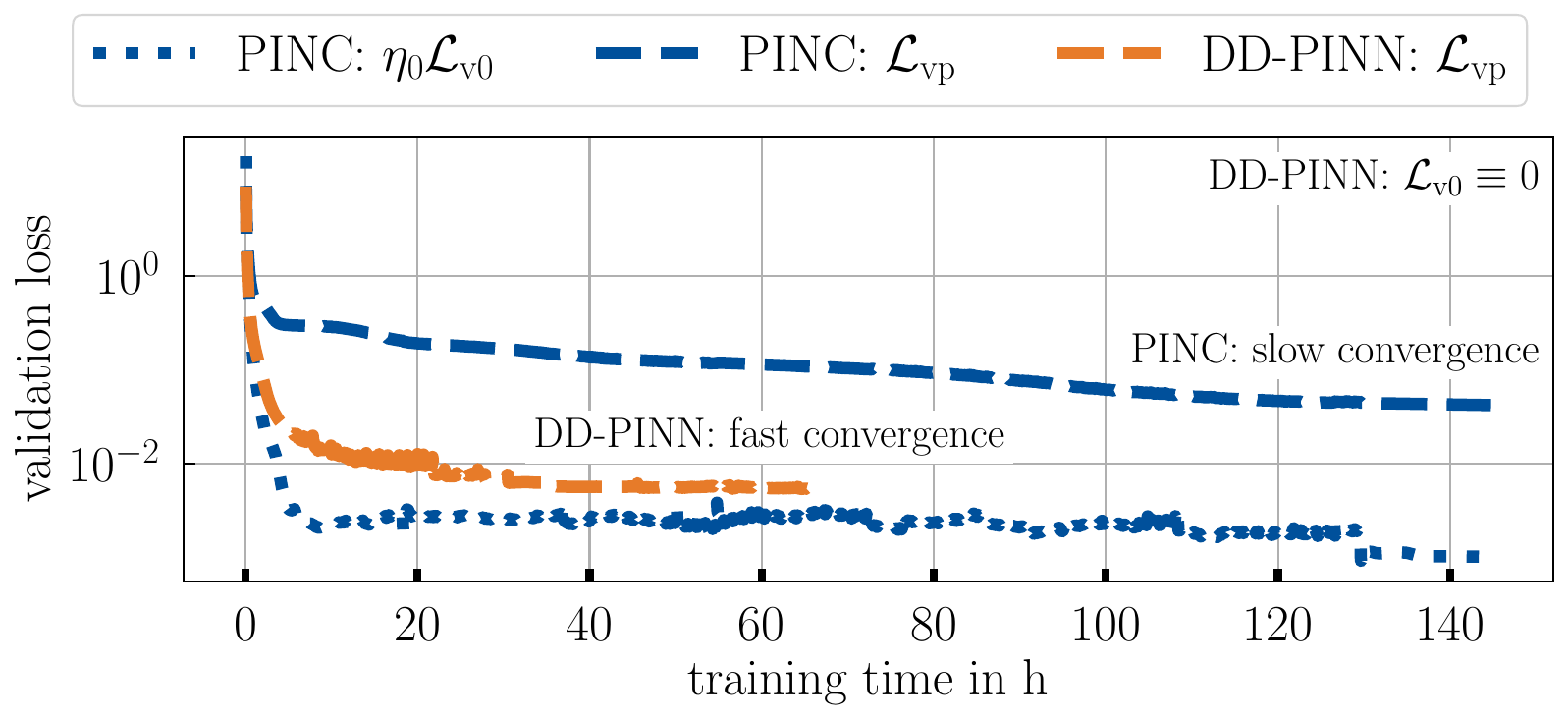}}
	\caption{Training convergence of two PINN architectures (PINC and DD-PINN), which are adapted for our system and extended to enable additional domain input $\mm{\delta}$. The validation physics losses $\mc{L}_\ind{vp}$ and validation initial-condition loss $\mc{L}_\ind{v0}$ are plotted over training time. Both networks are trained on the same hardware with identical parameters. For the PINC, two competing loss terms must be minimized, whereby the initial-condition loss is weighted with $\eta_0$ ($\eta_\ind{p}{=}1$). The DD-PINN has a considerably improved convergence, which is indicated by the \textit{lower physics loss}. Also, the initial-condition loss from the DD-PINN ($\mc{L}_\ind{v0}{\equiv}0$) is always less than PINC's $\mc{L}_\ind{v0}$. \highlightred{In order to also use the PINC as a reference model in this work, it was trained for a considerably longer duration of $\SI{144}{\hour}$ (six days). Even after this time, convergence is not yet complete.}}
	\label{fig:train_time}
\end{figure}

The training progress for both networks is illustrated in Fig.~\ref{fig:train_time}.
It can be seen that the DD-PINN converges much faster for two main reasons.
First, no initial-condition loss needs to be minimized. 
Second, $\frac{\partial\hat{\mm{x}}}{\partial t}$ is calculated in closed form for all $n_\ind{p}$~collocation points so that no computationally expensive automatic differentiation is required. 
The latter considerably influences the average duration of one training epoch $\bar{T}_\ind{e}$, allowing it to be reduced by \highlightred{$\SI{38.8}{\percent}$} (PINC: \highlightred{$\bar{T}_\ind{e}{=}\SI{260.9}{\second}$}, DD-PINN: $\bar{T}_\ind{e}{=}\SI{159.7}{\second}$).
Thus, the DD-PINN was trained for \SI{1500}{epochs}, while the PINC could only be trained for \highlightred{\SI{919}{epochs}} within the recorded training time of approx. \SI{67}{\hour} \highlightred{(\SI{2.8}{days})}.
A systematic HPO can be enabled in a reasonable time due to this considerably faster convergence \highlightred{of the DD-PINN.}
\highlightred{This is presented in the following.}
\subsubsection{Hyperparameter Optimization}\label{HPO_res}
Compared to RNNs, the training of PINNs is very time-consuming and requires considerable resources of computing hardware. 
We suspect that this is the main reason why systematic hyperparameter optimizations of PINNs are relatively rare in the state of the art.
This is in line with a recent overview~\cite{Nghiem.2023}, which declares such HPO as an open challenge.
However, since many hyperparameters significantly influence the network performance, we perform an HPO.
For this purpose, the computing cluster from the previous section was used, which allows several \highlightred{DD-}PINNs to be trained in parallel.

Before an HPO can be carried out, the most crucial training parameters must first be defined.
For the DD-PINN, these are $n_\ind{n}$, $n_\ind{h}$, $\lambda_0$ and $n_\ind{a}$.
All other parameters are taken from the previous section.
At this point, it must be noted that $T_\ind{s}$ is also a crucial parameter for the PINN performance.
For large sampling times $T_\ind{s}$, the prediction speed is improved, but the accuracy degrades.
The opposite is true for a small $T_\ind{s}$.
This tradeoff was already highlighted in~\cite{Zeipel.2024} for the PINC and applies analogously to the DD-PINN.
Since the physics loss also shows a tendency to decrease as $T_\ind{s}$ decreases, integration into an HPO with ASHA is not recommended. 
It would always tend towards a very small $T_\ind{s}$ without considering the prediction speed.
This parameter was therefore tuned iteratively, and a sample time $T_\ind{s}{=}\SI{20}{\milli\second}$ proved to be suitable.
The later use in the MPC was also taken into account here, as the sample times must not be too small.
Otherwise, there would be problems similar to those with the FP~model~(cf. Sec.~\ref{first_princ}).

During the HPO, 35 networks are trained simultaneously, with a total of 100 trials being examined.
As PINN training converges slowly, ASHA's grace period is set to 500 epochs with a reduction factor of 2.
To keep the computation time reasonable, only $n_\ind{e}{=}\SI{1000}{}$ epochs are trained during the HPO.
The best trial can be further trained afterward until $n_\ind{e}{=}\SI{1500}{}$.

	\begin{figure}[tbp]
		\centering
		\resizebox{\linewidth}{!}{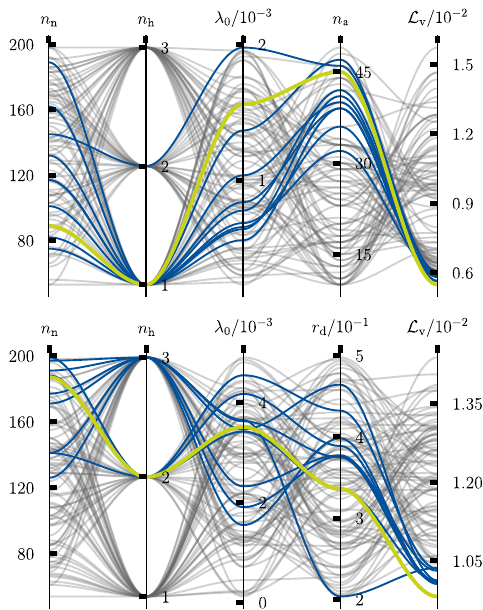}
		\caption{Results of hyperparameter optimization with $100$ trials each: Poorly performing trials are shown in gray, the best ten in blue, and the best one in green. The validation loss $\mc{L}_\ind{v}$ is considerably reduced by systematically tuning the networks' hyperparameters. (a)~The optimization of the most important \highlightred{DD-}PINN parameters ($n_\mathrm{n}$, $n_\mathrm{h}$, $\lambda_0$ and $n_\mathrm{a}$) took approx. six days due to the long training times on the specified cluster hardware. (b)~The optimization of the most important RNN parameters ($n_\mathrm{n}$, $n_\mathrm{h}$, $\lambda_0$ and $r_\mathrm{d}$) took ${<}\SI{14}{\hour}$ on the specified cluster hardware with minor performance.} \label{fig:HPO_results}
	\end{figure}

The results are visualized in Fig.~\ref{fig:HPO_results}(a).
It can be seen that the systematic HPO can considerably minimize the validation loss.
Note that with such an HPO, the limits of the hyperparameters must be adjusted iteratively so that no optimal parameters lie at their boundaries.
Otherwise, the true optimum of the parameter could be outside the limits.
This iterative process, in combination with the HPO duration of approx. six days, therefore, requires a substantial computation and time effort.

The HPO of the RNN is performed similarly, and the results are shown in Fig.~\ref{fig:HPO_results}(b).
Since RNNs converge much faster, $n_\ind{e}{=}300$, $n_\ind{\lambda}{=}10$ are chosen, and ASHA's grace period is set to 100 epochs.
The \highlightred{DD-}PINN-specific parameter $n_\ind{a}$ is replaced by the dropout rate $r_\ind{d}$ as an important training parameter of the RNN.
Although significantly inferior cluster hardware (\SI{2.3}{\giga\hertz} Intel Cascade Lake Xeon Gold 6230N CPU) was used, only ${<}\SI{14}{\hour}$ is required for the HPO of the RNNs.

\subsection{\highlightred{First Principles vs. PINNs vs. Black Box}}\label{performance}
The following experiments underline our \textit{first and second claims} regarding prediction speed and generalizability, which are illustrated in Fig.~\ref{fig:cover}(a).
As the model base, we use the identified first-principles model, as well as the \highlightred{DD-}PINN and RNN (both hyperparameter-optimized) from the previous section.
\highlightred{As already mentioned, the PINC training is too time-consuming for an HPO.
	In order to also have a PINC reference model, we trained the network even further.
	This is also shown in Fig.~\ref{fig:train_time}.
	In total, the PINC training of the single configuration took $\SI{144}{\hour}$ (six days).}

\subsubsection{Prediction Speed}\label{pred_speed}
For a fair comparison of the prediction times, the physics-driven state-space model~(\ref{eq:state-space}) and the forward passes of the RNN~(\ref{eq:RNN}) and \highlightred{PINNs}~(\ref{eq:pinn_forward}) were reimplemented and mex-compiled in Matlab R2023a.
A \SI{2.5}{\giga\hertz} Intel Core i5-10300H CPU with \SI{16}{\giga\byte} of RAM running Windows was used for evaluation.
The explicit Euler method and the Runge-Kutta fourth-order method (RK4) were used for the numerical integration of the physical model.
These are often used in the MPC as integrators due to the constant step size. 

The results for an exemplary \SI{100}{\second} trajectory of the input signals are listed in Table~\ref{tab:pred_speed}. 
The time for each prediction horizon of $T_\ind{s}{=}\SI{20}{\milli\second}$ was measured for all methods. 
The statistical data (mean, maximum and minimum), therefore, refer to $\SI{5000}{}$ time measurements, whereby a different number of function calls $n_\ind{calls}$ are required per measurement depending on the method.

Due to the stiff ODE~(\ref{eq:state-space}), fine step sizes (Euler: \SI{20}{\micro\second}, RK4: \SI{100}{\micro\second}) are necessary for robust forward simulation of the system.
In contrast, only \textit{one} function call of the networks' forward passes is required for a horizon of $T_\ind{s}$.
This demonstrates one key advantage of using model-learning approaches.
On average, the \highlightred{DD-PINN} is \highlightred{significantly} faster than numerical integration using explicit Euler or RK4 by a factor of \highlightred{467} and \highlightred{377}, respectively.
Due to the simpler network structure, the \highlightred{DD-}PINN is also slightly faster than the RNN, which is also presented in Table~\ref{tab:pred_speed}.
\highlightred{The PINC is slightly faster than the DD-PINN, as the PINC's forward pass is simpler without the ansatz function used.}

\begin{table}[b]
	\centering
	\caption{Computation time for predicting a horizon of $T_\ind{s}=\SI{20}{\milli\second}$. Statistics were computed for a \SI{100}{\second} input trajectory.}
	\begin{tabular}{|c|c|c|c|c|}
		\hline
		method&mean/\SI{}{\milli\second}& max/\SI{}{\milli\second} & min/\SI{}{\milli\second} & $n_\ind{calls}$ \\
		\hline
		\highlightred{DD-PINN} &$0.04$&$0.30$&$0.03$  &$\mathbf{1}$\\
		\hline
		\highlightred{PINC} &\highlightred{$\mathbf{0.02}$}&\highlightred{$\mathbf{0.25}$}&\highlightred{$\mathbf{0.01}$}  &\highlightred{$\mathbf{1}$}\\
		\hline
		RNN &$0.08$&$1.46$&$0.05$&$\mathbf{1}$\\
		\hline
		Euler &$18.67$&$35.74$&$17.37$&$1000$\\
		\hline
		RK4 &$15.07$&$27.96$&$13.98$&$200$\\
		\hline
	\end{tabular}
	\label{tab:pred_speed}
\end{table}

With regard to a real-time nonlinear MPC, the results show that the numerical integration of the physics-driven state-space model is unsuitable.
The required fine temporal discretization of the physics model and, thus, the high number of function calls is problematic.
The number of discrete time steps $m$ in the prediction horizon of the MPC influences the solution times considerably and is, therefore, usually set to a few time steps, e.g., $m{=}5$.
This would lead to very short prediction horizons (Euler: \SI{100}{\micro\second}, RK4: \SI{500}{\micro\second}), which results in difficulties such as aggressive MPC actions.
Online optimization cannot be solved in such a short time frame, as solver frequencies of $2{-}\SI{10}{\kilo\hertz}$ would be required.
Even if we ignore this aspect, \highlightred{and use constant input signals for long segments of the prediction horizon --- the long prediction times prevent the realization of fast real-time applications.}
\subsubsection{Generalizability}\label{pred_acc}
For all test datasets, random pressure combinations limited to $\SI{0.7}{\bar}$ with a linear transition of $\SI{1}{\second}$ were applied to each bellows.
In contrast to the training data, each combination is held for only $\SI{1}{\second}$, which leads to more dynamic motions.
\begin{figure}[t]
	\vspace{2.5mm}
	\centering{\includegraphics[width=\linewidth]{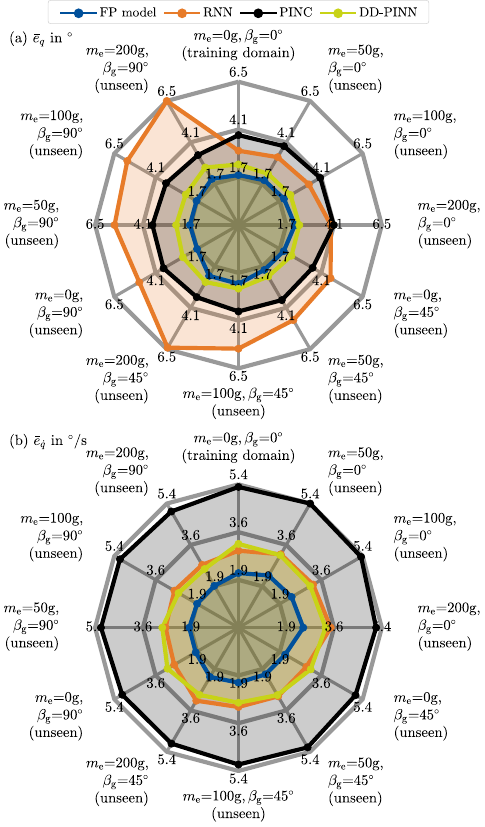}}
	\caption{Accuracy of FP model, RNN, and \highlightred{PINNs (PINC and DD-PINN)} on \highlightred{the training domain $\mm{\delta}_\ind{t}{=}[\SI{0}{\gram},\SI{0}{\degree}]\transpose$ and eleven unseen test domains $\mm{\delta}{\neq}\mm{\delta}_\ind{t}$ with \SI{10}{\minute} duration of each test dataset}. 
		The ODE of the FP model is numerically integrated with an advanced implicit solver to achieve results of the highest possible accuracy. 
		(a)~Mean absolute error (MAE) $\bar{e}_q$ between true and predicted positions averaged over all $n$ joints. 
		The \highlightred{DD-PINN} is only marginally inferior to the FP model, whereas the RNN has large errors due to poor generalization.
		\highlightred{The PINC shows better generalizability than the RNN. 
			However, the PINC is less accurate than the DD-PINN, although it has been trained considerably longer.}
		(b)~MAE $\bar{e}_{\dot{q}}$ between true and predicted velocities averaged over all $n$ joints. 
		\highlightred{DD-PINN} and RNN are only marginally less accurate than the FP model, whereby the error of the RNN increases with higher mass $m_\ind{e}$ due to poor generalization. \highlightred{Again, the PINC does not show satisfactory accuracy.}}
	\label{fig:spider_plot}
\end{figure}

Many publications only investigate a generalization for changed input trajectories, for which this would already be sufficient.
However, we also want to investigate the generalizability to changes in system dynamics after training-data acquisition.
For this purpose, test datasets with a duration of $\SI{10}{\minute}$ for twelve different domains $\mm{\delta}{\in}[\SI{0}{\gram},\SI{50}{\gram},\SI{100}{\gram},\SI{200}{\gram}]{\times}[\SI{0}{\degree},\SI{45}{\degree},\SI{90}{\degree}]$ were recorded. 
Note that an additional mass of $m_\ind{e}{=}\SI{200}{\gram}$ is approx. \SI{20}{\percent} of the total mass of the robot, and therefore, represents a considerable system change.
In order to also investigate possible bellows changes over time, the test datasets were recorded \textit{more than two months later} than the training dataset.
The robot continued to operate during this time, which required bellows to be replaced due to cracks.
We believe this is important because the identified parameters listed in Table~\ref{tab:ident_params} are fixed and serve as the basis for training the PINN.

The results for all test datasets are visualized in Fig.~\ref{fig:spider_plot}.
To have a first-principles baseline with the highest possible accuracy, an implicit Runge-Kutta method (Radau IIA of order five~\cite{Hairer.1996}) is used for numerical integration. 
This is more advanced than explicit Euler or RK4, and also has the advantage that no discrete step size has to be chosen/tuned.

Fig.~\ref{fig:spider_plot}(a) demonstrates the advantage of the PINN\highlightred{s} over the RNN.
The more changes are present after training, the larger the RNN error becomes: 
For $m_\ind{e}{=}\SI{200}{\gram}$ and $\beta_\ind{g}{\in}[\SI{45}{\degree},\SI{90}{\degree}]$, this position error is more than twice as large compared to the performance in the training domain~$\mm{\delta}_\ind{t}$.
In contrast, our proposed \highlightred{DD-}PINN consistently achieves an accuracy that is only slightly worse than the FP model, which has to be integrated with a substantial time effort.
Even in the training domain, the \highlightred{DD-}PINN achieves a lower position error than the RNN.
\highlightred{Also, the DD-PINN is considerably more accurate than the PINC. 
	Although the PINC was trained for six days, it has obviously not yet finally converged.}
\begin{figure}[t]
	\vspace{2.5mm}
	\centering{\includegraphics[width=\linewidth]{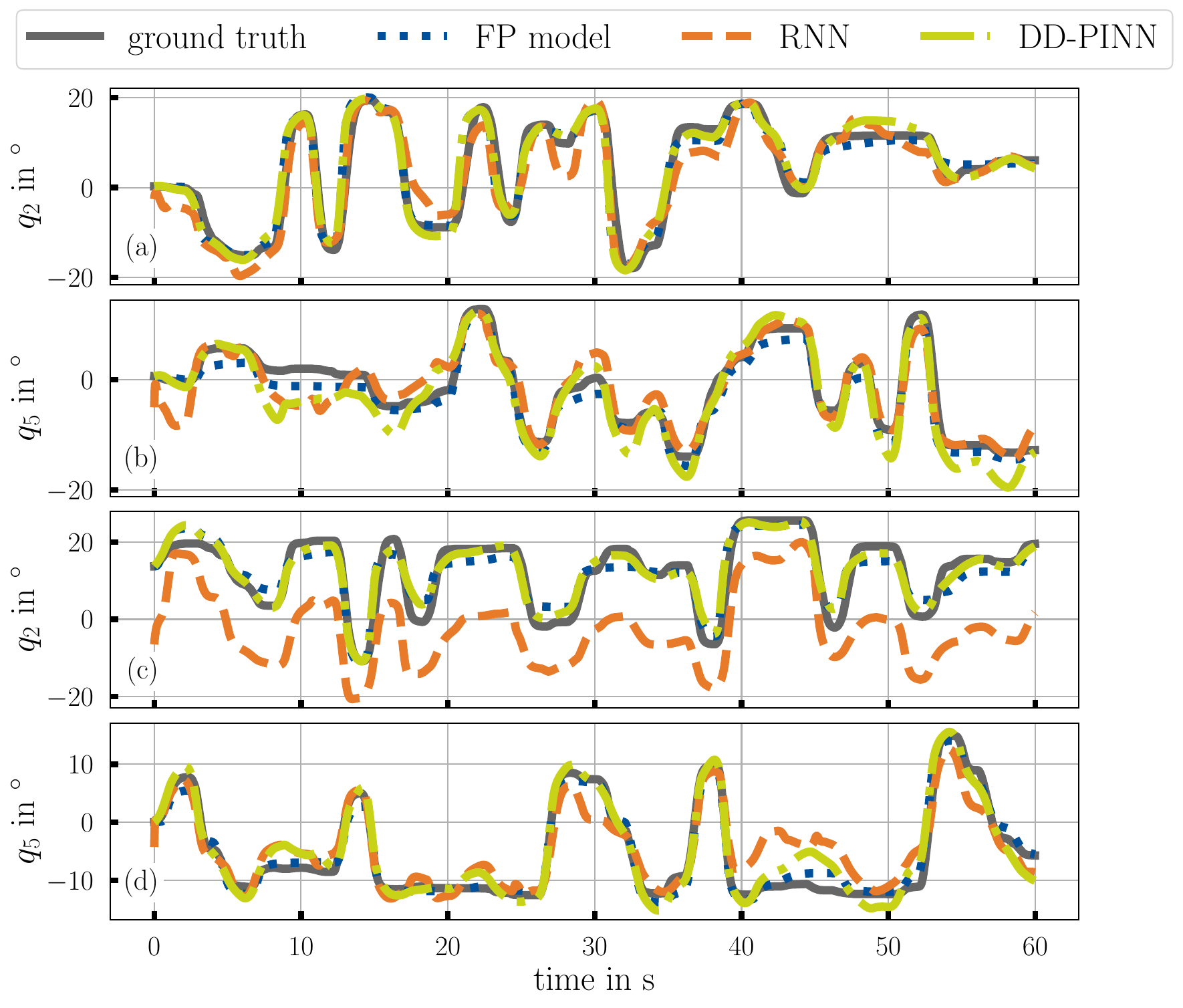}}
	\caption{Prediction results for $q_2$ and $q_5$ on test datasets in (a)--(b)~training domain $\mm{\delta}{=}\mm{\delta}_\ind{t}{=}[\SI{0}{\gram},\SI{0}{\degree}]\transpose$ and (c)--(d)~unseen test domain $\mm{\delta}{=}[\SI{100}{\gram},\SI{45}{\degree}]\transpose$. Within the unseen test domain, the dynamics of joint $i{=}2$ are mainly changed due to gravity. This is taken into account by the FP model and the \highlightred{DD-PINN} due to the exploited physical knowledge. During training, real-world data in this test domain was not available. Therefore, large deviations occur with the RNN due to poor generalization.}
	\label{fig:traj_test}
\end{figure}

There are \highlightred{somewhat different} results on the velocity level, which are shown in Fig.~\ref{fig:spider_plot}(b).
Since the base orientation~$\beta_\ind{g}$ only has a minor influence on the velocity, the poor generalizability of the RNN is less present.
However, the mass $m_\ind{e}$ influences the velocity.
Therefore, at higher inertia, a higher prediction error of the RNN can be seen.

\highlightred{The PINC clearly shows the worst accuracy at the velocity level, which can be explained by the suboptimal convergence.
	A systematic HPO of the PINC could find better network configurations, but is not realistic due to the excessive training time.
	Since the DD-PINN has a comparable prediction speed with considerably better accuracy compared to the PINC, it is used for the rest of this work.}

To elaborate further, Fig.~\ref{fig:traj_test} shows prediction results for a short excerpt of tests in two different domains. 
First, the RNN often shows poor accuracy at the beginning because the unmeasurable hidden states have to be initialized first. 
Moreover, it can be seen that the prediction of $q_2$, in particular, is worse for new domains, as this is most affected by the system change.
Such disadvantages do not exist with the \highlightred{DD-}PINN.

\subsection{Control Results}\label{control_res}
In final control experiments with the soft robot, the \textit{third claim} is confirmed.
It should first be mentioned that all controller gains were manually tuned once in the training domain and then remained unchanged.
The PI gains are set to $\mm{Q}_\ind{P}{=}\ind{diag}(40,60,110,110,100)\SI{}{\milli\bar/\degree}$ and $\mm{Q}_\ind{I}{=}\ind{diag}(40,40,80,60,110)\SI{}{\milli\bar/(\degree\second)}$.
For the nonlinear MPC problem, a prediction horizon with \highlightred{$m{=}3$} steps was chosen and the unitless gains were tuned to \highlightred{${Q}_{\ind{s}q}{=}{Q}_{\ind{t}q}{=}0.7$ and ${Q}_{\ind{s}\dot{q}}{=}{Q}_{\ind{t}\dot{q}}{=}{R}_\ind{s}{=}0.01$}.
Automated tuning could further improve the control performance.

\begin{table}[t]
	\centering
	\caption{Tracking errors and NMPC frequency for control experiments in different domains $\mm{\delta}{=}[m_\ind{e},\beta_\ind{g}]\transpose$ \\with $\SI{40}{\second}$ duration each.}
	\highlightred{
	\begin{tabular}{|c|c|c|c|c|}
		\hline
		$\beta_\ind{g}/\SI{}{\degree}$ &$m_\ind{e}/\SI{}{\gram}$&controller & $\bar{e}_q/\SI{}{\degree}$ & $f_\ind{NMPC}/\SI{}{\hertz}$ \\
		\hline
		\multirow{4}{*}{$0$} 	& \multirow{2}{*}{$0$} 		& DD-PINN+NMPC &$\mathbf{2.60}$ & $46.4$\\ \cline{3-5} 
		&                   & PI~\cite{Habich.2024}       &$3.04$ & --\\ \cline{2-5} 
		& \multirow{2}{*}{$200$} 	& DD-PINN+NMPC &$\mathbf{2.29}$ & $46.8$\\ \cline{3-5} 
		&                  			& PI~\cite{Habich.2024}       &$3.23$ & --\\ \hline
		\multirow{4}{*}{$45$} 	& \multirow{2}{*}{$0$} 		& DD-PINN+NMPC &$\mathbf{2.88}$ & $47.6$\\ \cline{3-5} 
		&                   		& PI~\cite{Habich.2024}      &$3.20$ & --\\ \cline{2-5} 
		& \multirow{2}{*}{$200$} 	& DD-PINN+NMPC &$\mathbf{2.80}$ & $46.3$\\ \cline{3-5} 
		&                   		& PI~\cite{Habich.2024}       &$3.22$ & --\\ \hline
		\multirow{4}{*}{$90$} 	& \multirow{2}{*}{$0$} 		& DD-PINN+NMPC &$\mathbf{2.66}$ & $47.7$\\ \cline{3-5} 
		&                   		& PI~\cite{Habich.2024}      &$2.93$ & --\\ \cline{2-5} 
		& \multirow{2}{*}{$200$} 	& DD-PINN+NMPC &$\mathbf{2.98}$ & $46.8$\\ \cline{3-5} 
		&                   		& PI~\cite{Habich.2024}       &$3.36$ & --\\ \hline
	\end{tabular}}
	\label{tab:mpc_exp}
\end{table}

Experiments were conducted in \highlightred{six} domains with different desired trajectories of the positions.
For this purpose, \highlightred{linear} trajectories were generated, whereby both the rise time in the range \highlightred{$0.4{-}\SI{1.6}{\second}$} and the height of the amplitudes in the range ${\pm}\SI{18}{\degree}$ are determined randomly.
\highlightred{In order to reduce the sharp transitions between the linear trajectories, the reference was slightly smoothed using a moving mean.}
\highlightred{The average of the absolute joint velocities during all experiments was $\SI{11.70}{\degree/\second}$.
Thus, the resulting trajectories correspond to dynamic movements of the robot, whereby some sections are not even feasible.
Please watch the supporting video of this article to see the soft robot and its movements during control.}

\begin{figure}[t]
	\vspace{2.5mm}
	\centering{\includegraphics[width=\linewidth]{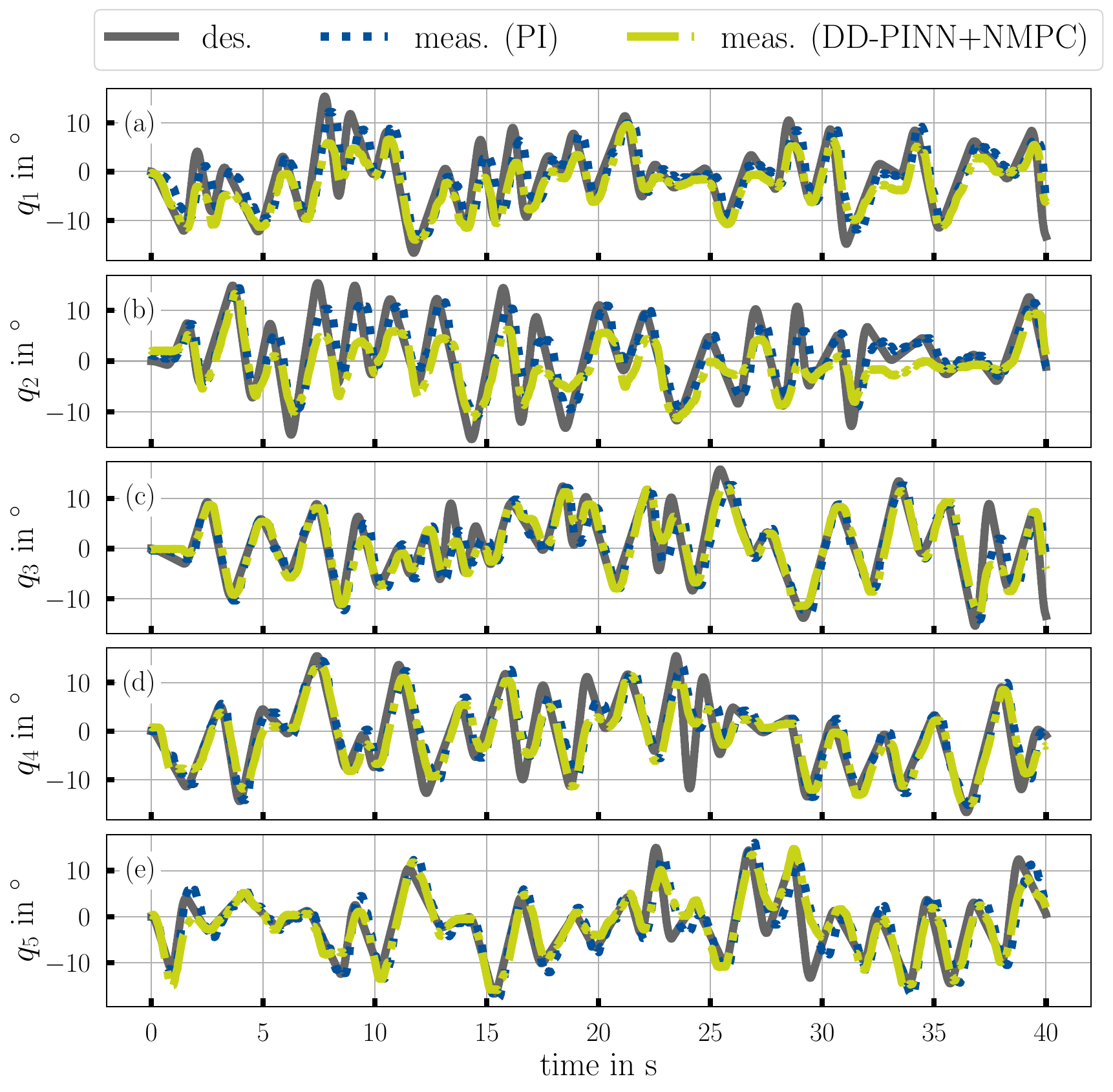}}
	\caption{\highlightred{(a)--(e) Control results for a randomly selected trajectory of all $n$~joints and $\mm{\delta}{=}[\SI{0}{\gram},\SI{90}{\degree}]\transpose$ using the proposed DD-PINN as a model within NMPC and comparison with PI control.}}
	\label{fig:control_plot}
\end{figure}
\begin{figure}[t]
	\vspace{2.5mm}
	\centering{\includegraphics[width=\linewidth]{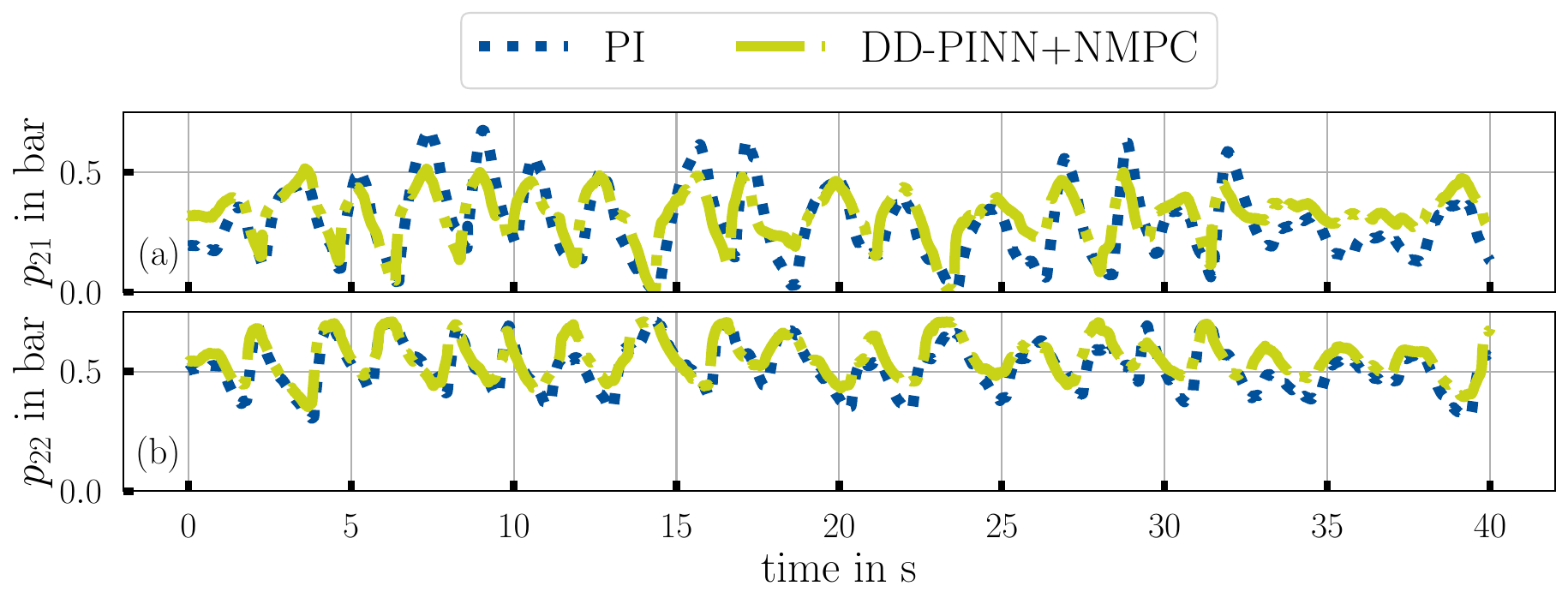}}
	\caption{\highlightred{Measured pressures (a)~$p_{21}$ and (b)~$p_{22}$ during the control experiment with $\mm{\delta}{=}[\SI{0}{\gram},\SI{90}{\degree}]\transpose$ for NMPC with the DD-PINN and PI control. The NMPC enables a more dynamic response due to the used model knowledge and online optimization.}}
	\label{fig:control_plot_pd}
\end{figure}
The results are presented in Table~\ref{tab:mpc_exp}.
In general, accurate trajectory tracking is realized with the nonlinear MPC, and the fast PINNs enable control frequencies $f_\ind{NMPC}$ up to \highlightred{\SI{47.7}{\hertz}}.
It can be seen that the NMPC approach achieves better tracking results due to its predictive nature, i.e., anticipating future system behavior and optimizing the input signal accordingly, rather than merely reacting to past control errors.
On average, an improvement of \highlightred{\SI{14.6}{\percent} can be realized by using the proposed DD-PINN+NMPC}.
\highlightred{The highest improvement of \SI{29.1}{\percent} was achieved for $\mm{\delta}{=}[\SI{200}{\gram},\SI{0}{\degree}]$. Thereby, especially poor tracking of the first joint was present during PI control due to the challenging task of balancing against gravity.}

It must be noted that the MPC runs with a significantly lower control frequency than the PI control ($\SI{1}{\kilo\hertz}$).
Better computing hardware in the future will improve the applicability of such MPC approaches \highlightred{further}.
For example, this would increase $f_\ind{NMPC}$, which will further improve the tracking performance.
\highlightred{However, the realized real-time nonlinear MPC with up to \highlightred{$\SI{47.7}{\hertz}$} is already a notable research achievement, which is only made possible by using PINNs.}
\highlightred{Depending on the system dynamics, the tracking accuracy of the MPC can also be higher in comparison to the PI control.}
If state constraints were required for other systems, these could also be easily handled using the MPC framework.

Exemplary control results for one test domain are visualized in Fig.~\ref{fig:control_plot}.
\highlightred{In some areas it can be seen that the steep changes can only be roughly tracked with a slight delay.}
In addition, for joint two in the present domain, several sections of the desired trajectory are not feasible due to gravity. 
This is illustrated in Fig.~\ref{fig:control_plot_pd} with the corresponding pressure curves, which run into saturation.
\highlightred{It can also often be seen that the PI controller has a delay and that the MPC reacts faster due to its predictive nature.}

\highlightred{\subsection{Overall Discussion}\label{results_discussion}
Three key claims of the article were given in Sec.~\ref{intro}.
Results show that the accurate physics-driven model can be accelerated by the fast surrogate by a factor of 467.
This comes at the cost of a slightly lower accuracy.
The hybrid model only requires data from one domain due to the use of prior physical knowledge and still allows a generalization to unknown dynamics.
Such a key advantage becomes apparent when comparing the proposed DD-PINN with a pure black-box model.
Finally, a real-time application is used to illustrate what becomes possible --- the generalizable and fast DD-PINN enables a high control frequency and accurate control for articulated soft robots with different payloads or base orientations.
\\Although the scenario presented in Fig.~\ref{fig:cover} is more academic, this article provides important insights for real-world applications.
In principle, changing conditions are always to be expected in reality, so generalizable models have great potential.
Grasping different masses or acting with a variable base orientation is therefore possible with model-based control and the proposed DD-PINN.
With longer versions of the snake robot, it is also possible that not every actuator is equipped with an encoder due to the cost.
Instead, the states could be obtained in real-time using a PINN-based estimator.
Furthermore, fast simulations are possible with PINNs.
This can be useful for a fast design optimization of the system or time-efficient (simulation-based) reinforcement learning, for example.
In general, utilizing PINNs on embedded or low-power computing platforms in wearable devices is also conceivable due to the simple feedforward structure.
}
\section{Conclusion}\label{conclusions}
Model-based estimation and control of soft robots require forward models that generalize to several system dynamics and provide a high prediction speed for real-time applications.
However, data-hungry black-box learning and slow physics-based models cannot fulfill both requirements.
In a literature overview, we analyze the current trend of hybrid model learning and identify a research gap regarding physics-informed neural networks for \textit{real-world, multi-DoF soft robots in small-data regimes}.
The latter is crucial: Real-world data is expensive due to the recording effort or possible maintenance/damage, and it is unrealistic that all future system conditions can be considered during training-data acquisition.

We extend two existing PINN architectures with domain knowledge and perform hybrid model learning for articulated soft robots.
In two hours of experiments, the presented model was tested in various system domains that were not seen during the recording of the training data.
Core results of our work (Table~\ref{tab:pred_speed} and Fig.~\ref{fig:spider_plot}) show that the proposed DD-PINN outperforms an elaborately identified first-principles model and a recurrent neural network \textit{when generalizability and prediction speed are considered jointly}.
As one possible real-time application, the DD-PINN enables nonlinear model predictive control and, thus, accurate position tracking in several system domains for dynamic references.

Diverse future research directions arise from our work.
The method is applicable to (soft) continuum robots, where dynamical models from Cosserat rod theory are accurate but require high computational costs.
Since the domain variables only need to appear in the dynamics equation, the PINN could be trained for other variable quantities, such as external contacts.
This could, in principle, be used for contact estimation during operation.
\highlightred{In order to further investigate the extent to which generalization is possible, considerably more than two domain variables should be examined.
In general, due to the required knowledge of the domain variables, online estimation is promising for unmeasurable quantities.
}
\highlightredfinal{Furthermore}, different ansatz functions of the DD-PINN could be compared, or suitable ansatz functions could be found using symbolic regression.
\highlightredfinal{Also, the feedforward structure of the DD-PINN could be replaced by an RNN to further improve the prediction accuracy.}
\highlightred{Since the DD-PINN training is still demanding, approaches from meta or transfer learning could be used to enhance generalization for system changes beyond the considered domain variables.}
Regarding control, GPU approaches~\cite{Hyatt.2020,Jeon.2025} could further improve the performance, and the combination with online learning could enable adaptivity.
Besides soft robotics, our architecture can be applied to various dynamical systems to provide \highlightred{generalizable and fast surrogates} for real-time estimation and control.


\section*{Acknowledgment}
We thank \highlightredfinal{Mehdi Belhadj} for collecting the datasets.
\bibliographystyle{IEEEtran}
\bibliography{literatur}
\vspace{-1.5cm}
\begin{IEEEbiography}[{\includegraphics[width=1in,height=1.25in,clip,keepaspectratio]{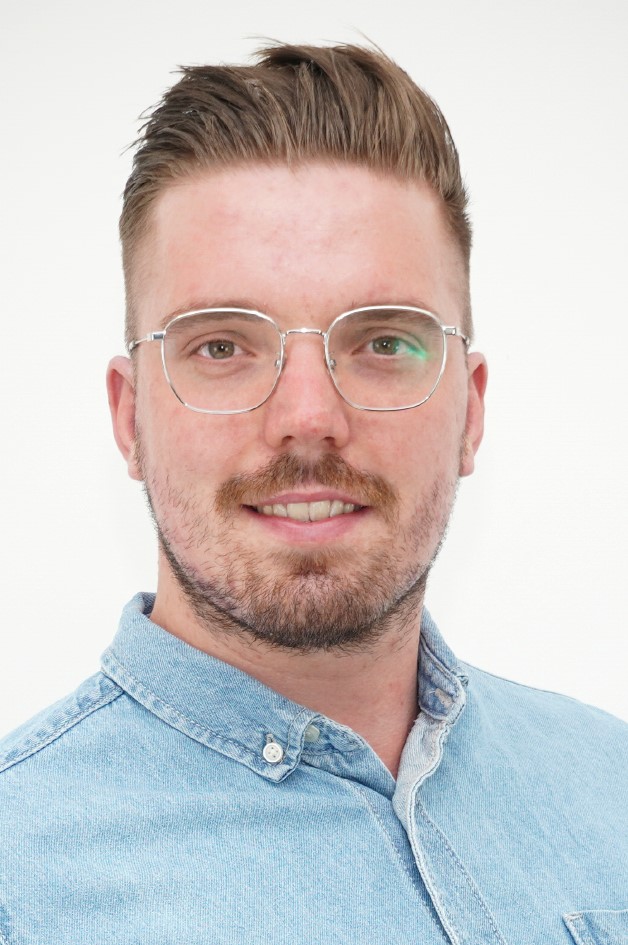}}]{Tim-Lukas Habich}
	\highlightredfinal{received his B.\,Sc. and M.\,Sc. degrees in Mechanical Engineering from Leibniz University Hannover (LUH), Germany, in 2017 and 2020, respectively. Currently, he is working towards his Ph.\,D. degree at the Institute of Mechatronic Systems (imes) at LUH. His research interests include physics-informed machine learning and learning-based control of mechatronic systems.}\end{IEEEbiography}
\vspace{-1.5cm}
\begin{IEEEbiography}[{\includegraphics[width=1in,height=1.25in,clip,keepaspectratio]{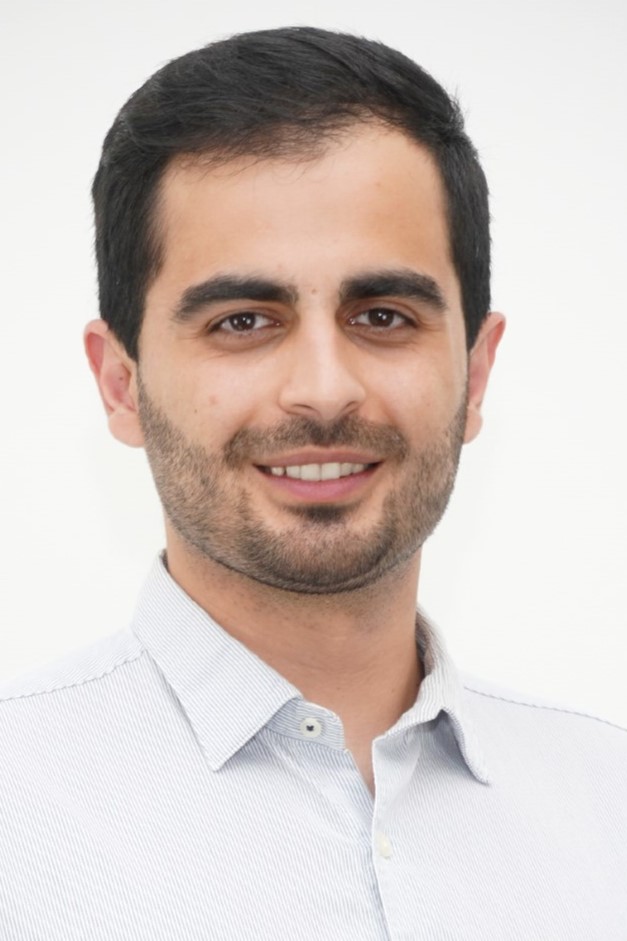}}]{Aran Mohammad}
	\highlightredfinal{received his B.\,Sc. and M.\,Sc. degrees in Mechanical Engineering from LUH, Germany, in 2017 and 2020, respectively. Currently, he is working towards his Ph.\,D. degree at the imes at LUH. His research interests include physics- and learning-based modeling, optimization and control of mechatronic systems.}\end{IEEEbiography}
\vspace{-1.5cm}
\begin{IEEEbiography}[{\includegraphics[width=1in,height=1.25in,clip,keepaspectratio]{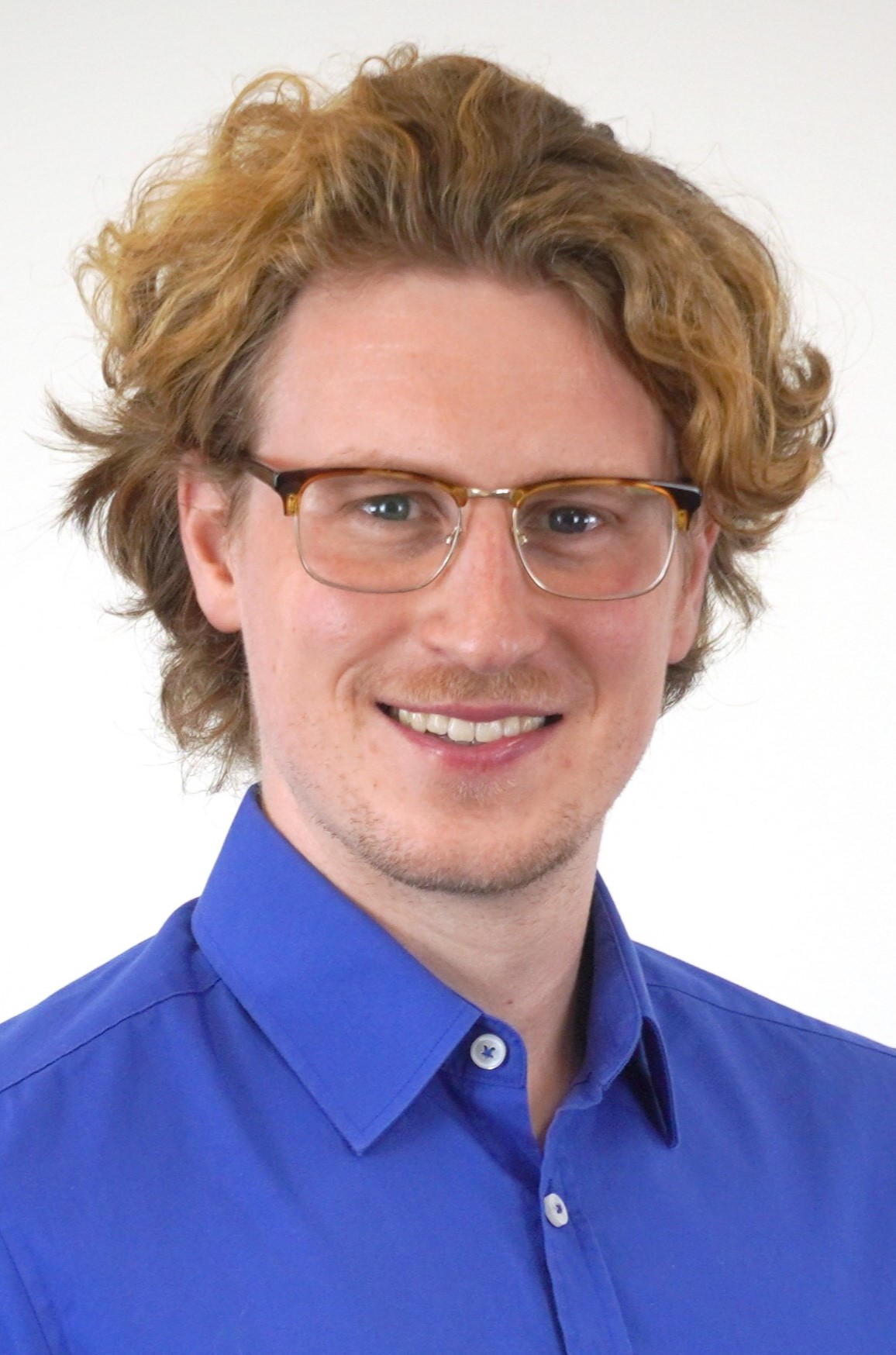}}]{Simon F. G. Ehlers}
	\highlightredfinal{received his B.\,Sc., M.\,Sc. and Ph.\,D. degrees in Mechanical Engineering from LUH, Germany, in 2016, 2019 and 2024, respectively. He is currently leading a research group on learning and control at the imes at LUH. His research interests include physics- and learning-based estimator concepts, especially in the field of vehicle dynamics.}\end{IEEEbiography}
\vspace{-1.5cm}
\begin{IEEEbiography}[{\includegraphics[width=1in,height=1.25in,clip,keepaspectratio]{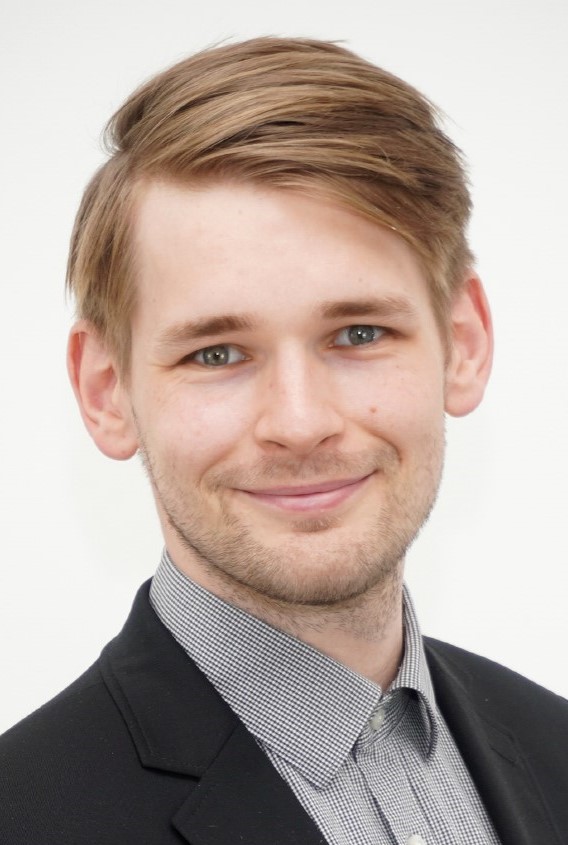}}]{Martin Bensch}
	\highlightredfinal{received his B.\,Sc. and M.\,Sc. degrees in Mechanical Engineering from LUH, Germany, in 2015 and 2019, respectively. Currently, he is working towards his Ph.\,D. degree at the imes at LUH. His research interests include continuum robots, machine learning, and path planning for continuum robots.}\end{IEEEbiography}
\vspace{-1.5cm}
\begin{IEEEbiography}[{\includegraphics[width=1in,height=1.25in,clip,keepaspectratio]{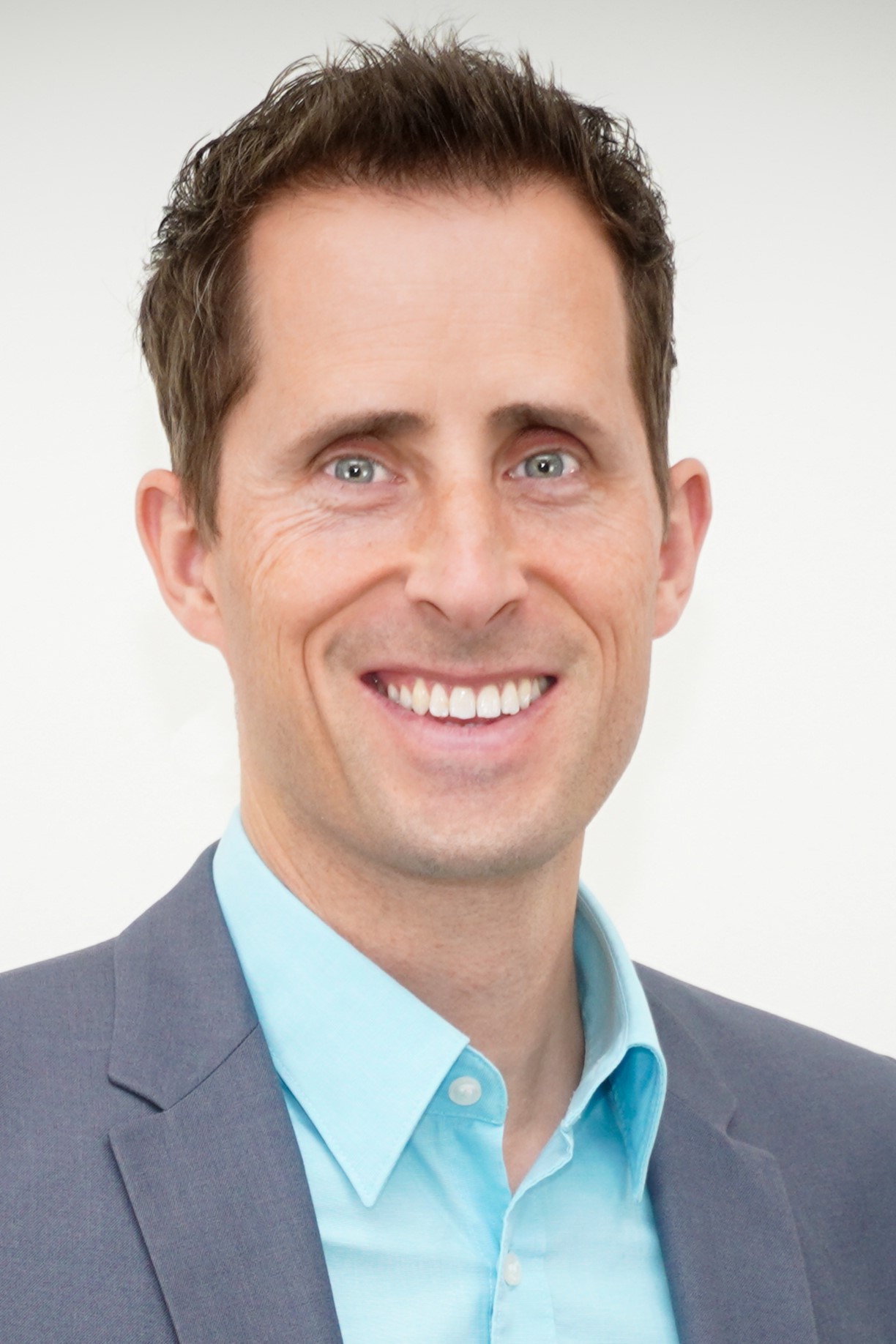}}]{Thomas Seel}
	\highlightredfinal{studied Engineering Cybernetics at Otto von Guericke University Magdeburg, Germany, and the University of California, USA. He received the Ph.\,D. degree from Technische Universität Berlin, Germany, in 2016. He is currently director of the imes at LUH. His research interests include dynamic inference and learning in biomedical and mechatronic systems.}\end{IEEEbiography}
\vspace{-1.5cm}
\begin{IEEEbiography}[{\includegraphics[width=1in,height=1.25in,clip,keepaspectratio]{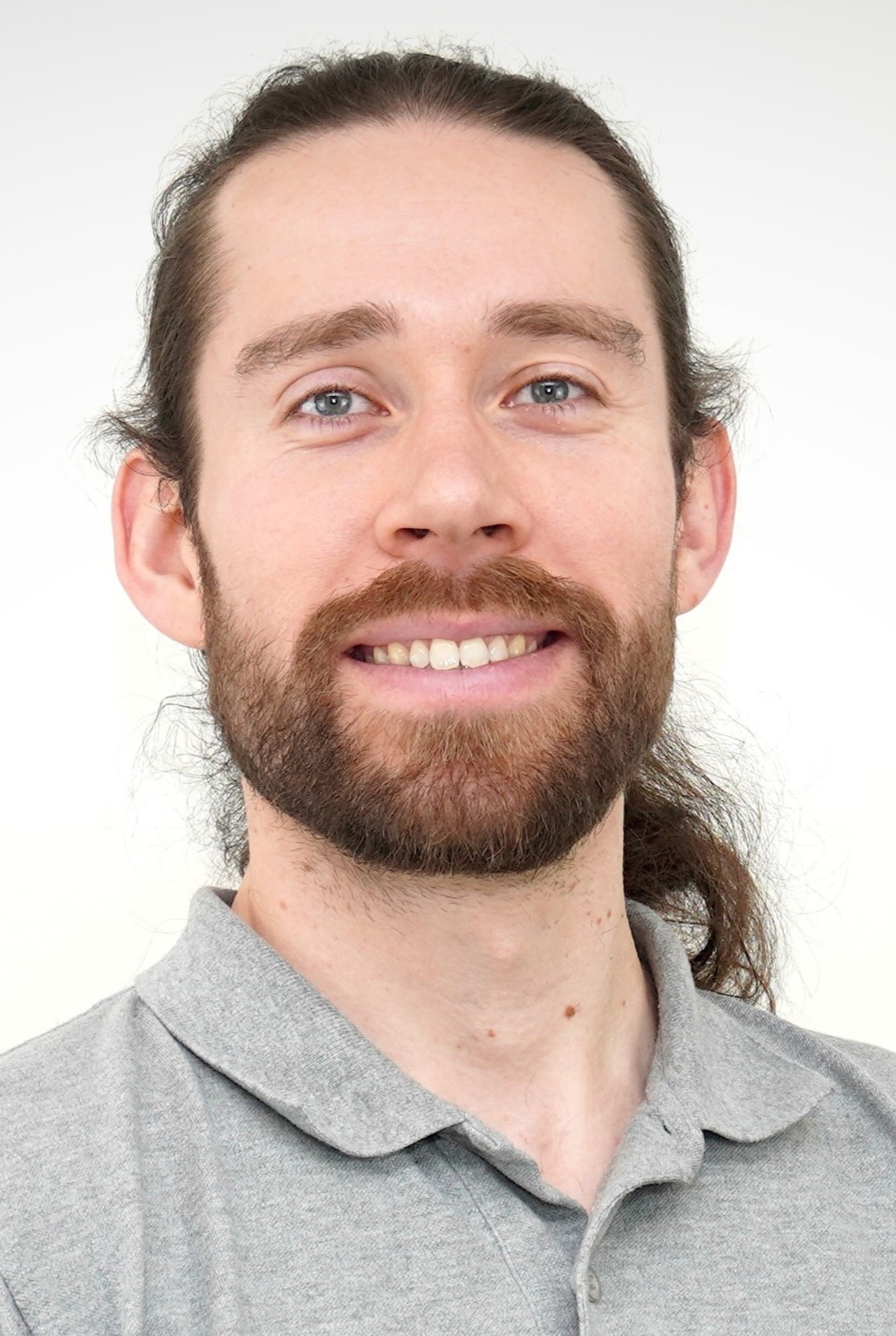}}]{Moritz Schappler}
	\highlightredfinal{received his B.\,Sc. and M.\,Sc. degrees in Mechatronics Engineering, and Ph.\,D. degree in Mechanical Engineering from LUH, Germany, in 2012, 2014 and 2025, respectively. He is currently leading a research group on robotic systems at the imes at LUH. His research interests include design, modeling and control of robotic and mechatronic systems with a focus on parallel robots, snake-like robots and methods for design optimization.}\end{IEEEbiography}
\end{document}

%% file: images/cover.pdf_tex
\begingroup%
  \makeatletter%
  \providecommand\color[2][]{%
    \errmessage{(Inkscape) Color is used for the text in Inkscape, but the package 'color.sty' is not loaded}%
    \renewcommand\color[2][]{}%
  }%
  \providecommand\transparent[1]{%
    \errmessage{(Inkscape) Transparency is used (non-zero) for the text in Inkscape, but the package 'transparent.sty' is not loaded}%
    \renewcommand\transparent[1]{}%
  }%
  \providecommand\rotatebox[2]{#2}%
  \newcommand*\fsize{\dimexpr\f@size pt\relax}%
  \newcommand*\lineheight[1]{\fontsize{\fsize}{#1\fsize}\selectfont}%
  \ifx\svgwidth\undefined%
    \setlength{\unitlength}{232.38542932bp}%
    \ifx\svgscale\undefined%
      \relax%
    \else%
      \setlength{\unitlength}{\unitlength * \real{\svgscale}}%
    \fi%
  \else%
    \setlength{\unitlength}{\svgwidth}%
  \fi%
  \global\let\svgwidth\undefined%
  \global\let\svgscale\undefined%
  \makeatother%
  \begin{picture}(1,0.57692211)%
    \lineheight{1}%
    \setlength\tabcolsep{0pt}%
    \put(0,0){\includegraphics[width=\unitlength,page=1]{cover.pdf}}%
    \put(0.51614529,0.004308){\makebox(0,0)[t]{\lineheight{1.25}\smash{\begin{tabular}[t]{c}\scriptsize{training domain}\end{tabular}}}}%
    \put(0.01291125,0.2604901){\rotatebox{90}{\makebox(0,0)[t]{\lineheight{1.25}\smash{\begin{tabular}[t]{c}\scriptsize{generalizability}\end{tabular}}}}}%
    \put(0.415529,0.30325312){\rotatebox{90}{\makebox(0,0)[t]{\lineheight{1.25}\smash{\begin{tabular}[t]{c}\scriptsize{orientation and payload constant}\end{tabular}}}}}%
    \put(0.18987098,0.00430316){\makebox(0,0)[t]{\lineheight{1.25}\smash{\begin{tabular}[t]{c}\scriptsize{prediction speed}\end{tabular}}}}%
    \put(0.25432481,0.11954412){\makebox(0,0)[t]{\lineheight{1.25}\smash{\begin{tabular}[t]{c}\scriptsize{black-box}\end{tabular}}}}%
    \put(0.25546568,0.08916071){\makebox(0,0)[t]{\lineheight{1.25}\smash{\begin{tabular}[t]{c}\scriptsize{learning}\end{tabular}}}}%
    \put(0.15750273,0.48181941){\makebox(0,0)[t]{\lineheight{1.25}\smash{\begin{tabular}[t]{c}\scriptsize{first-principles}\end{tabular}}}}%
    \put(-0.00097998,0.53749211){\makebox(0,0)[lt]{\lineheight{1.25}\smash{\begin{tabular}[t]{l}\scriptsize{(a)}\end{tabular}}}}%
    \put(0.4191353,0.53749211){\makebox(0,0)[rt]{\lineheight{1.25}\smash{\begin{tabular}[t]{r}\scriptsize{(b)}\end{tabular}}}}%
    \put(0.15864378,0.45143599){\makebox(0,0)[t]{\lineheight{1.25}\smash{\begin{tabular}[t]{c}\scriptsize{modeling}\end{tabular}}}}%
    \put(0,0){\includegraphics[width=\unitlength,page=2]{cover.pdf}}%
    \put(0.20510675,0.34789925){\makebox(0,0)[t]{\lineheight{1.25}\smash{\begin{tabular}[t]{c}\scriptsize{physics-informed ML}\end{tabular}}}}%
    \put(0.20551008,0.3156253){\makebox(0,0)[t]{\lineheight{1.25}\smash{\begin{tabular}[t]{c}\scriptsize{enables fast prediction}\end{tabular}}}}%
    \put(0.20551027,0.2833513){\makebox(0,0)[t]{\lineheight{1.25}\smash{\begin{tabular}[t]{c}\scriptsize{and generalization to}\end{tabular}}}}%
    \put(0.20510685,0.2510773){\makebox(0,0)[t]{\lineheight{1.25}\smash{\begin{tabular}[t]{c}\scriptsize{\orange{unseen test domains}}\end{tabular}}}}%
    \put(0,0){\includegraphics[width=\unitlength,page=3]{cover.pdf}}%
    \put(0.82893093,0.004308){\makebox(0,0)[t]{\lineheight{1.25}\smash{\begin{tabular}[t]{c}\scriptsize{\orange{unseen test domains}}\end{tabular}}}}%
    \put(0,0){\includegraphics[width=\unitlength,page=4]{cover.pdf}}%
    \put(0.6460092,0.1441468){\rotatebox{90}{\makebox(0,0)[t]{\lineheight{1.25}\smash{\begin{tabular}[t]{c}\scriptsize{\orange{orientation change}}\end{tabular}}}}}%
    \put(0.6460093,0.41433618){\rotatebox{90}{\makebox(0,0)[t]{\lineheight{1.25}\smash{\begin{tabular}[t]{c}\scriptsize{\orange{payload change}}\end{tabular}}}}}%
  \end{picture}%
\endgroup%

%% file: images/overview_main.pdf_tex
\begingroup%
  \makeatletter%
  \providecommand\color[2][]{%
    \errmessage{(Inkscape) Color is used for the text in Inkscape, but the package 'color.sty' is not loaded}%
    \renewcommand\color[2][]{}%
  }%
  \providecommand\transparent[1]{%
    \errmessage{(Inkscape) Transparency is used (non-zero) for the text in Inkscape, but the package 'transparent.sty' is not loaded}%
    \renewcommand\transparent[1]{}%
  }%
  \providecommand\rotatebox[2]{#2}%
  \newcommand*\fsize{\dimexpr\f@size pt\relax}%
  \newcommand*\lineheight[1]{\fontsize{\fsize}{#1\fsize}\selectfont}%
  \ifx\svgwidth\undefined%
    \setlength{\unitlength}{232.6540537bp}%
    \ifx\svgscale\undefined%
      \relax%
    \else%
      \setlength{\unitlength}{\unitlength * \real{\svgscale}}%
    \fi%
  \else%
    \setlength{\unitlength}{\svgwidth}%
  \fi%
  \global\let\svgwidth\undefined%
  \global\let\svgscale\undefined%
  \makeatother%
  \begin{picture}(1,0.46551503)%
    \lineheight{1}%
    \setlength\tabcolsep{0pt}%
    \put(0,0){\includegraphics[width=\unitlength,page=1]{overview_main.pdf}}%
    \put(0.46424209,0.39454051){\makebox(0,0)[t]{\smash{\begin{tabular}[t]{c}\scriptsize data\end{tabular}}}}%
    \put(0.75634408,0.31422354){\makebox(0,0)[lt]{\smash{\begin{tabular}[t]{l}\scriptsize ODEs\end{tabular}}}}%
    \put(0,0){\includegraphics[width=\unitlength,page=2]{overview_main.pdf}}%
    \put(0.07236343,0.17895748){\makebox(0,0)[lt]{\smash{\begin{tabular}[t]{l}\scriptsize soft-robot\end{tabular}}}}%
    \put(0.06591609,0.10890618){\makebox(0,0)[lt]{\smash{\begin{tabular}[t]{l}\scriptsize (Sec.~\ref{snake_robot})\end{tabular}}}}%
    \put(0.08525811,0.14393058){\makebox(0,0)[lt]{\smash{\begin{tabular}[t]{l}\scriptsize platform\end{tabular}}}}%
    \put(0,0){\includegraphics[width=\unitlength,page=3]{overview_main.pdf}}%
    \put(0.74345171,0.36300773){\makebox(0,0)[t]{\smash{\begin{tabular}[t]{c}\scriptsize (Sec.~\ref{first_princ})\end{tabular}}}}%
    \put(0.74352863,0.39803478){\makebox(0,0)[t]{\smash{\begin{tabular}[t]{c}\scriptsize FP modeling and identification\end{tabular}}}}%
    \put(0.50094741,0.01643561){\makebox(0,0)[t]{\smash{\begin{tabular}[t]{c}\scriptsize real-time application: nonlinear MPC of ASRs using PINNs (Sec.~\ref{mpc})\end{tabular}}}}%
    \put(0.74345172,0.23338732){\makebox(0,0)[t]{\smash{\begin{tabular}[t]{c}\scriptsize (Sec.~\ref{pinn_subsection}, \ref{pinn_impl} and \ref{dom_know})\end{tabular}}}}%
    \put(0.74617363,0.26841441){\makebox(0,0)[t]{\smash{\begin{tabular}[t]{c}\scriptsize generalizable and fast PINNs \end{tabular}}}}%
    \put(0,0){\includegraphics[width=\unitlength,page=4]{overview_main.pdf}}%
  \end{picture}%
\endgroup%

%% file: images/kinematic_chain.pdf_tex
\begingroup%
  \makeatletter%
  \providecommand\color[2][]{%
    \errmessage{(Inkscape) Color is used for the text in Inkscape, but the package 'color.sty' is not loaded}%
    \renewcommand\color[2][]{}%
  }%
  \providecommand\transparent[1]{%
    \errmessage{(Inkscape) Transparency is used (non-zero) for the text in Inkscape, but the package 'transparent.sty' is not loaded}%
    \renewcommand\transparent[1]{}%
  }%
  \providecommand\rotatebox[2]{#2}%
  \newcommand*\fsize{\dimexpr\f@size pt\relax}%
  \newcommand*\lineheight[1]{\fontsize{\fsize}{#1\fsize}\selectfont}%
  \ifx\svgwidth\undefined%
    \setlength{\unitlength}{232.37310214bp}%
    \ifx\svgscale\undefined%
      \relax%
    \else%
      \setlength{\unitlength}{\unitlength * \real{\svgscale}}%
    \fi%
  \else%
    \setlength{\unitlength}{\svgwidth}%
  \fi%
  \global\let\svgwidth\undefined%
  \global\let\svgscale\undefined%
  \makeatother%
  \begin{picture}(1,0.56980682)%
    \lineheight{1}%
    \setlength\tabcolsep{0pt}%
    \put(0,0){\includegraphics[width=\unitlength,page=1]{kinematic_chain.pdf}}%
    \put(1.00038531,0.17029172){\makebox(0,0)[rt]{\lineheight{1.25}\smash{\begin{tabular}[t]{r}\scriptsize$p_{12}$\end{tabular}}}}%
    \put(0,0){\includegraphics[width=\unitlength,page=2]{kinematic_chain.pdf}}%
    \put(0.51566144,0.09442595){\makebox(0,0)[t]{\lineheight{1.25}\smash{\begin{tabular}[t]{c}\scriptsize$q_1$\end{tabular}}}}%
    \put(0,0){\includegraphics[width=\unitlength,page=3]{kinematic_chain.pdf}}%
    \put(0.07535625,0.12857466){\makebox(0,0)[rt]{\lineheight{1.25}\smash{\begin{tabular}[t]{r}\scriptsize$h$\end{tabular}}}}%
    \put(0,0){\includegraphics[width=\unitlength,page=4]{kinematic_chain.pdf}}%
    \put(0.63435894,0.0545027){\makebox(0,0)[rt]{\lineheight{1.25}\smash{\begin{tabular}[t]{r}\scriptsize$p_{11}$\end{tabular}}}}%
    \put(0,0){\includegraphics[width=\unitlength,page=5]{kinematic_chain.pdf}}%
    \put(0.61539697,0.13075396){\makebox(0,0)[rt]{\lineheight{1.25}\smash{\begin{tabular}[t]{r}\scriptsize$p_{22}$\end{tabular}}}}%
    \put(0,0){\includegraphics[width=\unitlength,page=6]{kinematic_chain.pdf}}%
    \put(0.57949028,0.21789829){\makebox(0,0)[rt]{\lineheight{1.25}\smash{\begin{tabular}[t]{r}\scriptsize$p_{31}$\end{tabular}}}}%
    \put(0,0){\includegraphics[width=\unitlength,page=7]{kinematic_chain.pdf}}%
    \put(0.56173866,0.30544607){\makebox(0,0)[rt]{\lineheight{1.25}\smash{\begin{tabular}[t]{r}\scriptsize$p_{42}$\end{tabular}}}}%
    \put(0,0){\includegraphics[width=\unitlength,page=8]{kinematic_chain.pdf}}%
    \put(0.52018373,0.38169737){\makebox(0,0)[rt]{\lineheight{1.25}\smash{\begin{tabular}[t]{r}\scriptsize$p_{51}$\end{tabular}}}}%
    \put(0,0){\includegraphics[width=\unitlength,page=9]{kinematic_chain.pdf}}%
    \put(0.98815535,0.26928367){\makebox(0,0)[rt]{\lineheight{1.25}\smash{\begin{tabular}[t]{r}\scriptsize$p_{21}$\end{tabular}}}}%
    \put(0,0){\includegraphics[width=\unitlength,page=10]{kinematic_chain.pdf}}%
    \put(0.49226158,0.17081076){\makebox(0,0)[t]{\lineheight{1.25}\smash{\begin{tabular}[t]{c}\scriptsize$q_2$\end{tabular}}}}%
    \put(0,0){\includegraphics[width=\unitlength,page=11]{kinematic_chain.pdf}}%
    \put(0.46374058,0.25760576){\makebox(0,0)[t]{\lineheight{1.25}\smash{\begin{tabular}[t]{c}\scriptsize$q_3$\end{tabular}}}}%
    \put(0,0){\includegraphics[width=\unitlength,page=12]{kinematic_chain.pdf}}%
    \put(-0.00031108,0.53338521){\makebox(0,0)[lt]{\lineheight{1.25}\smash{\begin{tabular}[t]{l}\scriptsize$m_5{+}m_\mathrm{e}$\end{tabular}}}}%
    \put(0,0){\includegraphics[width=\unitlength,page=13]{kinematic_chain.pdf}}%
    \put(0.44024826,0.05221917){\makebox(0,0)[lt]{\lineheight{1.25}\smash{\begin{tabular}[t]{l}\scriptsize$\beta_\ind{g}$\end{tabular}}}}%
    \put(0.31494729,0.5457443){\makebox(0,0)[lt]{\lineheight{1.25}\smash{\begin{tabular}[t]{l}\scriptsize gravity\end{tabular}}}}%
    \put(0,0){\includegraphics[width=\unitlength,page=14]{kinematic_chain.pdf}}%
    \put(0.65527675,0.00767935){\makebox(0,0)[lt]{\lineheight{1.25}\smash{\begin{tabular}[t]{l}\scriptsize(b)\end{tabular}}}}%
    \put(0.04882566,0.0063789){\makebox(0,0)[lt]{\lineheight{1.25}\smash{\begin{tabular}[t]{l}\scriptsize(a)\end{tabular}}}}%
    \put(0,0){\includegraphics[width=\unitlength,page=15]{kinematic_chain.pdf}}%
    \put(0.14368086,0.54044135){\makebox(0,0)[lt]{\lineheight{1.25}\smash{\begin{tabular}[t]{l}\scriptsize$\fra{5}$\end{tabular}}}}%
    \put(0.25271579,0.11290446){\makebox(0,0)[lt]{\lineheight{1.25}\smash{\begin{tabular}[t]{l}\scriptsize$\fra{0}$\end{tabular}}}}%
    \put(0.2487335,0.21080256){\makebox(0,0)[lt]{\lineheight{1.25}\smash{\begin{tabular}[t]{l}\scriptsize$\fra{1}$\end{tabular}}}}%
    \put(0.20492824,0.30463102){\makebox(0,0)[lt]{\lineheight{1.25}\smash{\begin{tabular}[t]{l}\scriptsize$\fra{2}$\end{tabular}}}}%
    \put(0.19795927,0.40029348){\makebox(0,0)[lt]{\lineheight{1.25}\smash{\begin{tabular}[t]{l}\scriptsize$\fra{3}$\end{tabular}}}}%
    \put(0.15382205,0.49354555){\makebox(0,0)[lt]{\lineheight{1.25}\smash{\begin{tabular}[t]{l}\scriptsize$\fra{4}$\end{tabular}}}}%
    \put(0.94907205,0.37341064){\makebox(0,0)[rt]{\lineheight{1.25}\smash{\begin{tabular}[t]{r}\scriptsize$p_{32}$\end{tabular}}}}%
    \put(0,0){\includegraphics[width=\unitlength,page=16]{kinematic_chain.pdf}}%
    \put(0.87701948,0.46814019){\makebox(0,0)[rt]{\lineheight{1.25}\smash{\begin{tabular}[t]{r}\scriptsize$p_{41}$\end{tabular}}}}%
    \put(0,0){\includegraphics[width=\unitlength,page=17]{kinematic_chain.pdf}}%
    \put(0.78019244,0.56093276){\makebox(0,0)[rt]{\lineheight{1.25}\smash{\begin{tabular}[t]{r}\scriptsize$p_{52}$\end{tabular}}}}%
    \put(0,0){\includegraphics[width=\unitlength,page=18]{kinematic_chain.pdf}}%
    \put(0.44142739,0.33789335){\makebox(0,0)[t]{\lineheight{1.25}\smash{\begin{tabular}[t]{c}\scriptsize$q_4$\end{tabular}}}}%
    \put(0,0){\includegraphics[width=\unitlength,page=19]{kinematic_chain.pdf}}%
    \put(0.41524307,0.41736222){\makebox(0,0)[t]{\lineheight{1.25}\smash{\begin{tabular}[t]{c}\scriptsize$q_5$\end{tabular}}}}%
    \put(0,0){\includegraphics[width=\unitlength,page=20]{kinematic_chain.pdf}}%
  \end{picture}%
\endgroup%

%% file: images/KMS_experiment.pdf_tex
\begingroup%
  \makeatletter%
  \providecommand\color[2][]{%
    \errmessage{(Inkscape) Color is used for the text in Inkscape, but the package 'color.sty' is not loaded}%
    \renewcommand\color[2][]{}%
  }%
  \providecommand\transparent[1]{%
    \errmessage{(Inkscape) Transparency is used (non-zero) for the text in Inkscape, but the package 'transparent.sty' is not loaded}%
    \renewcommand\transparent[1]{}%
  }%
  \providecommand\rotatebox[2]{#2}%
  \newcommand*\fsize{\dimexpr\f@size pt\relax}%
  \newcommand*\lineheight[1]{\fontsize{\fsize}{#1\fsize}\selectfont}%
  \ifx\svgwidth\undefined%
    \setlength{\unitlength}{232.25688063bp}%
    \ifx\svgscale\undefined%
      \relax%
    \else%
      \setlength{\unitlength}{\unitlength * \real{\svgscale}}%
    \fi%
  \else%
    \setlength{\unitlength}{\svgwidth}%
  \fi%
  \global\let\svgwidth\undefined%
  \global\let\svgscale\undefined%
  \makeatother%
  \begin{picture}(1,0.40300211)%
    \lineheight{1}%
    \setlength\tabcolsep{0pt}%
    \put(0,0){\includegraphics[width=\unitlength,page=1]{KMS_experiment.pdf}}%
    \put(0.00660138,0.35687842){\makebox(0,0)[lt]{\lineheight{1.25}\smash{\begin{tabular}[t]{l}\scriptsize(a)\end{tabular}}}}%
    \put(0,0){\includegraphics[width=\unitlength,page=2]{KMS_experiment.pdf}}%
    \put(0.08586815,0.01313171){\makebox(0,0)[t]{\lineheight{1.25}\smash{\begin{tabular}[t]{c}\scriptsize torque sensor\end{tabular}}}}%
    \put(0,0){\includegraphics[width=\unitlength,page=3]{KMS_experiment.pdf}}%
    \put(0.25741844,0.01313171){\makebox(0,0)[t]{\lineheight{1.25}\smash{\begin{tabular}[t]{c}\scriptsize coupling\end{tabular}}}}%
    \put(0,0){\includegraphics[width=\unitlength,page=4]{KMS_experiment.pdf}}%
    \put(0.41443747,0.01312808){\makebox(0,0)[t]{\lineheight{1.25}\smash{\begin{tabular}[t]{c}\scriptsize actuator 1\end{tabular}}}}%
    \put(0.49743704,0.35687842){\makebox(0,0)[lt]{\lineheight{1.25}\smash{\begin{tabular}[t]{l}\scriptsize(b)\end{tabular}}}}%
  \end{picture}%
\endgroup%

%% file: images/pinn_structures.pdf_tex
\begingroup%
  \makeatletter%
  \providecommand\color[2][]{%
    \errmessage{(Inkscape) Color is used for the text in Inkscape, but the package 'color.sty' is not loaded}%
    \renewcommand\color[2][]{}%
  }%
  \providecommand\transparent[1]{%
    \errmessage{(Inkscape) Transparency is used (non-zero) for the text in Inkscape, but the package 'transparent.sty' is not loaded}%
    \renewcommand\transparent[1]{}%
  }%
  \providecommand\rotatebox[2]{#2}%
  \newcommand*\fsize{\dimexpr\f@size pt\relax}%
  \newcommand*\lineheight[1]{\fontsize{\fsize}{#1\fsize}\selectfont}%
  \ifx\svgwidth\undefined%
    \setlength{\unitlength}{229.32863491bp}%
    \ifx\svgscale\undefined%
      \relax%
    \else%
      \setlength{\unitlength}{\unitlength * \real{\svgscale}}%
    \fi%
  \else%
    \setlength{\unitlength}{\svgwidth}%
  \fi%
  \global\let\svgwidth\undefined%
  \global\let\svgscale\undefined%
  \makeatother%
  \begin{picture}(1,1.0672027)%
    \lineheight{1}%
    \setlength\tabcolsep{0pt}%
    \put(0,0){\includegraphics[width=\unitlength,page=1]{pinn_structures.pdf}}%
    \put(0.03056423,0.37908185){\makebox(0,0)[lt]{\lineheight{1.25}\smash{\begin{tabular}[t]{l}\scriptsize$\mm{x}_0$\end{tabular}}}}%
    \put(0.03104053,0.27304941){\makebox(0,0)[lt]{\lineheight{1.25}\smash{\begin{tabular}[t]{l}\scriptsize$\mm{u}_0$\end{tabular}}}}%
    \put(0.03931322,0.16413213){\makebox(0,0)[lt]{\lineheight{1.25}\smash{\begin{tabular}[t]{l}\scriptsize$\mm{\delta}$\end{tabular}}}}%
    \put(0,0){\includegraphics[width=\unitlength,page=2]{pinn_structures.pdf}}%
    \put(0.58549514,0.48452038){\makebox(0,0)[t]{\lineheight{1.25}\smash{\begin{tabular}[t]{c}\scriptsize$\mm{a}$\end{tabular}}}}%
    \put(0.76705192,0.16336557){\makebox(0,0)[lt]{\lineheight{1.25}\smash{\begin{tabular}[t]{l}\scriptsize$\mm{I}$\end{tabular}}}}%
    \put(0.76705192,0.26939357){\makebox(0,0)[lt]{\lineheight{1.25}\smash{\begin{tabular}[t]{l}\scriptsize$\mm{I}$\end{tabular}}}}%
    \put(0.09627121,0.05981219){\makebox(0,0)[t]{\lineheight{1.25}\smash{\begin{tabular}[t]{c}\scriptsize data $\mm{x}(T_\ind{s})$\end{tabular}}}}%
    \put(0,0){\includegraphics[width=\unitlength,page=3]{pinn_structures.pdf}}%
    \put(0.54718325,0.05711671){\makebox(0,0)[lt]{\lineheight{-13.36629963}\smash{\begin{tabular}[t]{l}\scriptsize $\eta_\ind{d}\mc{L}_\mathrm{d}$\end{tabular}}}}%
    \put(0.63370314,0.05711671){\makebox(0,0)[lt]{\lineheight{-13.36629963}\smash{\begin{tabular}[t]{l}\scriptsize $+$\end{tabular}}}}%
    \put(0.70992248,0.05711671){\makebox(0,0)[t]{\lineheight{-13.36629963}\smash{\begin{tabular}[t]{c}\scriptsize $\eta_\ind{p}\mc{L}_\mathrm{p}$\end{tabular}}}}%
    \put(0,0){\includegraphics[width=\unitlength,page=4]{pinn_structures.pdf}}%
    \put(0.86834152,0.42800293){\makebox(0,0)[lt]{\lineheight{1.25}\smash{\begin{tabular}[t]{l}\scriptsize$\frac{\partial\hat{\mm{x}}}{\partial t}$\end{tabular}}}}%
    \put(0.83154936,0.37853792){\makebox(0,0)[lt]{\lineheight{1.25}\smash{\begin{tabular}[t]{l}\scriptsize$\hat{\mm{x}}$\end{tabular}}}}%
    \put(0.99299407,0.32432423){\makebox(0,0)[rt]{\lineheight{1.25}\smash{\begin{tabular}[t]{r}\scriptsize(\ref{eq:r_physics})\end{tabular}}}}%
    \put(0,0){\includegraphics[width=\unitlength,page=5]{pinn_structures.pdf}}%
    \put(0.77613373,0.3782719){\makebox(0,0)[t]{\lineheight{1.25}\smash{\begin{tabular}[t]{c}\scriptsize$+$\end{tabular}}}}%
    \put(0,0){\includegraphics[width=\unitlength,page=6]{pinn_structures.pdf}}%
    \put(0.04303915,0.48386443){\makebox(0,0)[lt]{\lineheight{1.25}\smash{\begin{tabular}[t]{l}\scriptsize$t$\end{tabular}}}}%
    \put(-0.00094698,0.00415796){\makebox(0,0)[lt]{\lineheight{1.25}\smash{\begin{tabular}[t]{l}\scriptsize(b)\end{tabular}}}}%
    \put(0,0){\includegraphics[width=\unitlength,page=7]{pinn_structures.pdf}}%
    \put(0.03104053,0.81524408){\makebox(0,0)[lt]{\lineheight{1.25}\smash{\begin{tabular}[t]{l}\scriptsize$\mm{u}_0$\end{tabular}}}}%
    \put(0.03931322,0.70632681){\makebox(0,0)[lt]{\lineheight{1.25}\smash{\begin{tabular}[t]{l}\scriptsize$\mm{\delta}$\end{tabular}}}}%
    \put(0.76705192,0.70556024){\makebox(0,0)[lt]{\lineheight{1.25}\smash{\begin{tabular}[t]{l}\scriptsize$\mm{I}$\end{tabular}}}}%
    \put(0.76705192,0.81158825){\makebox(0,0)[lt]{\lineheight{1.25}\smash{\begin{tabular}[t]{l}\scriptsize$\mm{I}$\end{tabular}}}}%
    \put(0.09627121,0.60200691){\makebox(0,0)[t]{\lineheight{1.25}\smash{\begin{tabular}[t]{c}\scriptsize data $\mm{x}(T_\ind{s})$\end{tabular}}}}%
    \put(0,0){\includegraphics[width=\unitlength,page=8]{pinn_structures.pdf}}%
    \put(0.41357012,0.59931134){\makebox(0,0)[rt]{\lineheight{-13.36629963}\smash{\begin{tabular}[t]{r}\scriptsize loss $\mc{L}=$\end{tabular}}}}%
    \put(0,0){\includegraphics[width=\unitlength,page=9]{pinn_structures.pdf}}%
    \put(0.53656799,0.05711633){\makebox(0,0)[rt]{\lineheight{-13.36629963}\smash{\begin{tabular}[t]{r}\scriptsize loss $\mc{L}=$\end{tabular}}}}%
    \put(0.54718325,0.59931143){\makebox(0,0)[lt]{\lineheight{-13.36629963}\smash{\begin{tabular}[t]{l}\scriptsize $\eta_\ind{d}\mc{L}_\mathrm{d}$\end{tabular}}}}%
    \put(0.42521168,0.59931143){\makebox(0,0)[lt]{\lineheight{-13.36629963}\smash{\begin{tabular}[t]{l}\scriptsize $\eta_\ind{0}\mc{L}_\mathrm{0}$\end{tabular}}}}%
    \put(0,0){\includegraphics[width=\unitlength,page=10]{pinn_structures.pdf}}%
    \put(0.99299407,0.86651891){\makebox(0,0)[rt]{\lineheight{1.25}\smash{\begin{tabular}[t]{r}\scriptsize(\ref{eq:r_physics})\end{tabular}}}}%
    \put(0,0){\includegraphics[width=\unitlength,page=11]{pinn_structures.pdf}}%
    \put(0.77613373,1.02713217){\makebox(0,0)[t]{\lineheight{1.25}\smash{\begin{tabular}[t]{c}\scriptsize$\frac{\partial}{\partial t}$\end{tabular}}}}%
    \put(0.77613373,0.48437582){\makebox(0,0)[t]{\lineheight{1.25}\smash{\begin{tabular}[t]{c}\scriptsize$\frac{\partial}{\partial t}$\end{tabular}}}}%
    \put(0,0){\includegraphics[width=\unitlength,page=12]{pinn_structures.pdf}}%
    \put(0.04303915,1.02605915){\makebox(0,0)[lt]{\lineheight{1.25}\smash{\begin{tabular}[t]{l}\scriptsize$t$\end{tabular}}}}%
    \put(-0.00094698,0.54635259){\makebox(0,0)[lt]{\lineheight{1.25}\smash{\begin{tabular}[t]{l}\scriptsize(a)\end{tabular}}}}%
    \put(0,0){\includegraphics[width=\unitlength,page=13]{pinn_structures.pdf}}%
    \put(0.76795426,0.91933676){\makebox(0,0)[lt]{\lineheight{1.25}\smash{\begin{tabular}[t]{l}\scriptsize$\mm{I}$\end{tabular}}}}%
    \put(0,0){\includegraphics[width=\unitlength,page=14]{pinn_structures.pdf}}%
    \put(0.57478539,0.86659491){\makebox(0,0)[lt]{\lineheight{1.25}\smash{\begin{tabular}[t]{l}\scriptsize$\hat{\mm{x}}$\end{tabular}}}}%
    \put(0,0){\includegraphics[width=\unitlength,page=15]{pinn_structures.pdf}}%
    \put(0.03056423,0.92127652){\makebox(0,0)[lt]{\lineheight{1.25}\smash{\begin{tabular}[t]{l}\scriptsize$\mm{x}_0$\end{tabular}}}}%
    \put(0.63370314,0.59930845){\makebox(0,0)[lt]{\lineheight{-13.36629963}\smash{\begin{tabular}[t]{l}\scriptsize $+$\end{tabular}}}}%
    \put(0.52397034,0.59930846){\makebox(0,0)[t]{\lineheight{-13.36629963}\smash{\begin{tabular}[t]{c}\scriptsize $+$\end{tabular}}}}%
    \put(0.70992248,0.59930845){\makebox(0,0)[t]{\lineheight{-13.36629963}\smash{\begin{tabular}[t]{c}\scriptsize $\eta_\ind{p}\mc{L}_\mathrm{p}$\end{tabular}}}}%
    \put(0,0){\includegraphics[width=\unitlength,page=16]{pinn_structures.pdf}}%
    \put(0.58521245,0.32541198){\makebox(0,0)[t]{\lineheight{1.25}\smash{\begin{tabular}[t]{c}\scriptsize$\mm{\alpha}$\end{tabular}}}}%
    \put(0.90206273,0.49977321){\makebox(0,0)[t]{\lineheight{1.25}\smash{\begin{tabular}[t]{c}\scriptsize closed-form\end{tabular}}}}%
    \put(0.90206273,0.47122019){\makebox(0,0)[t]{\lineheight{1.25}\smash{\begin{tabular}[t]{c}\scriptsize differentiation\end{tabular}}}}%
    \put(0.90206273,1.04197213){\makebox(0,0)[t]{\lineheight{1.25}\smash{\begin{tabular}[t]{c}\scriptsize automatic\end{tabular}}}}%
    \put(0.90206273,1.01341911){\makebox(0,0)[t]{\lineheight{1.25}\smash{\begin{tabular}[t]{c}\scriptsize differentiation\end{tabular}}}}%
  \end{picture}%
\endgroup%

%% file: images/control_architecture.pdf_tex
\begingroup%
  \makeatletter%
  \providecommand\color[2][]{%
    \errmessage{(Inkscape) Color is used for the text in Inkscape, but the package 'color.sty' is not loaded}%
    \renewcommand\color[2][]{}%
  }%
  \providecommand\transparent[1]{%
    \errmessage{(Inkscape) Transparency is used (non-zero) for the text in Inkscape, but the package 'transparent.sty' is not loaded}%
    \renewcommand\transparent[1]{}%
  }%
  \providecommand\rotatebox[2]{#2}%
  \newcommand*\fsize{\dimexpr\f@size pt\relax}%
  \newcommand*\lineheight[1]{\fontsize{\fsize}{#1\fsize}\selectfont}%
  \ifx\svgwidth\undefined%
    \setlength{\unitlength}{232.35320564bp}%
    \ifx\svgscale\undefined%
      \relax%
    \else%
      \setlength{\unitlength}{\unitlength * \real{\svgscale}}%
    \fi%
  \else%
    \setlength{\unitlength}{\svgwidth}%
  \fi%
  \global\let\svgwidth\undefined%
  \global\let\svgscale\undefined%
  \makeatother%
  \begin{picture}(1,0.40063485)%
    \lineheight{1}%
    \setlength\tabcolsep{0pt}%
    \put(0,0){\includegraphics[width=\unitlength,page=1]{control_architecture.pdf}}%
    \put(0.96884314,0.23157862){\makebox(0,0)[t]{\smash{\begin{tabular}[t]{c}\scriptsize $\mm{q}$\end{tabular}}}}%
    \put(0.57031211,0.23157862){\makebox(0,0)[lt]{\smash{\begin{tabular}[t]{l}\scriptsize $\mm{p}$\end{tabular}}}}%
    \put(0.43630122,0.23158346){\makebox(0,0)[rt]{\smash{\begin{tabular}[t]{r}\scriptsize $\mm{u}$\end{tabular}}}}%
    \put(0.26670193,0.23158346){\makebox(0,0)[lt]{\smash{\begin{tabular}[t]{l}\scriptsize $\mm{u}^*_\ind{0}$\end{tabular}}}}%
    \put(0.39665434,0.31388936){\makebox(0,0)[lt]{\smash{\begin{tabular}[t]{l}\scriptsize $\mm{u}_\ind{PI}$\end{tabular}}}}%
    \put(0.16224894,0.1283924){\makebox(0,0)[rt]{\smash{\begin{tabular}[t]{r}\scriptsize $\mm{x}$\end{tabular}}}}%
    \put(0,0){\includegraphics[width=\unitlength,page=2]{control_architecture.pdf}}%
    \put(0.00730541,0.30098211){\makebox(0,0)[lt]{\smash{\begin{tabular}[t]{l}\scriptsize $\mm{q}_\ind{d0}$\end{tabular}}}}%
    \put(0,0){\includegraphics[width=\unitlength,page=3]{control_architecture.pdf}}%
    \put(0.00730541,0.16821584){\makebox(0,0)[lt]{\smash{\begin{tabular}[t]{l}\scriptsize $\mm{x}_{\ind{d}m}$\end{tabular}}}}%
    \put(0,0){\includegraphics[width=\unitlength,page=4]{control_architecture.pdf}}%
    \put(0.00730541,0.25540686){\makebox(0,0)[lt]{\smash{\begin{tabular}[t]{l}\scriptsize $\mm{x}_{\ind{d}1}$\end{tabular}}}}%
    \put(0,0){\includegraphics[width=\unitlength,page=5]{control_architecture.pdf}}%
    \put(0.50734737,0.20488688){\color[rgb]{0,0,0}\makebox(0,0)[t]{\smash{\begin{tabular}[t]{c}\scriptsize valves\end{tabular}}}}%
    \put(0.90867019,0.3398727){\color[rgb]{0,0,0}\makebox(0,0)[rt]{\smash{\begin{tabular}[t]{r}\scriptsize ASR\end{tabular}}}}%
    \put(0.11873754,0.31950732){\color[rgb]{0,0,0}\makebox(0,0)[t]{\smash{\begin{tabular}[t]{c}\scriptsize PI\end{tabular}}}}%
    \put(0.01506642,0.36339113){\makebox(0,0)[lt]{\smash{\begin{tabular}[t]{l}\scriptsize $-$\end{tabular}}}}%
    \put(0,0){\includegraphics[width=\unitlength,page=6]{control_architecture.pdf}}%
    \put(0.40781268,0.0259452){\makebox(0,0)[t]{\smash{\begin{tabular}[t]{c}\scriptsize $[\mm{I},\ind{diag}(\frac{\partial}{\partial t},\dots,\frac{\partial}{\partial t})]\transpose$\end{tabular}}}}%
    \put(0,0){\includegraphics[width=\unitlength,page=7]{control_architecture.pdf}}%
    \put(0.17387302,0.2208905){\makebox(0,0)[t]{\smash{\begin{tabular}[t]{c}\scriptsize NMPC\end{tabular}}}}%
    \put(0.17387303,0.18822045){\makebox(0,0)[t]{\smash{\begin{tabular}[t]{c}\scriptsize with PINN\end{tabular}}}}%
    \put(0.65362155,0.12687273){\makebox(0,0)[t]{\smash{\begin{tabular}[t]{c}\scriptsize five\end{tabular}}}}%
    \put(0.65362155,0.09420268){\makebox(0,0)[t]{\smash{\begin{tabular}[t]{c}\scriptsize DoF\end{tabular}}}}%
    \put(0,0){\includegraphics[width=\unitlength,page=8]{control_architecture.pdf}}%
    \put(0.30818564,0.32127859){\color[rgb]{0,0,0}\makebox(0,0)[t]{\smash{\begin{tabular}[t]{c}\scriptsize $\mm{g}_\ind{PI}(\cdot)$\end{tabular}}}}%
    \put(0,0){\includegraphics[width=\unitlength,page=9]{control_architecture.pdf}}%
    \put(0.15231188,0.34527276){\makebox(0,0)[lt]{\smash{\begin{tabular}[t]{l}\scriptsize $\Delta\mm{p}_\ind{PI}$\end{tabular}}}}%
  \end{picture}%
\endgroup%

%% file: images/test_bench_n_Act.pdf_tex
\begingroup%
  \makeatletter%
  \providecommand\color[2][]{%
    \errmessage{(Inkscape) Color is used for the text in Inkscape, but the package 'color.sty' is not loaded}%
    \renewcommand\color[2][]{}%
  }%
  \providecommand\transparent[1]{%
    \errmessage{(Inkscape) Transparency is used (non-zero) for the text in Inkscape, but the package 'transparent.sty' is not loaded}%
    \renewcommand\transparent[1]{}%
  }%
  \providecommand\rotatebox[2]{#2}%
  \newcommand*\fsize{\dimexpr\f@size pt\relax}%
  \newcommand*\lineheight[1]{\fontsize{\fsize}{#1\fsize}\selectfont}%
  \ifx\svgwidth\undefined%
    \setlength{\unitlength}{233.76843069bp}%
    \ifx\svgscale\undefined%
      \relax%
    \else%
      \setlength{\unitlength}{\unitlength * \real{\svgscale}}%
    \fi%
  \else%
    \setlength{\unitlength}{\svgwidth}%
  \fi%
  \global\let\svgwidth\undefined%
  \global\let\svgscale\undefined%
  \makeatother%
  \begin{picture}(1,0.60769171)%
    \lineheight{1}%
    \setlength\tabcolsep{0pt}%
    \put(0,0){\includegraphics[width=\unitlength,page=1]{test_bench_n_Act.pdf}}%
    \put(0.0440622,0.57535146){\makebox(0,0)[lt]{\lineheight{1.25}\smash{\begin{tabular}[t]{l}\scriptsize$\#n$\end{tabular}}}}%
    \put(0,0){\includegraphics[width=\unitlength,page=2]{test_bench_n_Act.pdf}}%
    \put(0.07063682,0.49810477){\makebox(0,0)[lt]{\lineheight{1.25}\smash{\begin{tabular}[t]{l}\scriptsize $\#2$\end{tabular}}}}%
    \put(0,0){\includegraphics[width=\unitlength,page=3]{test_bench_n_Act.pdf}}%
    \put(0.4713209,0.07145548){\makebox(0,0)[lt]{\lineheight{1.25}\smash{\begin{tabular}[t]{l}\scriptsize Ethernet\end{tabular}}}}%
    \put(0,0){\includegraphics[width=\unitlength,page=4]{test_bench_n_Act.pdf}}%
    \put(0.71972362,0.07145548){\makebox(0,0)[lt]{\lineheight{1.25}\smash{\begin{tabular}[t]{l}\scriptsize EtherLab\end{tabular}}}}%
    \put(0.14357342,0.01392376){\makebox(0,0)[lt]{\lineheight{1.25}\smash{\begin{tabular}[t]{l}\scriptsize{supply unit}\end{tabular}}}}%
    \put(0,0){\includegraphics[width=\unitlength,page=5]{test_bench_n_Act.pdf}}%
    \put(0.09506658,0.417325){\makebox(0,0)[lt]{\lineheight{1.25}\smash{\begin{tabular}[t]{l}\scriptsize $\#1$\end{tabular}}}}%
    \put(0,0){\includegraphics[width=\unitlength,page=6]{test_bench_n_Act.pdf}}%
    \put(0.76554435,0.19950306){\makebox(0,0)[lt]{\lineheight{1.25}\smash{\begin{tabular}[t]{l}\scriptsize{actuator}\end{tabular}}}}%
    \put(0.39210921,0.19950299){\makebox(0,0)[lt]{\lineheight{1.25}\smash{\begin{tabular}[t]{l}\scriptsize{valves}\end{tabular}}}}%
    \put(0,0){\includegraphics[width=\unitlength,page=7]{test_bench_n_Act.pdf}}%
    \put(0.33767889,0.23032626){\color[rgb]{0,0.31372549,0.60784314}\makebox(0,0)[lt]{\lineheight{1.25}\smash{\begin{tabular}[t]{l}\scriptsize$2$\end{tabular}}}}%
    \put(0.35180987,0.33329005){\color[rgb]{0,0.31372549,0.60784314}\makebox(0,0)[lt]{\lineheight{1.25}\smash{\begin{tabular}[t]{l}\scriptsize$3$\end{tabular}}}}%
    \put(0.29283073,0.33329005){\color[rgb]{0,0.31372549,0.60784314}\makebox(0,0)[lt]{\lineheight{1.25}\smash{\begin{tabular}[t]{l}\scriptsize$1$\end{tabular}}}}%
    \put(0,0){\includegraphics[width=\unitlength,page=8]{test_bench_n_Act.pdf}}%
    \put(0.41229498,0.25818402){\color[rgb]{0,0.31372549,0.60784314}\makebox(0,0)[lt]{\lineheight{1.25}\smash{\begin{tabular}[t]{l}\tiny P\end{tabular}}}}%
    \put(0,0){\includegraphics[width=\unitlength,page=9]{test_bench_n_Act.pdf}}%
    \put(0.17054388,0.23032626){\color[rgb]{0,0.31372549,0.60784314}\makebox(0,0)[lt]{\lineheight{1.25}\smash{\begin{tabular}[t]{l}\scriptsize$2$\end{tabular}}}}%
    \put(0.18467486,0.33329005){\color[rgb]{0,0.31372549,0.60784314}\makebox(0,0)[lt]{\lineheight{1.25}\smash{\begin{tabular}[t]{l}\scriptsize$3$\end{tabular}}}}%
    \put(0.12569572,0.33329005){\color[rgb]{0,0.31372549,0.60784314}\makebox(0,0)[lt]{\lineheight{1.25}\smash{\begin{tabular}[t]{l}\scriptsize$1$\end{tabular}}}}%
    \put(0,0){\includegraphics[width=\unitlength,page=10]{test_bench_n_Act.pdf}}%
    \put(0.24515997,0.25818402){\color[rgb]{0,0.31372549,0.60784314}\makebox(0,0)[lt]{\lineheight{1.25}\smash{\begin{tabular}[t]{l}\tiny P\end{tabular}}}}%
    \put(0,0){\includegraphics[width=\unitlength,page=11]{test_bench_n_Act.pdf}}%
    \put(0.91000653,0.12009963){\makebox(0,0)[lt]{\lineheight{1.25}\smash{\begin{tabular}[t]{l}\scriptsize$\mm{q}$\end{tabular}}}}%
    \put(0,0){\includegraphics[width=\unitlength,page=12]{test_bench_n_Act.pdf}}%
    \put(0.97812912,0.12009971){\makebox(0,0)[lt]{\lineheight{1.25}\smash{\begin{tabular}[t]{l}\scriptsize$\mm{p}$\end{tabular}}}}%
    \put(0.45179434,0.39754404){\makebox(0,0)[lt]{\lineheight{1.25}\smash{\begin{tabular}[t]{l}\scriptsize$\mm{p}_{\mathrm{d}1}$\end{tabular}}}}%
    \put(0,0){\includegraphics[width=\unitlength,page=13]{test_bench_n_Act.pdf}}%
    \put(0.3799689,0.05031816){\makebox(0,0)[t]{\lineheight{1.25}\smash{\begin{tabular}[t]{c}\scriptsize Dev-PC\end{tabular}}}}%
    \put(0,0){\includegraphics[width=\unitlength,page=14]{test_bench_n_Act.pdf}}%
    \put(0.60590338,0.04960492){\makebox(0,0)[lt]{\lineheight{1.25}\smash{\begin{tabular}[t]{l}\scriptsize RT-PC\end{tabular}}}}%
    \put(0.8641776,0.04767565){\makebox(0,0)[lt]{\lineheight{1.25}\smash{\begin{tabular}[t]{l}\scriptsize EtherCAT\end{tabular}}}}%
    \put(0.43254446,0.47952846){\makebox(0,0)[lt]{\lineheight{1.25}\smash{\begin{tabular}[t]{l}\scriptsize$\mm{p}_{\mathrm{d}2}$\end{tabular}}}}%
    \put(0.41329458,0.55253836){\makebox(0,0)[lt]{\lineheight{1.25}\smash{\begin{tabular}[t]{l}\scriptsize$\mm{p}_{\mathrm{d}n}$\end{tabular}}}}%
  \end{picture}%
\endgroup%

%% file: images/ident_results.pdf_tex
\begingroup%
  \makeatletter%
  \providecommand\color[2][]{%
    \errmessage{(Inkscape) Color is used for the text in Inkscape, but the package 'color.sty' is not loaded}%
    \renewcommand\color[2][]{}%
  }%
  \providecommand\transparent[1]{%
    \errmessage{(Inkscape) Transparency is used (non-zero) for the text in Inkscape, but the package 'transparent.sty' is not loaded}%
    \renewcommand\transparent[1]{}%
  }%
  \providecommand\rotatebox[2]{#2}%
  \newcommand*\fsize{\dimexpr\f@size pt\relax}%
  \newcommand*\lineheight[1]{\fontsize{\fsize}{#1\fsize}\selectfont}%
  \ifx\svgwidth\undefined%
    \setlength{\unitlength}{231.70005979bp}%
    \ifx\svgscale\undefined%
      \relax%
    \else%
      \setlength{\unitlength}{\unitlength * \real{\svgscale}}%
    \fi%
  \else%
    \setlength{\unitlength}{\svgwidth}%
  \fi%
  \global\let\svgwidth\undefined%
  \global\let\svgscale\undefined%
  \makeatother%
  \begin{picture}(1,0.57854581)%
    \lineheight{1}%
    \setlength\tabcolsep{0pt}%
    \put(0,0){\includegraphics[width=\unitlength,page=1]{ident_results.pdf}}%
    \put(0.4970469,0.04052653){\makebox(0,0)[lt]{\lineheight{1.25}\smash{\begin{tabular}[t]{l}\scriptsize(b)\end{tabular}}}}%
    \put(0.03318853,0.04052653){\makebox(0,0)[lt]{\lineheight{1.25}\smash{\begin{tabular}[t]{l}\scriptsize(a)\end{tabular}}}}%
  \end{picture}%
\endgroup%

%% file: images/HPO_results.pdf_tex
\begingroup%
  \makeatletter%
  \providecommand\color[2][]{%
    \errmessage{(Inkscape) Color is used for the text in Inkscape, but the package 'color.sty' is not loaded}%
    \renewcommand\color[2][]{}%
  }%
  \providecommand\transparent[1]{%
    \errmessage{(Inkscape) Transparency is used (non-zero) for the text in Inkscape, but the package 'transparent.sty' is not loaded}%
    \renewcommand\transparent[1]{}%
  }%
  \providecommand\rotatebox[2]{#2}%
  \newcommand*\fsize{\dimexpr\f@size pt\relax}%
  \newcommand*\lineheight[1]{\fontsize{\fsize}{#1\fsize}\selectfont}%
  \ifx\svgwidth\undefined%
    \setlength{\unitlength}{231.54664047bp}%
    \ifx\svgscale\undefined%
      \relax%
    \else%
      \setlength{\unitlength}{\unitlength * \real{\svgscale}}%
    \fi%
  \else%
    \setlength{\unitlength}{\svgwidth}%
  \fi%
  \global\let\svgwidth\undefined%
  \global\let\svgscale\undefined%
  \makeatother%
  \begin{picture}(1,1.26256137)%
    \lineheight{1}%
    \setlength\tabcolsep{0pt}%
    \put(0,0){\includegraphics[width=\unitlength,page=1]{HPO_results.pdf}}%
    \put(0.02662568,1.24540327){\makebox(0,0)[lt]{\lineheight{1.25}\smash{\begin{tabular}[t]{l}\scriptsize (a) \highlightred{DD-}PINN\end{tabular}}}}%
    \put(0.02662568,0.60524176){\makebox(0,0)[lt]{\lineheight{1.25}\smash{\begin{tabular}[t]{l}\scriptsize (b) RNN\end{tabular}}}}%
  \end{picture}%
\endgroup%

%% file: literatur.bib
@misc{Alessi.2024,
 author = {Alessi, Carlo and Agabiti, Camilla and Caradonna, Daniele and Laschi, Cecilia and Renda, Federico and Falotico, Egidio},
 date = {2024},
 title = {Rod models in continuum and soft robot control: a review},
 publisher = {arXiv},
 doi = {10.48550/arXiv.2407.05886}
}

@article{Andersson.2019,
 author = {Andersson, Joel A. E. and Gillis, Joris and Horn, Greg and Rawlings, James B. and Diehl, Moritz},
 year = {2019},
 title = {{C}as{A}{D}i: a software framework for nonlinear optimization and optimal control},
 pages = {1--36},
 volume = {11},
 number = {1},
 issn = {1867-2949},
 journal = {Mathematical Programming Computation},
 doi = {10.1007/s12532-018-0139-4}
}

@article{Antonelo.2024,
 author = {Antonelo, Eric Aislan and Camponogara, Eduardo and Seman, Laio Oriel and Jordanou, Jean Panaioti and de Souza, Eduardo Rehbein and H{\"u}bner, Jomi Fred},
 year = {2024},
 title = {Physics-informed neural nets for control of dynamical systems},
 pages = {127419},
 volume = {579},
 issn = {09252312},
 journal = {Neurocomputing},
 doi = {10.1016/j.neucom.2024.127419}
}

@article{Armanini.2023,
 author = {Armanini, Costanza and Boyer, Fr{\'e}d{\'e}ric and Mathew, Anup Teejo and Duriez, Christian and Renda, Federico},
 year = {2023},
 title = {Soft Robots Modeling: A Structured Overview},
 pages = {1728--1748},
 volume = {39},
 number = {3},
 issn = {1552-3098},
 journal = {IEEE Transactions on Robotics},
 doi = {10.1109/TRO.2022.3231360}
}

@article{Azad.2014,
 author = {Azad, Morteza and Featherstone, Roy},
 year = {2014},
 title = {A New Nonlinear Model of Contact Normal Force},
 pages = {736--739},
 volume = {30},
 number = {3},
 issn = {1552-3098},
 journal = {IEEE Transactions on Robotics},
 doi = {10.1109/TRO.2013.2293833}
}

@article{Beaber.2024,
 author = {Beaber, Sameh I. and Liu, Zhen and Sun, Ye},
 year = {2024},
 title = {Physics-Guided Deep Learning Enabled Surrogate Modeling for Pneumatic Soft Robots},
 pages = {11441--11448},
 volume = {9},
 number = {12},
 issn = {2377-3766},
 journal = {IEEE Robotics and Automation Letters},
 doi = {10.1109/LRA.2024.3490258}
}

@inproceedings{Bensch.2024,
 author = {Bensch, Martin and Job, Tim-David and Habich, Tim-Lukas and Seel, Thomas and Schappler, Moritz},
 title = {Physics-Informed Neural Networks for Continuum Robots: Towards Fast Approximation of Static {C}osserat Rod Theory},
 pages = {17293--17299},
 publisher = {IEEE},
 isbn = {979-8-3503-8457-4},
 booktitle = {2024 IEEE International Conference on Robotics and Automation},
 year = {2024},
 doi = {10.1109/ICRA57147.2024.10610742}
}

@inproceedings{Best.2015,
 author = {Best, Charles M. and Wilson, Joshua P. and Killpack, Marc D.},
 title = {Control of a pneumatically actuated, fully inflatable, fabric-based, humanoid robot},
 pages = {1133--1140},
 publisher = {IEEE},
 isbn = {978-1-4799-6885-5},
 booktitle = {2015 IEEE-RAS 15th International Conference on Humanoid Robots (Humanoids)},
 year = {2015},
 doi = {10.1109/HUMANOIDS.2015.7363495}
}

@article{Braganza.2007,
 author = {Braganza, D. and Dawson, D. M. and Walker, I. D. and Nath, N.},
 year = {2007},
 title = {A Neural Network Controller for Continuum Robots},
 pages = {1270--1277},
 volume = {23},
 number = {6},
 issn = {1941-0468},
 journal = {IEEE Transactions on Robotics},
 doi = {10.1109/TRO.2007.906248}
}

@article{Bruder.2021,
 author = {Bruder, Daniel and Fu, Xun and Gillespie, R. Brent and Remy, C. David and Vasudevan, Ram},
 year = {2021},
 title = {Data-Driven Control of Soft Robots Using {K}oopman Operator Theory},
 pages = {948--961},
 volume = {37},
 number = {3},
 issn = {1941-0468},
 journal = {IEEE Transactions on Robotics},
 doi = {10.1109/TRO.2020.3038693}
}

@article{Centurelli.2022,
 author = {Centurelli, Andrea and Arleo, Luca and Rizzo, Alessandro and Tolu, Silvia and Laschi, Cecilia and Falotico, Egidio},
 year = {2022},
 title = {Closed-Loop Dynamic Control of a Soft Manipulator Using Deep Reinforcement Learning},
 pages = {4741--4748},
 volume = {7},
 number = {2},
 issn = {2377-3766},
 journal = {IEEE Robotics and Automation Letters},
 doi = {10.1109/LRA.2022.3146903}
}

@article{Chen.2018,
 author = {Chen, Ricky T. Q. and Rubanova, Yulia and Bettencourt, Jesse and Duvenaud, David K.},
 year = {2018},
 title = {Neural Ordinary Differential Equations},
 volume = {31},
 journal = {Advances in Neural Information Processing Systems}
}

@inproceedings{Cheney.2024,
 author = {Cheney, Daniel G. and Killpack, Marc D.},
 title = {Mo{L}{D}y: Open-Source Library for Data-Based Modeling and Nonlinear Model Predictive Control of Soft Robots},
 pages = {958--964},
 isbn = {979-8-3503-8181-8},
 booktitle = {IEEE International Conference on Soft Robotics},
 year = {2024},
 doi = {10.1109/ROBOSOFT60065.2024.10522031}
}

@article{Chhatoi.2023,
 author = {Chhatoi, Saroj Prasad and Pierallini, Michele and Angelini, Franco and Mastalli, Carlos and Garabini, Manolo},
 year = {2023},
 title = {Optimal Control for Articulated Soft Robots},
 pages = {3671--3685},
 volume = {39},
 number = {5},
 issn = {1552-3098},
 journal = {IEEE Transactions on Robotics},
 doi = {10.1109/TRO.2023.3288837}
}

@article{Cho.2014,
 author = {Cho, Kyunghyun and {van Merrienboer}, Bart and Bahdanau, Dzmitry and Bengio, Yoshua},
 year = {2014},
 title = {On the Properties of Neural Machine Translation: Encoder-Decoder Approaches},
 url = {https://arxiv.org/pdf/1409.1259.pdf},
 doi = {10.48550/arXiv.1409.1259}
}

@article{Coevoet.2017,
 author = {Coevoet, E. and Morales-Bieze, T. and Largilliere, F. and Zhang, Z. and Thieffry, M. and Sanz-Lopez, M. and Carrez, B. and Marchal, D. and Goury, O. and Dequidt, J. and Duriez, C.},
 year = {2017},
 title = {Software toolkit for modeling, simulation, and control of soft robots},
 pages = {1208--1224},
 volume = {31},
 number = {22},
 issn = {0169-1864},
 journal = {Advanced Robotics},
 doi = {10.1080/01691864.2017.1395362}
}

@article{Cuomo.14.01.2022,
 author = {Cuomo, Salvatore and {Di Cola}, Vincenzo Schiano and Giampaolo, Fabio and Rozza, Gianluigi and Raissi, Maziar and Piccialli, Francesco},
 year = {2022},
 title = {Scientific Machine Learning through Physics-Informed Neural Networks:  Where we are and What's next},
 url = {https://arxiv.org/pdf/2201.05624}
}

@incollection{DellaSantina.2020,
 author = {{Della Santina}, Cosimo and Catalano, Manuel G. and Bicchi, Antonio},
 title = {Soft Robots},
 pages = {1--15},
 publisher = {{Springer Berlin Heidelberg}},
 isbn = {978-3-642-41610-1},
 editor = {Ang, Marcelo H. and Khatib, Oussama and Siciliano, Bruno},
 booktitle = {Encyclopedia of Robotics},
 year = {2020},
 address = {Berlin, Heidelberg},
 doi = {10.1007/978-3-642-41610-1{\textunderscore }146-2}
}

@article{DellaSantina.2023,
 author = {{Della Santina}, Cosimo and Duriez, Christian and Rus, Daniela},
 year = {2023},
 title = {Model-Based Control of Soft Robots: A Survey of the State of the Art and Open Challenges},
 pages = {30--65},
 volume = {43},
 number = {3},
 issn = {1066-033X},
 journal = {IEEE Control Systems},
 doi = {10.1109/MCS.2023.3253419}
}

@article{Falotico.2024,
 author = {{E. Falotico et al.}},
 year = {2024},
 title = {Learning Controllers for Continuum Soft Manipulators: Impact of Modeling and Looming Challenges},
 issn = {2640-4567},
 journal = {Advanced Intelligent Systems},
 doi = {10.1002/aisy.202400344}
}

@article{Gao.2024,
 author = {Gao, Junpeng and Michelis, Mike Y. and Spielberg, Andrew and Katzschmann, Robert K.},
 year = {2024},
 title = {Sim-to-Real of Soft Robots with Learned Residual Physics},
 pages = {1--8},
 issn = {2377-3766},
 journal = {IEEE Robotics and Automation Letters},
 doi = {10.1109/LRA.2024.3446287}
}

@article{GeorgeThuruthel.2018,
 author = {Thuruthel, Thomas George and Ansari, Yasmin and Falotico, Egidio and Laschi, Cecilia},
 year = {2018},
 title = {Control Strategies for Soft Robotic Manipulators: A Survey},
 pages = {149--163},
 volume = {5},
 number = {2},
 journal = {Soft {R}obotics},
 doi = {10.1089/soro.2017.0007}
}

@inproceedings{Gillespie.2018,
 author = {Gillespie, Morgan T. and Best, Charles M. and Townsend, Eric C. and Wingate, David and Killpack, Marc D.},
 title = {Learning nonlinear dynamic models of soft robots for model predictive control with neural networks},
 pages = {39--45},
 isbn = {978-1-5386-4516-1},
 booktitle = {IEEE International Conference on Soft Robotics},
 year = {2018},
 doi = {10.1109/ROBOSOFT.2018.8404894}
}

@inproceedings{Habich.2023,
 author = {Habich, Tim-Lukas and Kleinjohann, Sarah and Schappler, Moritz},
 title = {Learning-based Position and Stiffness Feedforward Control of Antagonistic Soft Pneumatic Actuators using {G}aussian Processes},
 pages = {1--7},
 publisher = {IEEE},
 isbn = {979-8-3503-3222-3},
 booktitle = {2023 IEEE International Conference on Soft Robotics},
 year = {2023},
 doi = {10.1109/ROBOSOFT55895.2023.10122057}
}

@article{Habich.2024,
 author = {Habich, Tim-Lukas and Haack, Jonas and Belhadj, Mehdi and Lehmann, Dustin and Seel, Thomas and Schappler, Moritz},
 year = {2024},
 title = {{S}{P}{O}{N}{G}{E}: Open-Source Designs of Modular Articulated Soft Robots},
 pages = {5346--5353},
 volume = {9},
 number = {6},
 issn = {2377-3766},
 journal = {IEEE Robotics and Automation Letters},
 doi = {10.1109/LRA.2024.3388855}
}

@article{HabichData.2025,
 author = {Habich, Tim-Lukas and Mohammad, Aran and Ehlers, Simon F. G. and Bensch, Martin and Seel, Thomas and Schappler, Moritz},
 year = {2025},
 title = {Experimental Data of an Articulated Soft Robot with Variable Payloads and Base Orientations},
 url = {https://doi.org/10.15488/19708},
 journal = {Institutional Repository of Leibniz University Hannover}
}

@book{Hairer.1996,
 author = {Hairer, Ernst and Wanner, Gerhard},
 year = {1996},
 title = {Solving Ordinary Differential Equations II},
 address = {Berlin, Heidelberg},
 volume = {14},
 publisher = {{Springer Berlin Heidelberg}},
 isbn = {978-3-642-05220-0},
 doi = {10.1007/978-3-642-05221-7}
}

@article{Hoffmann.2023,
 author = {Hoffmann, Kathrin and Trapp, Christian and Hildebrandt, Alexander and Sawodny, Oliver},
 year = {2023},
 title = {Nonlinear Model-Based Control of a Pneumatically Driven Robot},
 pages = {8770--8775},
 volume = {56},
 number = {2},
 issn = {24058963},
 journal = {IFAC-PapersOnLine},
 doi = {10.1016/j.ifacol.2023.10.062}
}

@article{Huang.2024,
 author = {Huang, Xinjia and Rong, Yu and Gu, GuoYing},
 year = {2024},
 title = {High-Precision Dynamic Control of Soft Robots With the Physics-Learning Hybrid Modeling Approach},
 pages = {1--12},
 issn = {1083-4435},
 journal = {IEEE/ASME Transactions on Mechatronics},
 doi = {10.1109/TMECH.2024.3403151}
}

@article{Hunt.1975,
 author = {Hunt, K. H. and Crossley, F. R. E.},
 year = {1975},
 title = {Coefficient of Restitution Interpreted as Damping in Vibroimpact},
 pages = {440--445},
 volume = {42},
 number = {2},
 issn = {0021-8936},
 journal = {Journal of Applied Mechanics},
 doi = {10.1115/1.3423596}
}

@article{Hyatt.2019,
 author = {Hyatt, Phillip and Wingate, David and Killpack, Marc D.},
 year = {2019},
 title = {Model-Based Control of Soft Actuators Using Learned Non-linear Discrete-Time Models},
 pages = {22},
 volume = {6},
 journal = {Frontiers in {R}obotics and AI},
 doi = {10.3389/frobt.2019.00022}
}

@article{Hyatt.2020,
 author = {Hyatt, Phillip and Killpack, Marc D.},
 year = {2020},
 title = {Real-Time Nonlinear Model Predictive Control of Robots Using a Graphics Processing Unit},
 pages = {1468--1475},
 volume = {5},
 number = {2},
 issn = {2377-3766},
 journal = {IEEE Robotics and Automation Letters},
 doi = {10.1109/LRA.2020.2965393}
}

@article{Jeon.2025,
 author = {Jeon, Se Hwan and Hong, Seungwoo and Lee, Ho Jae and Khazoom, Charles and Kim, Sangbae},
 year = {2025},
 title = {{C}us{A}{D}i: A {G}{P}{U} Parallelization Framework for Symbolic Expressions and Optimal Control},
 pages = {899--906},
 volume = {10},
 number = {2},
 issn = {2377-3766},
 journal = {IEEE Robotics and Automation Letters},
 doi = {10.1109/LRA.2024.3512254}
}

@article{Jiahao.2024,
 author = {Jiahao, Tom Z. and Adolf, Ryan and Sung, Cynthia and Hsieh, M. Ani},
 year = {2024},
 title = {Knowledge-based Neural Ordinary Differential Equations for {C}osserat Rod-based Soft Robots},
 url = {https://arxiv.org/pdf/2408.07776},
 journal = {arXiv},
 doi = {10.48550/arXiv.2408.07776}
}

@article{Johnson.2021,
 author = {Johnson, Curtis C. and Quackenbush, Tyler and Sorensen, Taylor and Wingate, David and Killpack, Marc D.},
 year = {2021},
 title = {Using First Principles for Deep Learning and Model-Based Control of Soft Robots},
 pages = {654398},
 volume = {8},
 journal = {Frontiers in {R}obotics and AI},
 doi = {10.3389/frobt.2021.654398}
}

@article{Karniadakis.2021,
 author = {Karniadakis, George Em and Kevrekidis, Ioannis G. and Lu, Lu and Perdikaris, Paris and Wang, Sifan and Yang, Liu},
 year = {2021},
 title = {Physics-informed machine learning},
 pages = {422--440},
 volume = {3},
 number = {6},
 journal = {Nature Reviews Physics},
 doi = {10.1038/s42254-021-00314-5}
}

@inproceedings{Kasaei.2023,
 author = {Kasaei, Mohammadreza and Babarahmati, Keyhan Kouhkiloui and Li, Zhibin and Khadem, Mohsen},
 title = {Data-efficient Non-parametric Modelling and Control of an Extensible Soft Manipulator},
 pages = {2641--2647},
 publisher = {IEEE},
 isbn = {979-8-3503-2365-8},
 booktitle = {2023 IEEE International Conference on Robotics and Automation},
 year = {2023},
 doi = {10.1109/ICRA48891.2023.10161275}
}

@inproceedings{Krauss.2024,
 author = {Krauss, Henrik and Habich, Tim-Lukas and Bartholdt, Max and Seel, Thomas and Schappler, Moritz},
 title = {Domain-Decoupled Physics-informed Neural Networks with Closed-Form Gradients for Fast Model Learning of Dynamical Systems},
 pages = {55--66},
 publisher = {SciTePress},
 isbn = {978-989-758-717-7},
 booktitle = {International Conference on Informatics in Control, Automation and Robotics},
 year = {2024},
 doi = {10.5220/0012935200003822}
}

@inproceedings{Lahariya.2022,
 author = {Lahariya, Manu and Innes, Craig and Develder, Chris and Ramamoorthy, Subramanian},
 title = {Learning physics-informed simulation models for soft robotic manipulation: A case study with dielectric elastomer actuators},
 pages = {11031--11038},
 publisher = {IEEE},
 isbn = {978-1-6654-7927-1},
 booktitle = {2022 IEEE/RSJ International Conference on Intelligent Robots and Systems},
 year = {2022},
 doi = {10.1109/IROS47612.2022.9981373}
}

@article{Laschi.2023,
 author = {Laschi, Cecilia and Thuruthel, Thomas George and Lida, Fumiya and Merzouki, Rochdi and Falotico, Egidio},
 year = {2023},
 title = {Learning-Based Control Strategies for Soft Robots: Theory, Achievements, and Future Challenges},
 pages = {100--113},
 volume = {43},
 number = {3},
 issn = {1066-033X},
 journal = {IEEE Control Systems},
 doi = {10.1109/MCS.2023.3253421}
}

@article{Li.,
 author = {{L. Li et al.}},
 year = {2020},
 title = {A System for Massively Parallel Hyperparameter Tuning},
 url = {https://arxiv.org/pdf/1810.05934},
 journal = {Conference on Machine Learning and Systems}
}

@article{Lipton.29.05.2015,
 author = {Lipton, Zachary C. and Berkowitz, John and Elkan, Charles},
 year = {2015},
 title = {A Critical Review of Recurrent Neural Networks for Sequence Learning},
 url = {https://arxiv.org/pdf/1506.00019}
}

@article{Liu.2024,
 author = {Liu, Jingyue and Borja, Pablo and {Della Santina}, Cosimo},
 year = {2024},
 title = {Physics--Informed Neural Networks to Model and Control Robots: A Theoretical and Experimental Investigation},
 issn = {2640-4567},
 journal = {Advanced Intelligent Systems},
 doi = {10.1002/aisy.202300385}
}

@article{Lou.2024,
 author = {Lou, Gaoming and Wang, Chuang and Xu, Zefeng and Liang, Jiaqiao and Zhou, Yitong},
 year = {2024},
 title = {Controlling Soft Robotic Arms Using Hybrid Modelling and Reinforcement Learning},
 pages = {7070--7077},
 volume = {9},
 number = {8},
 issn = {2377-3766},
 journal = {IEEE Robotics and Automation Letters},
 doi = {10.1109/LRA.2024.3418312}
}

@article{Luong.2021,
 author = {{T. Luong et al.}},
 year = {2021},
 title = {Long Short Term Memory Model Based Position-Stiffness Control of Antagonistically Driven Twisted-Coiled Polymer Actuators Using Model Predictive Control},
 pages = {4141--4148},
 volume = {6},
 number = {2},
 journal = {IEEE Robotics and Automation Letters},
 doi = {10.1109/LRA.2021.3068905}
}

@article{Lutter.05.10.2021,
 author = {Lutter, Michael and Peters, Jan},
 year = {2021},
 title = {Combining Physics and Deep Learning to learn Continuous-Time Dynamics  Models},
 url = {https://arxiv.org/pdf/2110.01894}
}

@inproceedings{Lutter.2019,
 author = {Lutter, Michael and Listmann, Kim and Peters, Jan},
 title = {Deep {L}agrangian Networks for end-to-end learning of energy-based control for under-actuated systems},
 pages = {7718--7725},
 isbn = {978-1-7281-4004-9},
 booktitle = {2019 IEEE/RSJ International Conference on Intelligent Robots and Systems},
 year = {2019},
 doi = {10.1109/IROS40897.2019.8968268}
}

@inproceedings{Mendenhall.2024,
 author = {Mendenhall, Carly A. and Hardan, Jonathan and Chiang, Trysta D. and Blumenschein, Laura H. and Tepole, Adrian Buganza},
 title = {Physics-Informed Neural Network for Scalable Soft Multi-Actuator Systems},
 pages = {716--721},
 isbn = {979-8-3503-8181-8},
 booktitle = {IEEE International Conference on Soft Robotics},
 year = {2024},
 doi = {10.1109/ROBOSOFT60065.2024.10522053}
}

@book{Nelles.2020,
 author = {Nelles, Oliver},
 year = {2020},
 title = {Nonlinear System Identification: From Classical Approaches to Neural Networks, Fuzzy Models, and Gaussian Processes},
 edition = {2nd ed. 2020},
 publisher = {{Springer Cham}},
 isbn = {978-3-030-47438-6},
 doi = {10.1007/978-3-030-47439-3}
}

@article{NEUMANN.2013,
 author = {Neumann, Klaus and Rolf, Matthias and Steil, Jochen Jakob},
 year = {2013},
 title = {RELIABLE INTEGRATION OF CONTINUOUS CONSTRAINTS INTO EXTREME LEARNING MACHINES},
 pages = {35--50},
 volume = {21},
 number = {supp02},
 issn = {0218-4885},
 journal = {International Journal of Uncertainty, Fuzziness and Knowledge-Based Systems},
 doi = {10.1142/S021848851340014X}
}

@inproceedings{Nghiem.2023,
 author = {{T. X. Nghiem et al.}},
 title = {Physics-Informed Machine Learning for Modeling and Control of Dynamical Systems},
 pages = {3735--3750},
 publisher = {IEEE},
 isbn = {979-8-3503-2806-6},
 booktitle = {2023 American Control Conference},
 year = {2023},
 doi = {10.23919/ACC55779.2023.10155901}
}

@inproceedings{NguyenTuong.2010,
 author = {Nguyen-Tuong, Duy and Peters, Jan},
 title = {Using model knowledge for learning inverse dynamics},
 pages = {2677--2682},
 publisher = {IEEE},
 isbn = {978-1-4244-5038-1},
 booktitle = {2010 IEEE International Conference on Robotics and Automation},
 year = {2010},
 doi = {10.1109/ROBOT.2010.5509858}
}

@article{Nicodemus.2022,
 author = {Nicodemus, Jonas and Kneifl, Jonas and Fehr, J{\"o}rg and Unger, Benjamin},
 year = {2022},
 title = {Physics-informed Neural Networks-based Model Predictive Control for Multi-link Manipulators},
 pages = {331--336},
 volume = {55},
 number = {20},
 issn = {24058963},
 journal = {IFAC-PapersOnLine},
 doi = {10.1016/j.ifacol.2022.09.117}
}

@article{Paszke.2019,
 author = {{Paszke et al.}, Adam},
 year = {2019},
 title = {{P}y{T}orch: An Imperative Style, High-Performance Deep Learning Library},
 volume = {32},
 journal = {Advances in Neural Information Processing Systems},
 doi = {10.48550/arXiv.1912.01703}
}

@article{Raissi.2019,
 author = {Raissi, M. and Perdikaris, P. and Karniadakis, G. E.},
 year = {2019},
 title = {Physics-informed neural networks: A deep learning framework for solving forward and inverse problems involving nonlinear partial differential equations},
 pages = {686--707},
 volume = {378},
 issn = {00219991},
 journal = {Journal of Computational Physics},
 doi = {10.1016/j.jcp.2018.10.045}
}

@article{Reinhart.2017,
 author = {Reinhart, Ren{\'e} Felix and Shareef, Zeeshan and Steil, Jochen Jakob},
 year = {2017},
 title = {Hybrid Analytical and Data-Driven Modeling for Feed-Forward Robot Control},
 volume = {17},
 number = {2},
 journal = {Sensors},
 doi = {10.3390/s17020311}
}

@inproceedings{Runge.2017,
 author = {Runge, Gundula and Wiese, Mats and Raatz, Annika},
 title = {F{E}{M}-based training of artificial neural networks for modular soft robots},
 pages = {385--392},
 publisher = {IEEE},
 isbn = {978-1-5386-3742-5},
 booktitle = {2017 IEEE International Conference on Robotics and Biomimetics},
 year = {2017},
 address = {Piscataway, NJ},
 doi = {10.1109/ROBIO.2017.8324448}
}

@article{Rus.2015,
 author = {Rus, Daniela and Tolley, Michael T.},
 year = {2015},
 title = {Design, fabrication and control of soft robots},
 pages = {467--475},
 journal = {Nature},
 doi = {10.1038/nature14543}
}

@article{Schafke.2024,
 author = {Sch{\"a}fke, Hendrik and Habich, Tim-Lukas and Muhmann, Christian and Ehlers, Simon F. G. and Seel, Thomas and Schappler, Moritz},
 year = {2024},
 title = {Learning-Based Nonlinear Model Predictive Control of Articulated Soft Robots Using Recurrent Neural Networks},
 pages = {11609--11616},
 volume = {9},
 number = {12},
 issn = {2377-3766},
 journal = {IEEE Robotics and Automation Letters},
 doi = {10.1109/LRA.2024.3495579}
}

@inproceedings{Schussler.2019,
 author = {Sch{\"u}ssler, Max and Munker, Tobias and Nelles, Oliver},
 title = {Deep Recurrent Neural Networks for Nonlinear System Identification},
 pages = {448--454},
 publisher = {IEEE},
 isbn = {978-1-7281-2485-8},
 booktitle = {2019 IEEE Symposium Series on Computational Intelligence},
 year = {2019},
 doi = {10.1109/SSCI44817.2019.9003133}
}

@article{Sun.2022,
 author = {Sun, Wentao and Akashi, Nozomi and Kuniyoshi, Yasuo and Nakajima, Kohei},
 year = {2022},
 title = {Physics-Informed Recurrent Neural Networks for Soft Pneumatic Actuators},
 pages = {6862--6869},
 volume = {7},
 number = {3},
 journal = {IEEE Robotics and Automation Letters},
 doi = {10.1109/LRA.2022.3178496}
}

@article{Tariverdi.2021,
 author = {{A. Tariverdi et al.}},
 year = {2021},
 title = {A Recurrent Neural-Network-Based Real-Time Dynamic Model for Soft Continuum Manipulators},
 pages = {631303},
 volume = {8},
 journal = {Frontiers in {R}obotics and AI},
 doi = {10.3389/frobt.2021.631303}
}

@article{Thuruthel.2017,
 author = {Thuruthel, Thomas George and Falotico, Egidio and Renda, Federico and Laschi, Cecilia},
 year = {2017},
 title = {Learning dynamic models for open loop predictive control of soft robotic manipulators},
 pages = {066003},
 volume = {12},
 number = {6},
 journal = {Bioinspiration {\&} {B}iomimetics},
 doi = {10.1088/1748-3190/aa839f}
}

@article{Thuruthel.2019,
 author = {Thuruthel, Thomas George and Falotico, Egidio and Renda, Federico and Laschi, Cecilia},
 year = {2019},
 title = {Model-Based Reinforcement Learning for Closed-Loop Dynamic Control of Soft Robotic Manipulators},
 pages = {124--134},
 volume = {35},
 number = {1},
 issn = {1552-3098},
 journal = {IEEE Transactions on Robotics},
 doi = {10.1109/TRO.2018.2878318}
}

@article{Wang.02.07.2021,
 author = {Wang, Rui and Yu, Rose},
 year = {2021},
 title = {Physics-Guided Deep Learning for Dynamical Systems: A Survey},
 url = {http://arxiv.org/abs/2107.01272v6}
}

@article{Wang.2021b,
 author = {Wang, Xiaomei and Li, Yingqi and Kwok, Ka-Wai},
 year = {2021},
 title = {A Survey for Machine Learning-Based Control of Continuum Robots},
 pages = {730330},
 volume = {8},
 journal = {Frontiers in Robotics and AI},
 doi = {10.3389/frobt.2021.730330}
}

@article{Wang.2024,
 author = {{X. Wang et al.}},
 year = {2024},
 title = {P{I}{N}{N}-{R}ay: A Physics-Informed Neural Network to Model Soft Robotic Fin Ray Fingers},
 url = {https://arxiv.org/abs/2407.08222},
 doi = {10.48550/arXiv.2407.08222}
}

@article{Wang.2024c,
 author = {Wang, Shuopeng and Wang, Rixin and Yang, Junjie and Hao, Lina},
 year = {2024},
 title = {Online Incremental Dynamic Modeling Using Physics-Informed Long Short-Term Memory Networks for the Pneumatic Artificial Muscle},
 pages = {8435--8442},
 volume = {9},
 number = {10},
 issn = {2377-3766},
 journal = {IEEE Robotics and Automation Letters},
 doi = {10.1109/LRA.2024.3446288}
}

@article{Webster.2010,
 author = {Webster, Robert J. and Jones, Bryan A.},
 year = {2010},
 title = {Design and Kinematic Modeling of Constant Curvature Continuum Robots: A Review},
 pages = {1661--1683},
 volume = {29},
 number = {13},
 issn = {0278-3649},
 journal = {The International Journal of Robotics Research},
 doi = {10.1177/0278364910368147}
}

@article{Werbos.1990,
 author = {Werbos, P. J.},
 year = {1990},
 title = {Backpropagation through time: what it does and how to do it},
 pages = {1550--1560},
 volume = {78},
 number = {10},
 issn = {00189219},
 journal = {Proceedings of the IEEE},
 doi = {10.1109/5.58337}
}

@article{Xavier.2022,
 author = {{M. S. Xavier et al.}},
 year = {2022},
 title = {Soft Pneumatic Actuators: A Review of Design, Fabrication, Modeling, Sensing, Control and Applications},
 pages = {59442--59485},
 volume = {10},
 journal = {IEEE Access},
 doi = {10.1109/ACCESS.2022.3179589}
}

@article{Yoon.2024,
 author = {{T. Yoon et al.}},
 year = {2024},
 title = {Kinematics-Informed Neural Networks: Enhancing Generalization Performance of Soft Robot Model Identification},
 pages = {3068--3075},
 volume = {9},
 number = {4},
 issn = {2377-3766},
 journal = {IEEE Robotics and Automation Letters},
 doi = {10.1109/LRA.2024.3362644}
}

@article{Zeipel.2024,
 author = {Zeipel, Henrik and Habich, Tim-Lukas and Seel, Thomas and Ehlers, Simon F. G.},
 year = {2024},
 title = {Fast and Accurate Prediction of Vehicle Dynamics using Physics-Informed Neural Networks},
 url = {https://doi.org/10.36227/techrxiv.173398186.65085317/v1},
 journal = {{T}echrXiv},
 doi = {10.36227/techrxiv.173398186.65085317/v1}
}
